\documentclass[10pt,twocolumn,letterpaper]{article}

\usepackage{wacv}
\usepackage{times}
\usepackage{graphicx}
\usepackage{amsmath}
\usepackage{amssymb}
\usepackage{booktabs}
\usepackage{float}

\usepackage{comment}
\usepackage{amsmath,amssymb} 
\usepackage{color}

\usepackage{pdflscape}

\usepackage{bm}
\usepackage{array}
\usepackage{arydshln}
\usepackage{booktabs}
\usepackage{subcaption}
\definecolor{myGray}{RGB}{80,80,80}
\definecolor{myGreen}{RGB}{120, 170, 120}
\definecolor{myOrange}{RGB}{255, 120, 40}
\definecolor{myBlue}{RGB}{60, 130, 180}

\newcommand{\bfstatG}[1]{{\textcolor{myGreen}{\bf #1}}}
\newcommand{\bfstatO}[1]{{\textcolor{myOrange}{\bf #1}}}
\newcommand{\bfstatB}[1]{{\textcolor{myBlue}{\bf #1}}}
\newcommand\sbullet[1][.8]{\mathbin{\vcenter{\hbox{\scalebox{#1}{$\bullet$}}}}}

\def\wacvPaperID{} 

\wacvfinalcopy 

\ifwacvfinal
\usepackage[breaklinks=true,bookmarks=false,backref=page]{hyperref}
\else
\usepackage[pagebackref=true,breaklinks=true,colorlinks,bookmarks=false]{hyperref}
\fi

\begin{document}

\title{CYBORG: Blending Human Saliency Into the Loss Improves Deep Learning}

\author{Aidan Boyd,~Patrick Tinsley,~Kevin Bowyer,~Adam Czajka \\
University of Notre Dame, Notre Dame IN 46556, USA \\
\{aboyd3,ptinsley,kwb,aczajka\}@nd.edu
}

\newcommand{\teaser}{
{
   \begin{center}
        \centering
            \begin{minipage}{\textwidth}
                \centering
                \includegraphics[width=\textwidth]{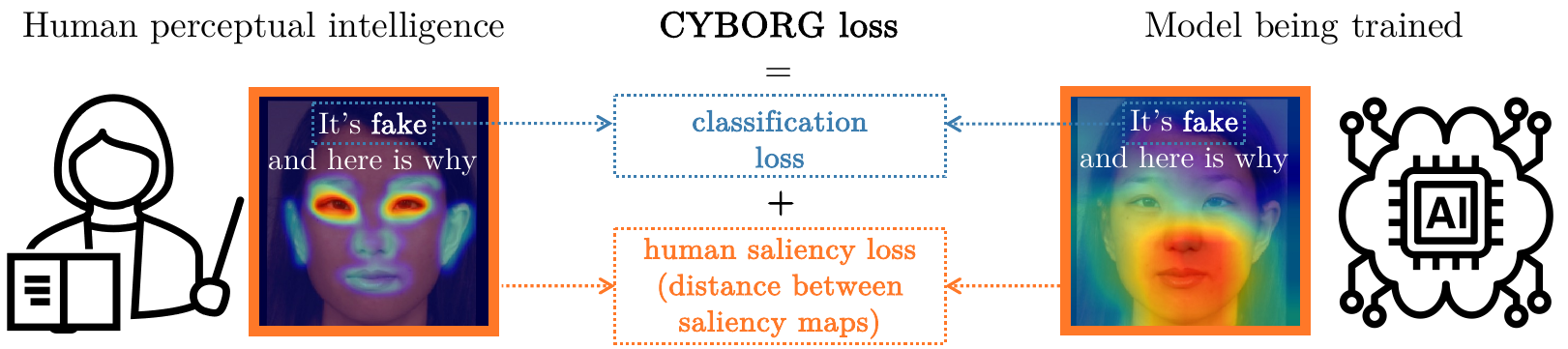}
            \end{minipage}%
            \captionof{figure}{
            Our proposed training strategy {\bf C}onve{\bf Y}s {\bf B}rain {\bf O}versight to {\bf R}aise {\bf G}eneralization. CYBORG continually encourages the training process to look at image regions judged as salient for human visual perception. This results in a model that is more likely to learn features from regions that are salient to humans, and less likely to learn features that are accidentally correlated with class labels. A boost in generalization performance is demonstrated.
            }
            \label{fig:teaser}
    \end{center}
    \vspace{-1em}
}
}

\maketitle
\thispagestyle{empty}

\begin{abstract}
Can deep learning models achieve greater generalization if their training is guided by reference to human perceptual abilities?
And how can we implement this in a practical manner? This paper proposes a training strategy to ConveY Brain Oversight to Raise Generalization (CYBORG). This new approach incorporates human-annotated saliency maps into a loss function that guides the model's learning to focus on image regions that humans deem salient for the task.
The Class Activation Mapping (CAM) mechanism is used to probe the model's current saliency in each training batch, juxtapose this model saliency with human saliency, and penalize large differences. 
Results on the task of synthetic face detection, selected to illustrate the effectiveness of the approach, show that CYBORG leads to significant improvement in accuracy on unseen samples consisting of face images generated from six Generative Adversarial Networks across multiple classification network architectures. We also show that scaling to even seven times the training data, or using non-human-saliency auxiliary information, such as segmentation masks, and standard loss cannot beat the performance of CYBORG-trained models. As a side effect of this work, we observe that the addition of explicit region annotation to the task of synthetic face detection increased human classification accuracy.
This work opens a new area of research on how to incorporate human visual saliency into loss functions in practice. 
All data, code and pre-trained models used in this work are offered with this paper.
\end{abstract}

\vspace{-1em}
\section{Introduction}
\label{sec:intro}

How do you teach a child to ride a bicycle? The passive option is to set the child on the bike, give the bike a push and then stand back silently, watching what happens. The active option is to set the child on the bike, give them a push, and then run alongside, continually giving advice on what to do. We argue that current state of training of deep learning-based models is more passive than active. We introduce a new training process that -- by incorporating human visual perception into a loss function -- continually reminds the model being trained of the image regions judged as salient to humans, as illustrated in Fig. \ref{fig:teaser}. 

The main goal of the proposed CYBORG approach is to {\bf c}onve{\bf y} {\bf b}rain {\bf o}versight to {\bf r}aise {\bf g}eneralization by 
encouraging the deep learning model to focus on human-salient regions.
This is achieved by adding a new component to the loss, based on the difference between human saliency heatmaps and the model's class activation mapping-based \cite{zhou2016learning} heatmaps in each training batch.
Thus our new loss function blends classical data-driven optimization with human-derived oversight or ``coaching'' about the parts of the image that are salient to the problem.

To demonstrate the advantages of CYBORG training,
we apply it to the challenging task of distinguishing face images that are authentic versus generated by various modern Generative Adversarial Nets (GANs). 
To generate human-derived saliency maps, we presented 1,000 pairs of face images to 363 humans.
Each image pair contained one authentic (real) and one synthetic image (generated by an example deep learning-based approach, StyleGAN2~\cite{karras2020analyzing}, and non deep learning-based method  SREFI~\cite{Banerjee_IJCB_2017}). Viewers were asked to (a) choose which face was authentic and which was synthetic, and (b) annotate regions that support their decision. For each image, annotations from the viewers were compiled into a {\it saliency map} summarizing human judgment about the salient image regions.

Our experiments show that CYBORG learning increases the accuracy of detecting synthetic data in an open-set classification regime, in which test samples are generated with six different GAN architectures withheld during the training process. 
We also demonstrate that although adding human saliency to an example model implementing an attention mechanism~\cite{dang2020detection} increases performance, this improvement is small compared to when the CYBORG approach is employed.
The {\bf main contributions} of this work are:

\noindent
~~$\sbullet$ {\bf Introduction of the CYBORG training strategy}, which benefits from human judgment about salient regions by incorporating perceptual intelligence into loss function. 

\noindent
~~$\sbullet$ {\bf Open-set evaluation of CYBORG training} that shows a significant improvement for multiple state-of-the-art deep learning models (ResNet, DenseNet, Inception and Xception), as well as for the existing synthetic face detector.

\noindent
~~$\sbullet$ {\bf Experiments assessing the ``value'' of human annotations in two ways:} (a) demonstrating that at least 7 times more training data is needed to train a model in classical fashion to achieve performance competitive with CYBORG, and (b) replacing human saliency maps with non-human-sourced cues offered by face segmentation masks, which did not achieve the level of generalization achieved by the CYBORG-trained models.

\noindent
~~$\sbullet$ {\bf Evaluation of state-of-the-art ``deep fake'' detector on GAN-generated face images}, which illustrates that the solutions to ``deep fake'' detection and synthetic face detection are not cross-applicable.

\noindent
~~$\sbullet$ Results demonstrating that {\bf human classification accuracy increases when participants are asked to annotate image regions that support their decisions}, compared to the same experiment without annotations.

\noindent
~~$\sbullet$ {\bf Data and source codes to reproduce all experiments:} a test set containing 600,000 synthetic faces generated by six GAN architectures (ProGAN, StarGANv2, StyleGAN, StyleGAN2, StyleGAN2-ADA and StyleGAN3), human annotation data, and all neural network models.

\section{Related Work}
\label{sec:related_work}

\noindent{\bf Synthetic Image Generation and Detection.} Since Goodfellow \etal ~\cite{goodfellow2014generative}, many  
open-source, and (often) pre-trained GANs for image synthesis have become available~\cite{karras2017progressive,karras2019style,karras2020analyzing,Karras2020ada,Karras2021,choi2020stargan,brock2018large,zhu2017unpaired,park2019gaugan}. 
The authors of~\cite{qian2020thinking,frank2020leveraging} maintain that frequency domain analysis can unveil artifacts or manipulations in GAN-generated images across different model architectures, datasets, and resolutions. However, as documented by Marra \etal~\cite{marra2018detection}, conventional, non-deep-learning methods (such as steganalysis~\cite{cozzolino2014image}) fail in the presence of compression. With a virtually infinite number of fake samples in their training processes, deep networks such as ResNet~\cite{he2016deep}, DenseNet~\cite{huang2017densely}, InceptionNet~\cite{szegedy2016rethinking}, and Xception-Net~\cite{chollet2017xception} have achieved over 99\% accuracy in fake image recall~\cite{tariq2019gan}. Even before public release of StyleGAN3~\cite{Karras2021}, there were several proactive efforts towards detecting StyleGAN3 images~\cite{kitware,littlejuyan,yaseryacoob,grip-unina,polimi-ispl,wang2019cnngenerated}. These models can be complemented with the proposed CYBORG loss, and such an attempt, with a model proposed by Wang \etal~\cite{wang2019cnngenerated}, is described in the supp. materials.

Although the generation of never-before-seen images lends itself naturally to the creative process, the ability to manipulate existing images poses a significant security problem ~\cite{botha2020fake,chesney2019deepfakes}. A commonly commercialized scapegoat is deepfakes~\cite{deepfakes2021}, which splices real identities onto realistic-looking videos. We demonstrate in this paper that state of the art deep fake detectors may not be effective in detecting fully-synthetic samples, which this paper focuses on.

\vskip1mm\noindent{\bf Using Human Perception to Understand / Improve Computer Vision.}
O'Toole \etal~\cite{o2012comparing} demonstrated that machines were never less accurate than humans on face images of various quality. RichardWebster \etal~\cite{richardwebster2018visual} showed that observing human face recognition behavior in certain contexts can be used to explain why face matchers succeed or fail, leading to better model explainability. 
In biometrics, human saliency was found complementary to algorithm saliency and thus beneficial to combine them~\cite{trokielewicz2019perception,moreira2019performance}. Czajka \etal measured human visual saliency via eye tracking and used it to build human-driven filtering kernels for iris recognition~\cite{czajka2019domain}, achieving better performance than non-human-driven approaches. Human-guided training data augmentation was proposed by Boyd \etal~\cite{boyd2021human} to build deep learning-based iris presentation attack detection methods generalizing exceptionally well to unknown attack types. 

In broader machine learning, incorporation of results from psychophysics has aided in deep learning tasks such as image captioning for scene understanding~\cite{he2019human,huang2021specific}, handwriting analysis~\cite{grieggs2021measuring}, and natural language processing~\cite{zhang2020human}. Linsley \etal~\cite{linsley2018learning} proposed to incorporate bottom-up, human-sourced saliency into a self-attention mechanism, combining global and local attention in the ``GALA'' module in their deep learning models. We demonstrate in the supp. materials how our human saliency maps can be incorporated into the attention mechanism, and show that CYBORG allows for a better gain in accuracy than using human saliency in the attention mechanism.

\vskip1mm\noindent{\bf Differences between the proposed CYBORG method and previous works:} (a) human spatial saliency and model spatial saliency have never before been directly compared and blended into overall loss; (b) CYBORG does not require architectural changes to the model \textit{e.g.}, a specialized attention module.

\section{Experimental Datasets}
\label{sec:data_collection}

\begin{figure}
    \centering
    \includegraphics[width=\columnwidth]{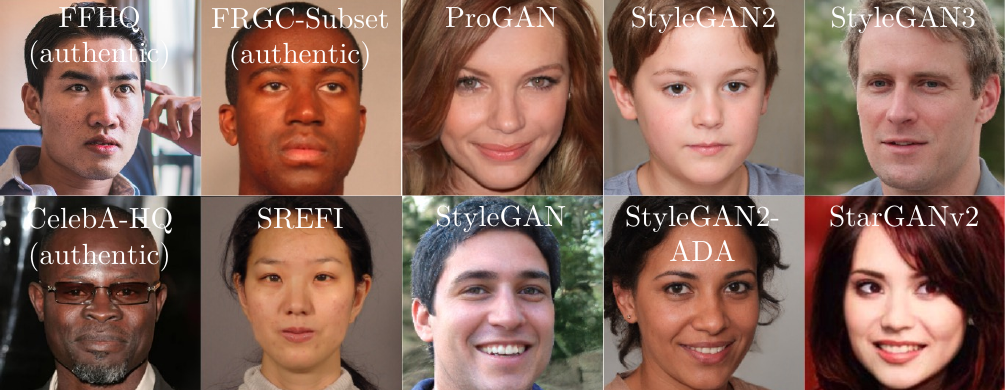}
    \vskip-1mm
    \caption{Examples from each data source.}
    \label{fig:test_data}
    \null\vskip-5mm
\end{figure}

Two types of face image datasets are used: authentic datasets consisting of real images from three sources (CelebA-HQ~\cite{karras2017progressive}, Flickr-Faces-HQ~\cite{karras2019style} and FRGC-Subset~\cite{Phillips_IVC_2017}), and synthetic datasets consisting of fake images from seven different generators (ProGAN, StyleGAN, StyleGAN2, StyleGAN2-ADA, StyleGAN3, StarGANv2 and SREFI~\cite{karras2017progressive,karras2019style,karras2020analyzing,Karras2020ada,Karras2021,choi2020stargan,Banerjee_IJCB_2017}).
Along with Fig. \ref{fig:test_data}, the following sections briefly characterize data sources.

\vskip1mm\noindent{\bf CelebA-HQ}
~\cite{karras2017progressive} is a high-quality version of the original CelebA dataset~\cite{liu2015faceattributes}, containing 30,000 images of celebrities at a resolution of $1024\times1024$. 

\vskip1mm\noindent{\bf Flickr-Faces-HQ (FFHQ)} includes 70,000 $1024\times1024$ images of faces varying in age, ethnicity, and facial accessories (glasses, hats, etc.) \cite{karras2019style}. 

\vskip1mm\noindent{\bf FRGC-Subset} dataset contains 16,433 face images, randomly sampled from a set of publicly available datasets collected by Phillips \etal~\cite{Phillips_IVC_2017}. 
Images show frontal faces varying in expression, ethnicity, gender, and age.

\vskip1mm\noindent{\bf SREFI} was generated by the ``synthesis of realistic face images'' (SREFI)~\cite{Banerjee_IJCB_2017} method, which works by first matching similar face images based on VGG-Face features, splitting them into region-specific triangles, and implanting from donor faces onto a base face to create a blended identity. To ensure consistency, important facial features, such as the mouth and eyes, on the generated image are required to come from the same donor.

\vskip1mm\noindent{\bf ProGAN } contains 100,000 images downloaded from~\cite{karras2017repo}. Unlike its successors (StyleGAN), Karras \etal's ProGAN generator network was trained on CelebA-HQ images~\cite{karras2017progressive}.

\vskip1mm\noindent{\bf The StyleGAN Family.} The next four synthetic datasets were generated with StyleGAN architectures~\cite{karras2019style,karras2020analyzing,Karras2020ada,Karras2021}. The original StyleGAN was trained in a similar manner to its predecessor ProGAN~\cite{karras2017progressive}, but with the added feature of mixable disentangled layers for style transfer.
The next version, StyleGAN2~\cite{karras2020analyzing}, removed artifacts found in original StyleGAN images and improved image reconstruction via path length regularization.
The third iteration of StyleGAN, StyleGAN2 with adaptive discriminator augmentation)~\cite{Karras2020ada}, solves for training GANs in data-limited scenarios. 
Finally, StyleGAN3~\cite{Karras2021} mitigates aliasing in rotation- and translation-invariant generator networks.

For original StyleGAN and StyleGAN2, sets of 100,000 fake face images were downloaded from their GitHub repositories. For StyleGAN2-ADA and StyleGAN3, sets of 100,000 images were generated using default generator settings, including a truncation of ($\psi$) of 0.5 (as recommended by StyleGAN authors).

\vskip1mm\noindent{\bf StarGANv2} produces images with the main focus of style transfer~\cite{choi2020stargan}, unlike StyleGAN. Generated images show source identities ``dressed'' in the style of the supplied reference images. In order to ensure high facial quality of StarGANv2 generated images, 250,000 images were initially synthesized using a supplied network (pre-trained on CelebA-HQ). 
These synthetic samples were then scored and sorted according to facial quality using FaceQNet~\cite{hernandez2020biometric}, a CNN designed to assess input images' suitability for face recognition tasks. The final dataset consisted of the top-ranked 100,000 images. 

\section{Human Saliency}
\label{sec:human_saliency}

\subsection{Acquisition of Human Salient Regions}

We replicate an experiment of Shen \etal~\cite{shen2021study}, in which subjects judge pairs of non-masked face images as fake or real, but we require subjects to annotate regions supporting their decisions. 
Specifically, participants are presented with a pair of face images (one a synthetically-generated identity, and the other an authentic facial image), and asked to decide which image is either the synthetic image or the real image in a two-alternative forced choice (2AFC) manner. The prompt question alternated between asking \textit{which is real} versus \textit{which is fake}\footnote{The online annotation tool developed for this work is presented in supp. materials.}. Next, users were asked to highlight regions (not size- nor location-constrained) of the image supporting their classification decision.

\begin{figure}[t!]
    \centering
    \begin{subfigure}[t]{0.44\columnwidth}
        \centering
        \includegraphics[width=\columnwidth]{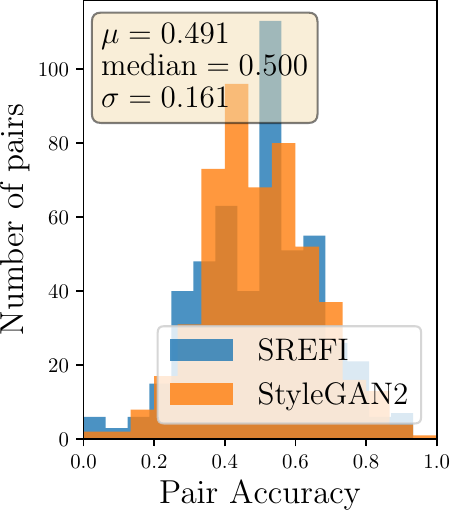}
        \caption{Only classification}
    \end{subfigure}%
    \hskip 10mm
    \begin{subfigure}[t]{0.44\columnwidth}
        \centering
        \includegraphics[width=\columnwidth]{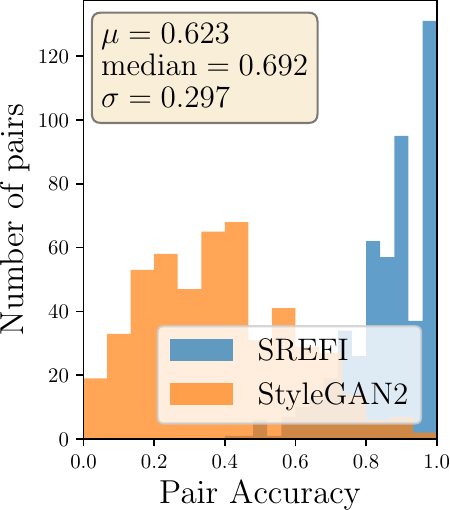}
        \caption{With annotations}
    \end{subfigure}\vskip-1mm
    \caption{Human classification of face images as real/fake is more accurate when subjects annotate images: (a) Shen \etal \cite{shen2021study} found that humans did not accurately classify face images as real/fake; (b) the same experiment in which we asked subjects to annotate the image regions that supported their classification. Average human accuracy is substantially higher in study (b). These histograms detail the accuracy of humans on the 1,000 pairs. 
    }
    \label{fig:annot_results}
    \vskip-4mm
\end{figure}

Data (decisions and annotations) was collected from 363 subjects (recruited via Amazon Mechanical Turk), with an average of 29.6 image pairs processed by each subject. The synthetic images consisted of 500 images generated using SREFI from the FRGC-Subset dataset, and 500 images synthesized by StyleGAN2 and downloaded from \url{thispersondoesnotexist.com}. 10,750 annotations were obtained, matching exactly the number of question/pair samples in Shen's work for fair comparison. For evaluating the CYBORG approach, only annotations for \textbf{correctly} classified pairs were used in the training process. 

\subsection{Do Annotations Improve Human Accuracy?}

As the only difference between our protocol and that used in ~\cite{shen2021study} was the annotation requirement, we can properly diagnose how annotating images impacts decision accuracy.
Fig. \ref{fig:annot_results}(a) outlines the original results\footnote{We thank the authors of~\cite{shen2021study} for sharing the raw results with us.}, where the blue and orange histograms represent results attained for 1,000 image pairs with synthetic data generated by SREFI and StyleGAN2 approaches, respectively. As can be seen, human accuracy is at random chance level when not asked to annotate regions to support their decision. However, Fig. \ref{fig:annot_results}(b) shows that the accuracy increased from 50\% (random chance) to 69.2\% simply by requesting users to annotate the images (and so spend more time on each pair). This experiment suggests that human accuracy in detecting synthetic faces can be improved simply by forcing annotations that support classification decisions. A side observation is that StyleGAN2 images may appear more realistic than SREFI images in such new setup (see the shift between distributions in Fig. \ref{fig:annot_results}(b)).

\begin{figure*}[!ht]
    \centering
    \begin{subfigure}[b]{1\textwidth}
        \begin{subfigure}[b]{0.19\textwidth}
          \centering
          \includegraphics[width=1\columnwidth]{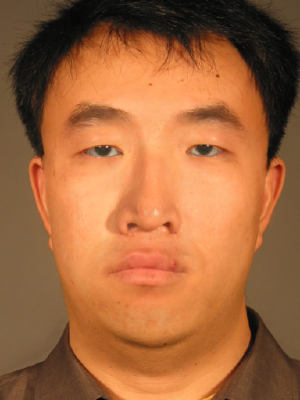}
          \caption{~}
        \end{subfigure}
        \hfill
        \begin{subfigure}[b]{0.38\textwidth}
            \centering
            \includegraphics[width=1\columnwidth]{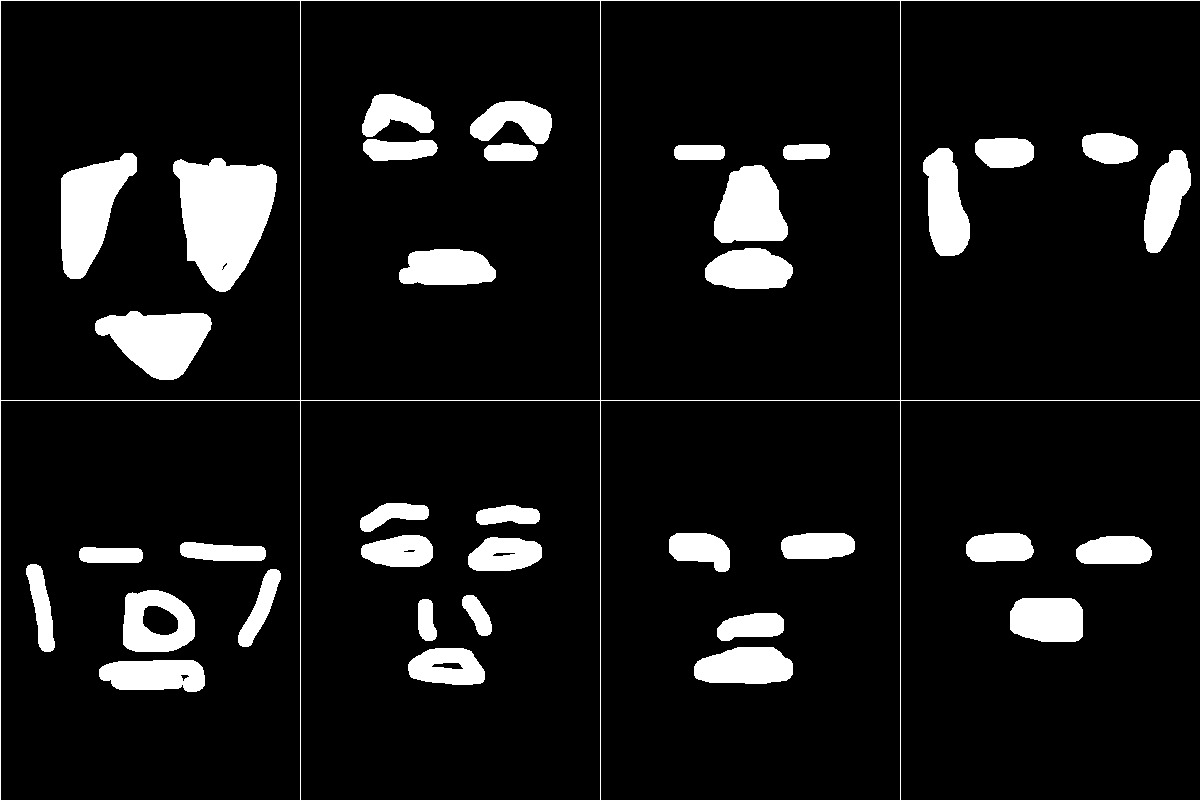}
            \caption{~}
        \end{subfigure}
        \hfill
        \begin{subfigure}[b]{0.19\textwidth}
          \centering
          \includegraphics[width=1\columnwidth]{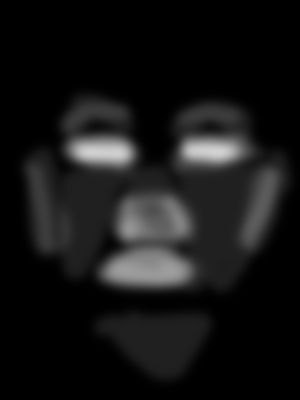}
          \caption{~}
        \end{subfigure}
        \hfill
        \begin{subfigure}[b]{0.19\textwidth}
          \centering
          \includegraphics[width=1\columnwidth]{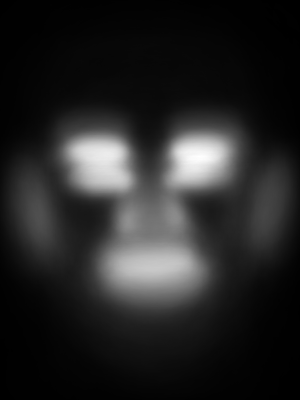}
          \caption{~}
        \end{subfigure}
    \end{subfigure}\vskip-1mm
    \caption{
    Creation of human saliency map: (a) an image (in this example generated by SREFI) presented to human annotators; (b) eight annotations from viewers who correctly classified the image; (c) averaged annotations defining the human salient features for that sample, as used by CYBORG; (d) average of all human annotations for all images in the training set.
    }
    \label{fig:incorp_process}
    \null\vskip-5mm
\end{figure*}

\subsection{Building Human Saliency Maps}

All correct annotations, as shown in eight individual images in Fig. \ref{fig:incorp_process}(b), are combined together with equal weight to create image representations called {\bf human saliency maps} shown in Fig. \ref{fig:incorp_process}(c). A Gaussian blur of $\sigma=5$ is applied to the combined array to smooth edges between regions of varying annotation density, and the map is scaled to the range of $\langle 0,1 \rangle$.
White pixels in the saliency map correspond to regions that more subjects had annotated as important. Black pixels correspond to areas not annotated by any subject.  

After data collection, there were 1,821 correctly classified images with annotations, consisting of 919 authentic images and 902 synthetic images. These 1,821 images represent the training set for the CYBORG loss experiments.

\section{CYBORG Loss}

In the same way a cyborg is a human-machine hybrid, the proposed CYBORG training strategy combines the human saliency information attained through annotations (\textit{human saliency loss component}) with a requirement for high classification accuracy (\textit{classification loss component}). The former component steers activations in the feature maps in the last convolutional layer to be aligned with human-defined regions of importance, while the model may still benefit from a data-driven learning approach owing to the latter component.

More specifically, the human saliency loss directly compares the difference in salient regions between machine and human during training. 
To accomplish this, we implemented a fully-differentiable Class Activation Mapping (CAM) approach~\cite{zhou2016learning} that, given the current weights, can generate CAMs for all samples in each training batch. Resultant CAMs are scaled to the range of $\langle 0,1 \rangle$, human saliency maps are downsized to the same size as CAMs, and then both heatmaps are compared via $\ell_2$ norm. Formally, we define CYBORG loss $ \mathcal{L}$ as:

\vskip-3mm
\begin{equation}
\begin{split}
\mathcal{L} = \frac{1}{K}\sum_{k=1}^K\sum_{c=1}^{C}\bm{1}_{y_k \in \mathcal{C}_c} \Bigg[\underbrace{(1-\alpha)\|\textbf{s}_k^{\text{(human)}}  - \textbf{s}_k^{\text{(model)}}\|^2}_{\text{human saliency loss component}} \\-\underbrace{\alpha\log p_{\text{model}}\big(y_k \in \mathcal{C}_c\big)}_{\text{classification loss component}}\Bigg]
\end{split}
\end{equation}

\noindent
where $\|\cdot\|$ is the $\ell_2$ norm, $y_k$ is a class label for the $k$-th sample, $\bm{1}$ is a class indicator function equal to $1$ when $y_k \in \mathcal{C}_c$ (0 otherwise), $C$ is the total number of classes, $K$ is the number of samples in a batch, $\alpha=0.5$ is a trade-off parameter weighting human- and model-based saliencies, $\textbf{s}_k^{\text{(human)}}$ is the human saliency for the $k$-th sample, and

$$
\textbf{s}_k^{(\text{model})} = \textbf{f}_1w_1^{(c)} + \textbf{f}_2w_2^{(c)} + \dots + \textbf{f}_Nw_N^{(c)}
$$
\noindent
is a class activation map-based model's saliency for the $k$-th sample, where $N$ is the number of feature maps $\textbf{f}$ in the last convolutional layer, and $w^{(c)}$ are the weights in the last classification layer belonging to predicted class $\mathcal{C}_c$. Both $\textbf{s}_k^{(\text{model})}$ and $\textbf{s}_k^{(\text{human})}$ are normalized to the range $\langle 0,1\rangle$.

The reason that the CAM method was implemented rather than a more modern approach (GradCAM~\cite{Selvaraju_ICCV_2017} or EigenCAM~\cite{Muhammad_IJCNN_2020}) is that the latter approaches require backpropagation to calculate gradients with respect to the input to determine salient regions (in addition to gradients with respect to the weights). This is expensive to do during training while maintaining differentiability, and these methods are typically only used on fully trained models where backward calls can be completed in a post-hoc fashion. For CAM, only a forward pass is necessary, meaning it can be bootstrapped into the training strategy directly.

\section{Experimental Setup for CYBORG}
\label{sec:experiments}

Face images are aligned using {\it img2pose}~\cite{Albiero_CVPR_2021}, cropped, and resized to $224\times224$. Face bounding boxes are expanded 20\% in all directions before cropping, with an additional 30\% on the forehead to ensure the head is fully in view. Human saliency maps are resized and cropped the same, to keep spatial correspondence. 

\subsection{Training Scenarios}
\label{sec:scenarios}

\vskip1mm\noindent{\bf Scenario 1: Classical Training.} The 
basic scenario consists of training the studied architectures in a task of synthetic face image detection, on image data for which human saliency information was collected, but 
using only the classification component in the loss function (\ie no human annotations are used). 
The training set in this scenario consists of 919 authentic and 902 synthetic images. The validation set consists of 20,000 images: 10,000 authentic images, 5,000 images generated using SREFI, and 5,000 images downloaded from \textit{thispersondoesnotexist.com}. The training and validation set used in this scenario will be further referred to as the \textit{original data}.

\vskip1mm\noindent{\bf Scenario 2: Classical Training with Large Data.} 
To evaluate how much additional data is required to achieve CYBORG-level performance from learning with only classification loss (as in Scenario 1), we curate a larger dataset than that used in Scenario 1. 
Starting from the original data, we add six times more samples resulting in a training set $7\times$ the initial size.
Scarcity of authentic images in the source datasets prevented going beyond $7\times$, as adding data from different source could add bias to the comparison.

\vskip1mm\noindent{\bf Scenario 3: CYBORG Training.} Using the
\textit{original data} as in Scenario 1, we apply the same training strategy but also include the human saliency component in the loss function to create CYBORG loss. 
The only difference between Scenarios 1 and 3 is the loss function, so observations can be directly correlated with the effectiveness of CYBORG training. 

\vskip1mm\noindent{\bf Experimental Parameters.} To ensure that observations are not architecture-specific, the base experiments are completed on four out-of-the-box architectures: DenseNet-121~\cite{huang2017densely}, ResNet50~\cite{he2016deep}, Inception v3~\cite{szegedy2016rethinking} and XceptionNet~\cite{chollet2017xception}. 
For all methods, Stochastic Gradient Descent (SGD) is used, with learning rate of $0.005$, modified by a factor of $0.1$ every 12 epochs. Training ran for 50 epochs, and weights giving the highest validation accuracy were selected as the final model. 
The validation set is constant, as described in Scenario 1. 
Networks are instantiated from the pre-trained ImageNet weights~\cite{pytorchModelZoo}. For all experiments using CYBORG loss, the human saliency and the classification components are given equal weight ($\alpha=0.5$). Each architecture/scenario pair is independently trained 10 times, to generate error statistics on the test set.

\subsection{Testing Protocol}

To evaluate accuracy of the models trained under the three scenarios, we composed a comprehensive test set of 100,000 synthetically generated images from each of six different GAN architectures, ending up with 600,000 total test samples. The authentic face datasets used for testing are the FFHQ dataset (70,000 images) and the CelebA-HQ dataset (30,000 images). For ProGAN and StarGANv2, the training data is CelebA-HQ; for the remaining four StyleGAN sets, the training data is FFHQ. This setup aims at demonstrating whether models can differentiate between authentic samples and synthetic samples, where the latter are generated by a GAN trained on the former.

\subsection{Evaluating State-Of-The-Art DeepFake Detector on Test Data}
\label{sec:pretrained_sota}

In order to properly compare our CYBORG models against existing deepfake detectors, we evaluated the state-of-the-art ensemble method from Bonettini \etal~\cite{bonettini2021video} on \textit{our} test set of synthetic images. 
Of the ten available models, five were trained on the DeepFake Detection Challenge (DFDC) dataset~\cite{DFDC2020}, and five were trained on the FaceForensics++ (FF++) dataset~\cite{roessler2019faceforensicspp}. 

For each dataset, ensemble methods were composed, using models trained on DFDC or models trained on FF++. Before evaluating on \textit{our} test data of synthetic images, we verified model performance by evaluating the reported top two ensemble methods (one for DFDC, one for FF++) on Bonettini \etal test deepfake data. We then ran the same two top-performing ensemble methods on our test data to compare results with CYBORG-trained models. 

\subsection{Assessing The Value Of Human Annotations} 
\label{sec:face_segmentation}

To determine the usefulness of human annotations in the CYBORG loss function, a comparison to a non-human-saliency-guided baseline is needed. A face parsing tool, BiSeNet \cite{bisenet2019}, is applied to the training images to attain a mask detailing all facial regions\footnote{example masks can be found in the supp. materials} and CYBORG training is applied with BiSeNet segmentation masks instead of human saliency maps. The goal of this experiment is to determine  
whether human saliency maps provide better cues than automatically-determined face masks. An affirmative answer could limit the costs of human saliency acquisition.

\section{Evaluation Results}
\label{sec:evaluation}

\begin{figure*}[t]
  \begin{subfigure}[b]{1\textwidth}
      \begin{subfigure}[b]{0.32\textwidth}
          \centering
            \includegraphics[width=1\columnwidth]{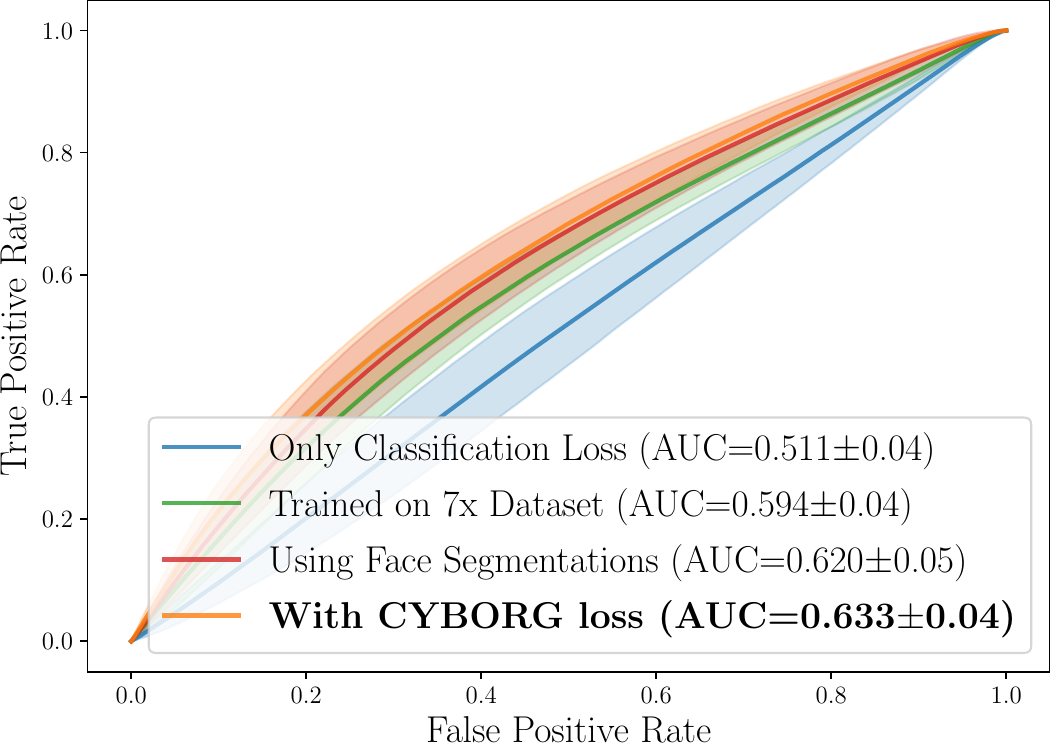}
          \caption{DenseNet-121}
      \end{subfigure}
      \hfill
      \begin{subfigure}[b]{0.32\textwidth}
          \centering
          \includegraphics[width=1\columnwidth]{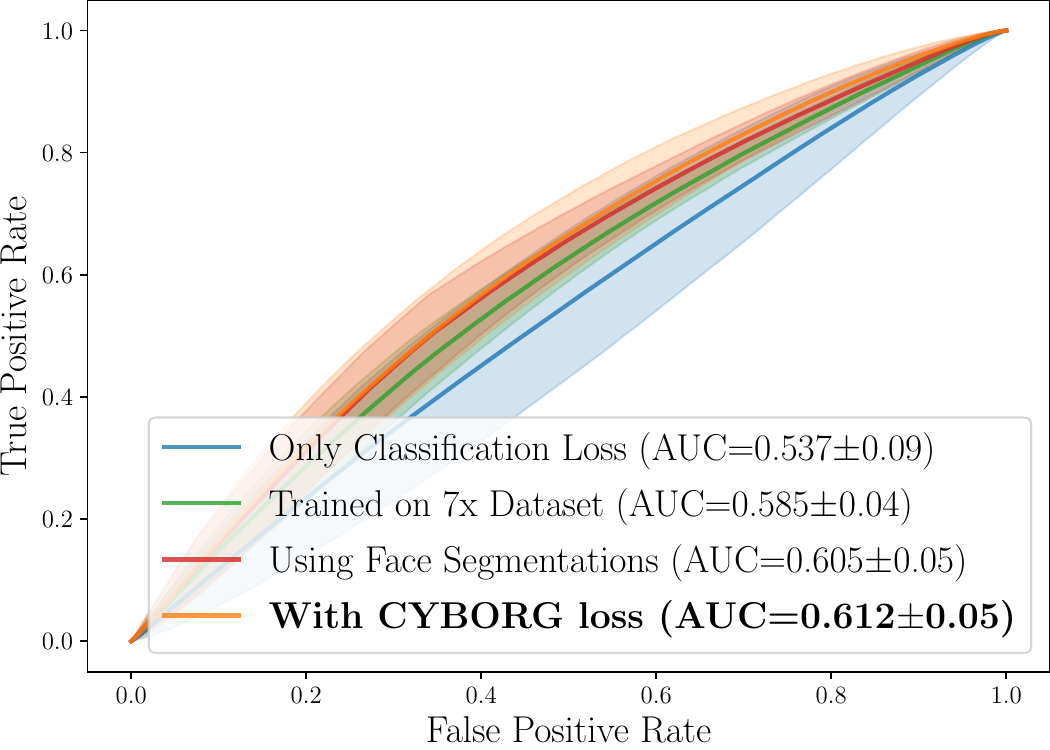}
          \caption{ResNet50}
      \end{subfigure}
      \hfill
      \begin{subfigure}[b]{0.32\textwidth}
          \centering
          \includegraphics[width=1\columnwidth]{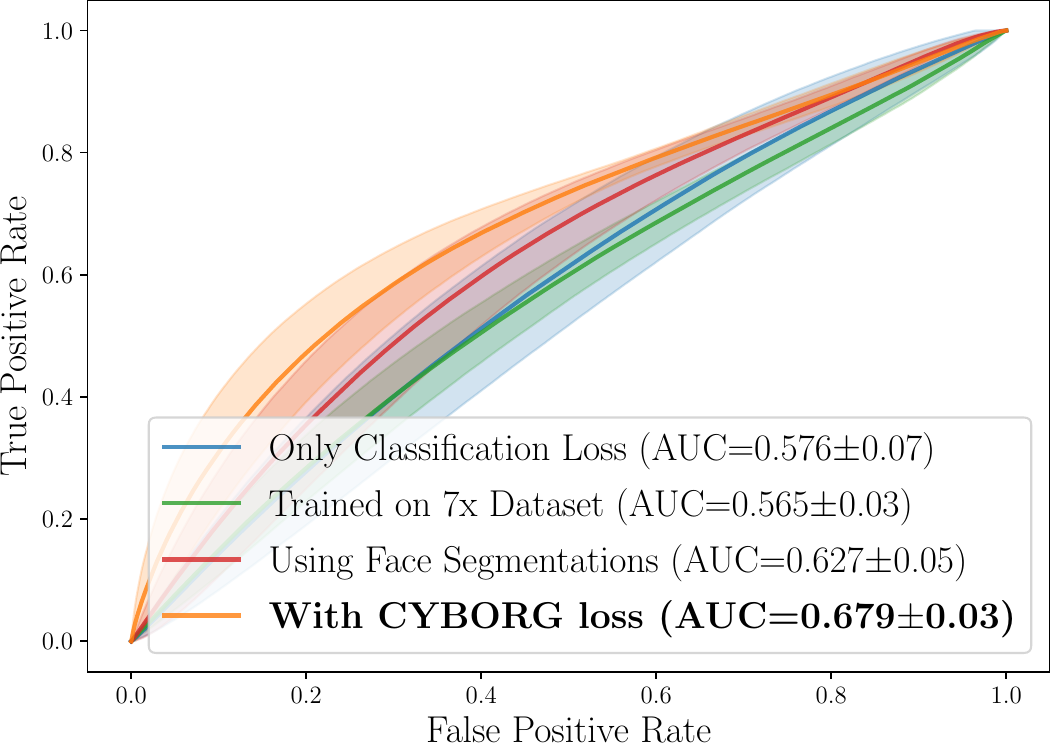}
          \caption{Inception-v3}
      \end{subfigure}

  \end{subfigure}\vskip3mm
  \begin{subfigure}[b]{1\textwidth}
      \centering
      \begin{subfigure}[b]{0.32\textwidth}
          \centering
          \includegraphics[width=1\columnwidth]{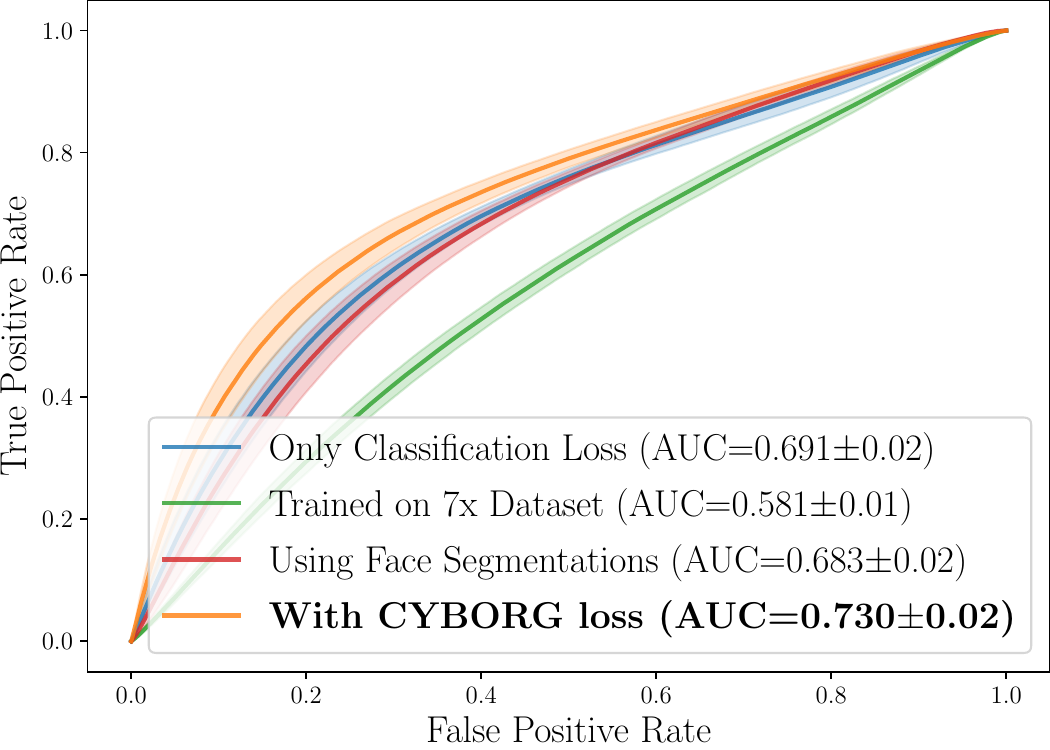}
          \caption{Xception}
      \end{subfigure}
     \begin{subfigure}[b]{0.305\textwidth}
          \centering
          \includegraphics[width=1\columnwidth]{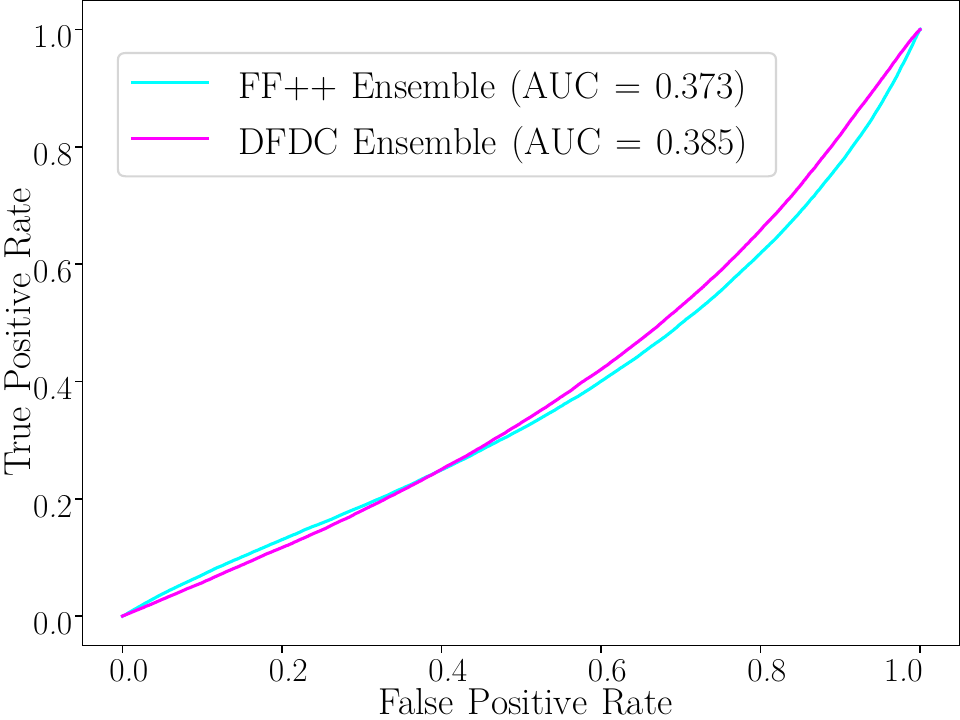}
          \caption{SOTA Deepfake Detector~\cite{bonettini2021video}}
      \end{subfigure}
  \end{subfigure}\vskip-1mm
  \caption{ROC curves presenting results on the test set consisting of six GAN types for four architectures and one off-the-shelf deepfake detector. Shaded regions in (a)-(d) correspond to $\pm1$ standard deviation of the False Positive Rate (FPR) for a given True Positive Rate (TPR). Results outline that in all cases that CYBORG loss was employed (a-d), an increase in performance compared to classification loss alone can be observed. 
  Additionally, in all (a-d) results CYBORG outperforms the models trained even on seven times the training set with just classification loss, and models trained with face segmentation masks instead of human saliency maps. 
  }
  \label{fig:roc}
  \null\vskip-5mm
\end{figure*}

Figure \ref{fig:roc} summarizes the performance observed for each of the four studied architectures by presenting ROC curves obtained for the comprehensive set of all $100,000$ authentic and $600,000$ synthetically-generated test samples\footnote{ROC curves for individual GANs can be found in the supp. material.}. 
For all experiments, training and validation is repeated 10 times in order to assess statistical significance of the observed differences in the results. Area Under the Curve (AUC) is given along with $\pm$ one standard deviation across the 10 runs.

\begin{figure}[t] 
    \centering
    \includegraphics[width=0.95\columnwidth]{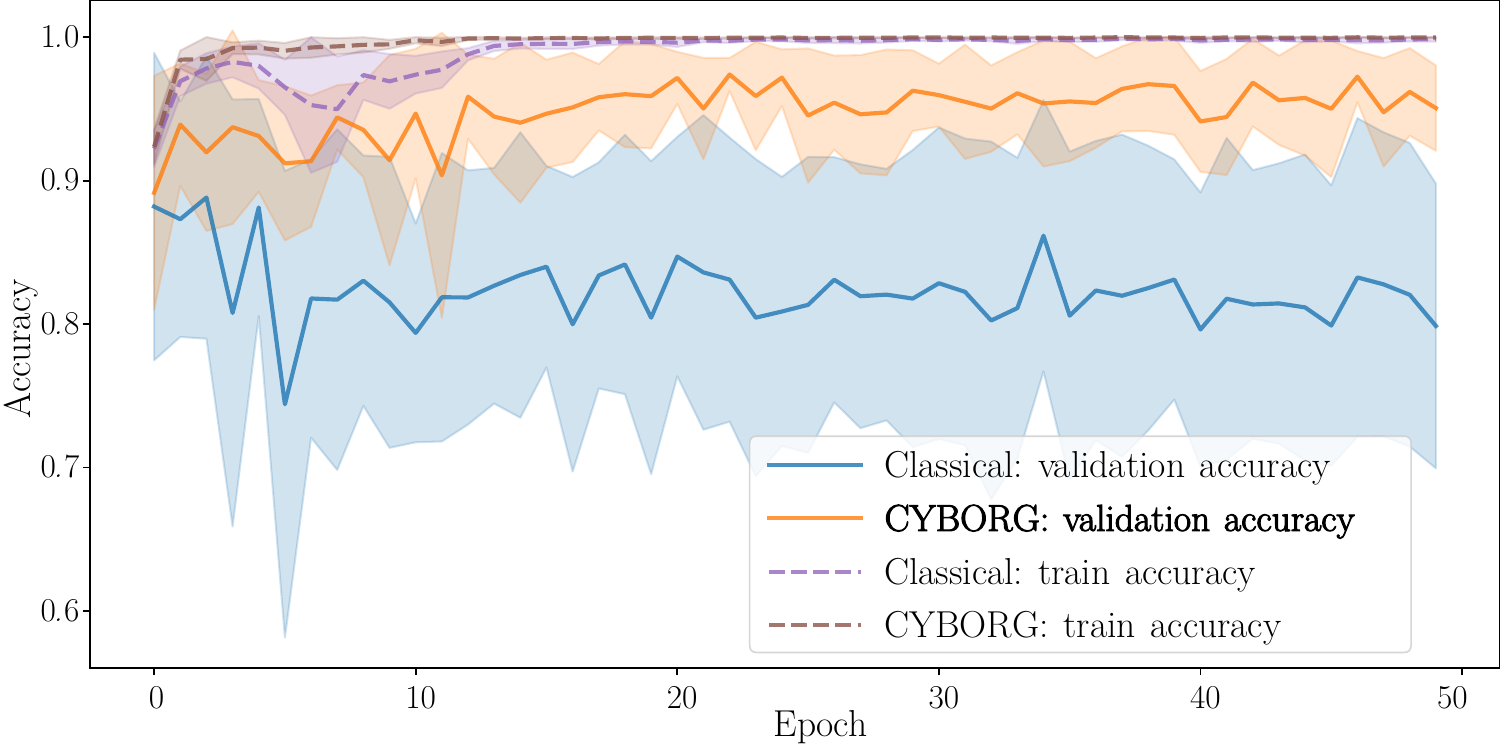}
    \vskip-1mm
    \caption{
    Comparison of training and validation accuracy for ResNet50 with only classification accuracy loss versus with CYBORG loss.
    Training accuracy quickly approaches 100\% for both.
    But CYBORG-trained models achieve significantly higher validation accuracy throughout, indicating more effective learning.
    Shaded area represents $\pm1$ standard deviation of the accuracy by epoch.
    }
    \label{fig:resnet_training}
    \null\vskip-5mm
\end{figure}

\vskip1mm\noindent{\bf Scenario 1 vs Scenario 3 (\ie Classical vs CYBORG Training).} As shown in Fig. \ref{fig:roc}, the models trained just using the \textit{original data} do not generalize well to the test sets. 
In contrast, when CYBORG training is applied on the same data, accuracy on the test sets increases significantly.
Fig. \ref{fig:resnet_training} outlines the training and validation accuracies for both Scenario 1 (only classification loss) and Scenario 3 (with CYBORG loss) for training of the ResNet50 models. As it can be seen, training accuracy quickly reaches 100\%, meaning both sets learn representative features of the training samples. However, the {\bf CYBORG-trained model shows better validation accuracy across all epochs.} The decrease in validation accuracy for Scenario 1 models suggests overfitting, and the subsequent plateau (and even slight decline) can be explained by the training accuracy reaching 100\% resulting in minimal optimization. The supplementary materials include plots for DenseNet, Inception-v3 and Xception models, showing very similar trends.

\vskip1mm\noindent{\bf Scenario 2 (\ie Classical Training with Large Data).} Experiments were conducted to determine whether simply adding more data from the same sources as the original data to the Scenario 1 approach would bridge the performance gap. 
Giving the classical training process additional data, up to $7\times$ the original amount, does not enable it to achieve CYBORG-level accuracy. In some cases, the performance of models trained on larger sets is even inferior to models trained with less data and CYBORG. The classical training simply overfits to the training data and so cannot generalize to samples generated by unknown GAN architecture. The CYBORG-trained models generalize better.

\vskip1mm\noindent{\bf Evaluating An Off-The-Shelf Deepfake Detector on Test Data.} The en\-sem\-ble-based ``deep fake'' detection methods \cite{bonettini2021video} demonstrated very high performance on the DFDC and FF++ test data with AUCs of 0.957 and 0.920, respectively. That means we were able to replicate the original results without any issues.
However, when applied to the task of synthetic image detection, these top-performing ``deep fake'' ensemble methods are incapable of distinguishing between authentic and synthetically-generated images, as demonstrated by AUCs of less than 0.5 (0.385 and 0.373) for these methods in Fig. \ref{fig:roc}(e).

\begin{figure}[ht]
\centering
    \includegraphics[width=0.7\columnwidth]{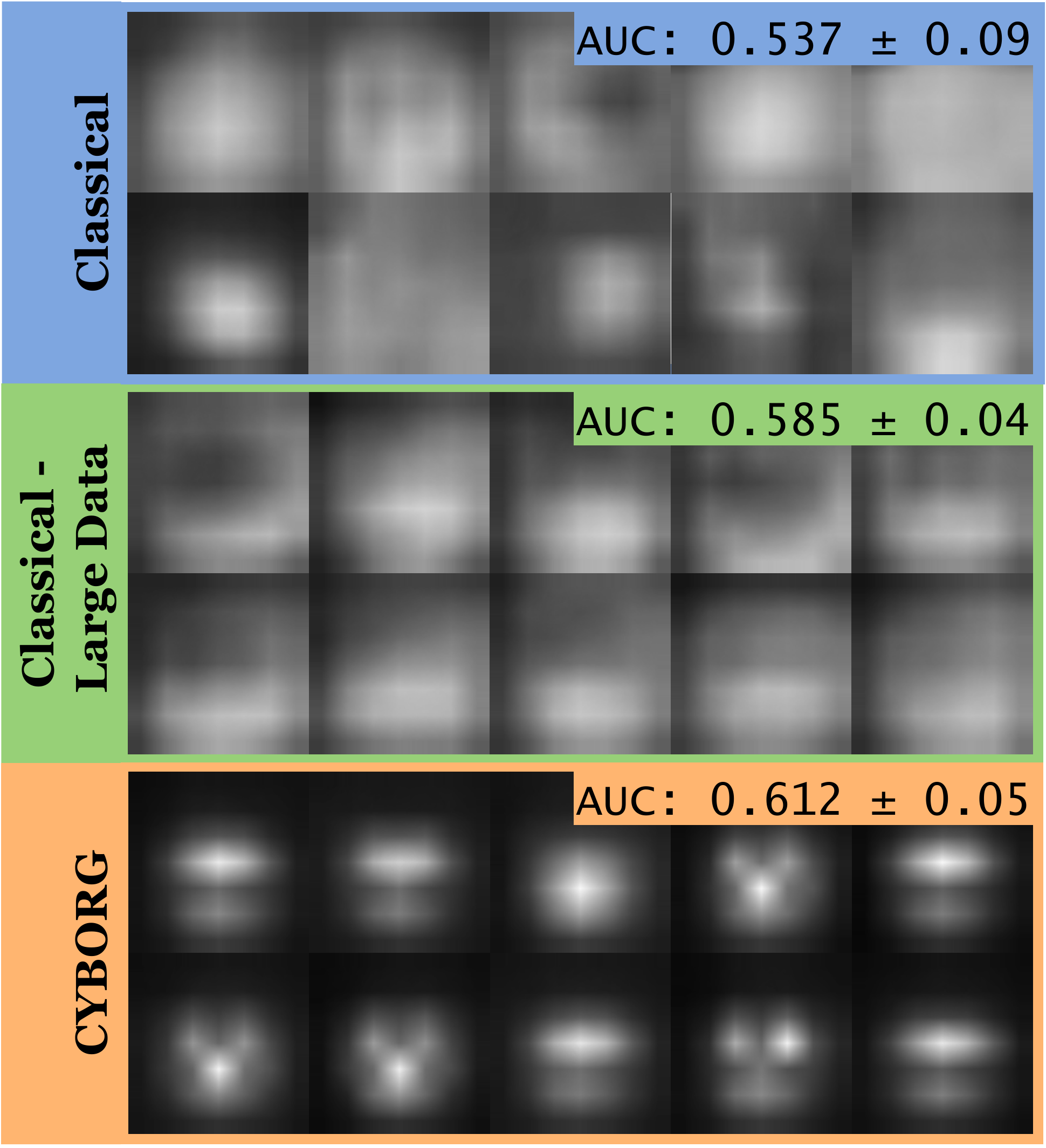}
  \caption{Average CAMs across the entire test set for 10 independently trained ResNet50 models in three experimental settings. Each individual plot is the average CAM obtained for all test images for a given model. (Similar results can be observed for other CNN architectures, included into the supp. materials). Compared to the average human annotation, shown in Fig. \ref{fig:incorp_process}(d), it is clear CYBORG models are guided effectively by human annotation.}
  \label{fig:visualizations}
  \null\vskip-5mm
\end{figure}

\vskip1mm\noindent{\bf What The CYBORG-trained Models ``Look'' At?} The results presented so far suggest that the CYBORG approach does guide deep learning towards models generalizing better on samples generated by never-seen-before GANs. However, are these CYBORG-trained models visually exhibiting behaviour akin to human annotators? To answer this question, visualizations of model saliency are generated on the test set, and illustrated in Fig. \ref{fig:visualizations}. For experimental Scenarios 1-3, a plot is created for each of the 10 independently trained models. To create each of these individual plots, the CAM is generated using the same mechanism as the model saliency probing during training~\cite{zhou2016learning}, but for each sample in the test set. The average of all 700,000 CAMs (100k authentic, 600k synthetic) is calculated. This details where the model deems important for classification on average over the entire test set for both classes combined. Because all images are aligned, facial features present in similar locations across test samples. 

For both DenseNet and ResNet, the difference between Scenario 1 (``Classical'') and Scenario 3 (``CYBORG'') is immediately evident. Models trained with CYBORG exhibit CAMs that resemble facial features such as the mouth, nose and eyes. The models trained with classification loss alone show less compact CAMs, meaning there is no consensus of importance across the test images. 

While the dominating features are comparable for Scenario 1 and Scenario 3 in the Inception-v3 experiment, the CYBORG models are more precisely focused on the facial region. For Xception, both the Scenario 1 and Scenario 3 models present similar CAMs, which is also indicated by the performance. However, CYBORG models exhibit more certainty as indicated by higher compactness of the corresponding CAMs. Tuning of the $\alpha$ value may be required for this model to attain average CAMs similar to ResNet.

For the Scenario 2 (``Classical -- Large Data''), roughly similar CAMs are observed across all four architectures. For DenseNet and ResNet, this results in greater performance than classification alone. In these two cases, the Scenario 2 models are more concise than the Scenario 1 models. For Inception-v3 and Xception, Scenario 2 results in poorer performance than Scenario 1. These models loosely focused on similar features as in Scenarios 1 and 3. However, there is less consensus and more uncertainty indicating increased overfitting to the training data. This added uncertainty explains the degradation in performance.

By direct comparison to the average correct human annotation in Fig. \ref{fig:visualizations}(e), the models trained with CYBORG focus on features that are more similar to human detailed salient regions than models trained with classical cross-entropy loss in all cases. Thus, it can be concluded these models are effectively guided by the human annotations supplied during the training process.

\vskip1mm\noindent{\bf Assessing Value Of Human Saliency.} As demonstrated in Fig. \ref{fig:roc}, 
using human saliency maps results in a larger increase in performance than automatically determined segmentation masks. 
Thus, to answer the two questions posed in Sec. \ref{sec:face_segmentation}, deep learning-based segmentation masks can effectively guide the models using CYBORG loss, however, human saliency maps guide the network to {\bf more generalized features}, thus achieving a {\bf greater performance in open-set classification problem}. 

\vskip1mm\noindent{\bf Incorporation Of CYBORG Into An Existing Synthetic Face Detector.} 
To determine whether the incorporation of CYBORG loss would improve upon existing methods, the CYBORG loss was added to Wang \etal's ~\cite{wang2019cnngenerated} publicly available, re-trainable synthetic face detection model~\cite{wang2021repo}. 
This resulted in a performance increase from AUC=$0.554\pm0.03$ in the classical scenario to AUC=$0.591\pm0.03$. 

\vskip1mm\noindent{\bf Incorporation Of Human Saliency Into An Attention Mechanism.} 
A popular approach to force networks to focus on specified regions is self-attention. 
As an additional experiment, we investigated whether replacement of the attention masks proposed in~\cite{dang2020detection} with human saliency results in higher accuracy. We train two models: (1) using the original approach with no human saliency, and (2) using our human saliency maps as the attention masks for authentic and synthetic images. In both cases, the parameters proposed by the authors are used. 
We found that replacement of the ground truth masks with human saliency increased performance from AUC=$0.428\pm0.04$ to AUC=$0.498\pm0.06$, suggesting that implanting human perception into the self-attention module narrows the model's search for areas of importance (even in the absence of ground truth) and boosts performance.

\section{Conclusions}
\label{sec:conclusion}

We proposed the CYBORG approach to CNN training, in which the learning is guided by information distilled from human visual abilities.
CYBORG uses a human perception-based loss term for the disagreement between the CNN’s class activation map and a human-derived saliency map. 
To emphasize that CYBORG is independent of CNN backbone, results are shown for four different models: ResNet, DenseNet, Inception and Xception.
Applying CYBORG improved performance in detecting synthetic face images from six different GANs unseen in training (Fig.~\ref{fig:roc}).
CYBORG-trained models produced CAMs closer to human-annotated regions of saliency than classically trained models (Fig.~\ref{fig:visualizations}). 
Comparing the training and validation accuracy of classical versus CYBORG training
(Fig.~\ref{fig:resnet_training}) makes it clear that CYBORG results in a model that generalizes better to samples generated by never-seen-before GANs.
Evaluation of a state-of-the-art ``deep fake'' detection model on our test set shows that this task and synthetic image detection are different domains.

Application of the CYBORG approach is possible for tasks in which human accuracy is not at the ``expert level''. The human saliency maps used in this work came from a task in which median human accuracy was 69.2\% (Fig.~\ref{fig:annot_results}), and only saliency maps associated with correct human decisions were applied.

Finally, we have shown that models trained classicaly with 7 times more training data does not achieve the performance of CYBORG-trained models, and that replacing human perception-driven maps with automatic face segmentation masks ends up with performance inferior to CYBORG. This shows value of the proposed mechanism to apply human perception to use limited data effectively in training.


{\small
\bibliographystyle{ieee_fullname}
\bibliography{main}

\begin{thebibliography}{10}\itemsep=-1pt

\bibitem{Albiero_CVPR_2021}
Vítor Albiero, Xingyu Chen, Xi Yin, Guan Pang, and Tal Hassner.
\newblock {img2pose: Face Alignment and Detection via 6DoF, Face Pose
  Estimation}.
\newblock In {\em IEEE/CVF Conf. Comput. Vis. Pattern Recog. (CVPR)}, pages
  7613--7623, 2021.

\bibitem{Banerjee_IJCB_2017}
Sandipan Banerjee, John~S. Bernhard, Walter~J. Scheirer, Kevin~W. Bowyer, and
  Patrick~J. Flynn.
\newblock Srefi: Synthesis of realistic example face images.
\newblock In {\em IEEE Int. Joint Conf.on Biometrics (IJCB)}, pages 37--45,
  2017.

\bibitem{bonettini2021video}
Nicolo Bonettini, Edoardo~Daniele Cannas, Sara Mandelli, Luca Bondi, Paolo
  Bestagini, and Stefano Tubaro.
\newblock Video face manipulation detection through ensemble of cnns.
\newblock In {\em 2020 25th International Conference on Pattern Recognition
  (ICPR)}, pages 5012--5019. IEEE, 2021.

\bibitem{botha2020fake}
Johnny Botha and Heloise Pieterse.
\newblock Fake news and deepfakes: A dangerous threat for 21st century
  information security.
\newblock In {\em Int. Conf. on Cyber Warfare and Security (ICCWS)}, page~57.
  Academic Conferences and Publishing Limited, 2020.

\bibitem{boyd2021human}
Aidan Boyd, Kevin Bowyer, and Adam Czajka.
\newblock {Human-Aided Saliency Maps Improve Generalization of Deep Learning}.
\newblock {\em arXiv preprint arXiv:2105.03492}, 2021.

\bibitem{brock2018large}
Andrew Brock, Jeff Donahue, and Karen Simonyan.
\newblock {Large scale GAN training for high fidelity natural image synthesis}.
\newblock {\em arXiv preprint arXiv:1809.11096}, 2018.

\bibitem{chesney2019deepfakes}
Robert Chesney and Danielle Citron.
\newblock Deepfakes and the new disinformation war: The coming age of
  post-truth geopolitics.
\newblock {\em Foreign Aff.}, 98:147, 2019.

\bibitem{choi2020stargan}
Yunjey Choi, Youngjung Uh, Jaejun Yoo, and Jung-Woo Ha.
\newblock {StarGAN v2: Diverse image synthesis for multiple domains}.
\newblock In {\em IEEE/CVF Conf. Comput. Vis. Pattern Recog. (CVPR)}, pages
  8185--8194, 2020.

\bibitem{chollet2017xception}
Fran{\c{c}}ois Chollet.
\newblock {Xception: Deep learning with depthwise separable convolutions}.
\newblock In {\em IEEE/CVF Conf. Comput. Vis. Pattern Recog. (CVPR)}, pages
  1251--1258, 2017.

\bibitem{cozzolino2014image}
Davide Cozzolino, Diego Gragnaniello, and Luisa Verdoliva.
\newblock Image forgery detection through residual-based local descriptors and
  block-matching.
\newblock In {\em IEEE Int. Conf. Image Process. (ICIP)}, pages 5297--5301.
  IEEE, 2014.

\bibitem{czajka2019domain}
Adam Czajka, Daniel Moreira, Kevin Bowyer, and Patrick Flynn.
\newblock Domain-specific human-inspired binarized statistical image features
  for iris recognition.
\newblock In {\em IEEE/CVF Winter Conf. on App. of Comp. Vis. (WACV)}, pages
  959--967. IEEE, 2019.

\bibitem{dang2020detection}
Hao Dang, Feng Liu, Joel Stehouwer, Xiaoming Liu, and Anil~K Jain.
\newblock On the detection of digital face manipulation.
\newblock In {\em IEEE/CVF Conf. Comput. Vis. Pattern Recog. (CVPR)}, pages
  5781--5790, 2020.

\bibitem{deepfakes2021}
DeepFakes.
\newblock {Faceswap}.
\newblock \url{https://github.com/deepfakes/faceswap}, 2021.

\bibitem{DFDC2020}
Brian Dolhansky, Joanna Bitton, Ben Pflaum, Jikuo Lu, Russ Howes, Menglin Wang,
  and Cristian~Canton Ferrer.
\newblock The deepfake detection challenge (dfdc) dataset.
\newblock {\em arXiv preprint arXiv:2006.07397}, 2020.

\bibitem{frank2020leveraging}
Joel Frank, Thorsten Eisenhofer, Lea Sch{\"o}nherr, Asja Fischer, Dorothea
  Kolossa, and Thorsten Holz.
\newblock Leveraging frequency analysis for deep fake image recognition.
\newblock In {\em Int. Conf. on Machine Learning (ICML)}, pages 3247--3258.
  PMLR, 2020.

\bibitem{goodfellow2014generative}
Ian Goodfellow, Jean Pouget-Abadie, Mehdi Mirza, Bing Xu, David Warde-Farley,
  Sherjil Ozair, Aaron Courville, and Yoshua Bengio.
\newblock Generative adversarial nets.
\newblock {\em Advances in Neural Information Processing Systems}, 27, 2014.

\bibitem{grip-unina}
Diego Gragnaniello, Davide Cozzolino, Francesco Marra, Giovanni Poggi, and
  Luisa Verdoliva.
\newblock {GANimageDetection}.
\newblock \url{https://github.com/grip-unina/GANimageDetection}, 2021.

\bibitem{grieggs2021measuring}
Samuel Grieggs, Bingyu Shen, Greta Rauch, Pei Li, Jiaqi Ma, David Chiang, Brian
  Price, and Walter Scheirer.
\newblock Measuring human perception to improve handwritten document
  transcription.
\newblock {\em IEEE Trans. Pattern Anal. Mach. Intell.}, 2021.

\bibitem{he2016deep}
Kaiming He, Xiangyu Zhang, Shaoqing Ren, and Jian Sun.
\newblock Deep residual learning for image recognition.
\newblock In {\em IEEE/CVF Conf. Comput. Vis. Pattern Recog. (CVPR)}, pages
  770--778, 2016.

\bibitem{he2019human}
Sen He, Hamed~R Tavakoli, Ali Borji, and Nicolas Pugeault.
\newblock Human attention in image captioning: Dataset and analysis.
\newblock In {\em IEEE/CVF Int. Conf. Comput. Vis. (ICCV)}, pages 8529--8538,
  2019.

\bibitem{hernandez2020biometric}
Javier Hernandez-Ortega, Javier Galbally, Julian Fierrez, and Laurent Beslay.
\newblock {Biometric Quality: Review and Application to Face Recognition with
  FaceQnet}.
\newblock {\em arXiv preprint arXiv:2006.03298}, 2020.

\bibitem{huang2017densely}
Gao Huang, Zhuang Liu, Laurens Van Der~Maaten, and Kilian~Q Weinberger.
\newblock Densely connected convolutional networks.
\newblock In {\em IEEE/CVF Conf. Comput. Vis. Pattern Recog. (CVPR)}, pages
  4700--4708, 2017.

\bibitem{huang2021specific}
Yupan Huang, Zhaoyang Zeng, and Yutong Lu.
\newblock {Be Specific, Be Clear: Bridging Machine and Human Captions by
  Scene-Guided Transformer}.
\newblock In {\em Workshop on Multi-Modal Pre-Training for Multimedia
  Understanding}, pages 4--13, 2021.

\bibitem{littlejuyan}
Yan Ju.
\newblock {GAN-generated-image-detector}.
\newblock \url{https://gitlab.com/littlejuyan/GAN-generated-image-detector},
  2021.

\bibitem{karras2017progressive}
Tero Karras, Timo Aila, Samuli Laine, and Jaakko Lehtinen.
\newblock {Progressive Growing of GANs for Improved Quality, Stability, and
  Variation}.
\newblock {\em arXiv preprint arXiv:1710.10196}, 2017.

\bibitem{karras2017repo}
Tero Karras, Timo Aila, Samuli Laine, and Jaakko Lehtinen.
\newblock {Progressive Growing of GANs for Improved Quality, Stability, and
  Variation: Official TensorFlow Implementation}.
\newblock \url{https://github.com/tkarras/progressive\_growing\_of\_gans},
  2021.

\bibitem{Karras2020ada}
Tero Karras, Miika Aittala, Janne Hellsten, Samuli Laine, Jaakko Lehtinen, and
  Timo Aila.
\newblock {Training Generative Adversarial Networks with Limited Data}.
\newblock In {\em Adv. Neural Inform. Process. Syst. (NeurIPS)}, 2020.

\bibitem{Karras2021}
Tero Karras, Miika Aittala, Samuli Laine, Erik H\"ark\"onen, Janne Hellsten,
  Jaakko Lehtinen, and Timo Aila.
\newblock {Alias-Free Generative Adversarial Networks}.
\newblock In {\em Adv. Neural Inform. Process. Syst. (NeurIPS)}, 2021.

\bibitem{karras2019style}
Tero Karras, Samuli Laine, and Timo Aila.
\newblock {A Style-Based Generator Architecture for Generative Adversarial
  Networks}.
\newblock In {\em IEEE/CVF Conf. Comput. Vis. Pattern Recog. (CVPR)}, pages
  4401--4410, 2019.

\bibitem{karras2020analyzing}
Tero Karras, Samuli Laine, Miika Aittala, Janne Hellsten, Jaakko Lehtinen, and
  Timo Aila.
\newblock {Analyzing and Improving the Image Quality of StyleGAN}.
\newblock In {\em IEEE/CVF Conf. Comput. Vis. Pattern Recog. (CVPR)}, pages
  8110--8119, 2020.

\bibitem{linsley2018learning}
Drew Linsley, Dan Shiebler, Sven Eberhardt, and Thomas Serre.
\newblock Learning what and where to attend.
\newblock {\em arXiv preprint arXiv:1805.08819}, 2018.

\bibitem{liu2015faceattributes}
Ziwei Liu, Ping Luo, Xiaogang Wang, and Xiaoou Tang.
\newblock {Deep Learning Face Attributes in the Wild}.
\newblock In {\em IEEE/CVF Int. Conf. Comput. Vis. (ICCV)}, December 2015.

\bibitem{polimi-ispl}
Sara Mandelli, Nicol{\'o} Bonettini, Paolo Bestagini, and Stefano Tubaro.
\newblock Training cnns in presence of jpeg compression: Multimedia forensics
  vs computer vision.
\newblock In {\em IEEE Int. Workshop on Inf. Forensics and Sec. (WIFS)}, pages
  1--6, 2020.

\bibitem{marra2018detection}
Francesco Marra, Diego Gragnaniello, Davide Cozzolino, and Luisa Verdoliva.
\newblock {Detection of GAN-generated fake images over social networks}.
\newblock In {\em IEEE Conf. on Multimedia Inf. Proc. and Retrieval (MIPR)},
  pages 384--389. IEEE, 2018.

\bibitem{moreira2019performance}
Daniel Moreira, Mateusz Trokielewicz, Adam Czajka, Kevin Bowyer, and Patrick
  Flynn.
\newblock {Performance of Humans in Iris Recognition: The Impact of Iris
  Condition and Annotation-driven Verification}.
\newblock In {\em IEEE/CVF Winter Conf. on App. of Comp. Vis. (WACV)}, pages
  941--949. IEEE, 2019.

\bibitem{Muhammad_IJCNN_2020}
Mohammed~Bany Muhammad and Mohammed Yeasin.
\newblock {Eigen-CAM: Class Activation Map using Principal Components}.
\newblock In {\em IEEE Int. Joint Conf. on Neural Networks (IJCNN)}. IEEE, Jul
  2020.

\bibitem{o2012comparing}
Alice~J O'Toole, Xaiobo An, Joseph Dunlop, Vaidehi Natu, and P~Jonathon
  Phillips.
\newblock Comparing face recognition algorithms to humans on challenging tasks.
\newblock {\em ACM Trans. on Applied Perception}, 9(4):1--13, 2012.

\bibitem{park2019gaugan}
Taesung Park, Ming-Yu Liu, Ting-Chun Wang, and Jun-Yan Zhu.
\newblock {GauGAN: semantic image synthesis with spatially adaptive
  normalization}.
\newblock In {\em ACM SIGGRAPH Real-Time Live!}, pages 1--1. IEEE/CVF Conf.
  Comput. Vis. Pattern Recog. (CVPR), 2019.

\bibitem{Phillips_IVC_2017}
P.~Jonathon Phillips, Patrick~J. Flynn, and Kevin~W. Bowyer.
\newblock Lessons from collecting a million biometric samples.
\newblock {\em Image and Vision Computing}, 58:96--107, 2017.

\bibitem{pytorchModelZoo}
PyTorch.
\newblock {Pytorch Model Zoo}.
\newblock \url{https://pytorch.org/serve/model\_zoo.html}, 2021.

\bibitem{qian2020thinking}
Yuyang Qian, Guojun Yin, Lu Sheng, Zixuan Chen, and Jing Shao.
\newblock Thinking in frequency: Face forgery detection by mining
  frequency-aware clues.
\newblock In {\em IEEE Eur. Conf. Comput. Vis. (ECCV)}, pages 86--103.
  Springer, 2020.

\bibitem{richardwebster2018visual}
Brandon RichardWebster, So~Yon Kwon, Christopher Clarizio, Samuel~E Anthony,
  and Walter~J Scheirer.
\newblock Visual psychophysics for making face recognition algorithms more
  explainable.
\newblock In {\em IEEE Eur. Conf. Comput. Vis. (ECCV)}, pages 252--270, 2018.

\bibitem{roessler2019faceforensicspp}
Andreas R\"ossler, Davide Cozzolino, Luisa Verdoliva, Christian Riess, Justus
  Thies, and Matthias Nie{\ss}ner.
\newblock Face{F}orensics++: Learning to detect manipulated facial images.
\newblock In {\em International Conference on Computer Vision (ICCV)}, 2019.

\bibitem{Selvaraju_ICCV_2017}
Ramprasaath~R. Selvaraju, Michael Cogswell, Abhishek Das, Ramakrishna Vedantam,
  Devi Parikh, and Dhruv Batra.
\newblock {Grad-CAM: Visual Explanations from Deep Networks via Gradient-Based
  Localization}.
\newblock In {\em IEEE/CVF Int. Conf. Comput. Vis. (ICCV)}, pages 618--626,
  2017.

\bibitem{shen2021study}
Bingyu Shen, Brandon RichardWebster, Alice O'Toole, Kevin Bowyer, and Walter~J.
  Scheirer.
\newblock A study of the human perception of synthetic faces.
\newblock {\em arXiv preprint arXiv:2111.04230}, 2021.

\bibitem{kitware}
Harry Sun.
\newblock {Kitware Generated Image Detector}.
\newblock \url{https://github.com/Kitware/generated-image-detection}, 2021.

\bibitem{szegedy2016rethinking}
Christian Szegedy, Vincent Vanhoucke, Sergey Ioffe, Jon Shlens, and Zbigniew
  Wojna.
\newblock Rethinking the inception architecture for computer vision.
\newblock In {\em IEEE/CVF Conf. Comput. Vis. Pattern Recog. (CVPR)}, pages
  2818--2826, 2016.

\bibitem{tariq2019gan}
Shahroz Tariq, Sangyup Lee, Hoyoung Kim, Youjin Shin, and Simon~S Woo.
\newblock Gan is a friend or foe? a framework to detect various fake face
  images.
\newblock In {\em ACM/SIGAPP Symp. on Applied Comp.}, pages 1296--1303, 2019.

\bibitem{trokielewicz2019perception}
Mateusz Trokielewicz, Adam Czajka, and Piotr Maciejewicz.
\newblock Perception of image features in post-mortem iris recognition: Humans
  vs machines.
\newblock In {\em IEEE Int. Conf. on Biometrics Theory, Applications and
  Systems (BTAS)}, pages 1--8. IEEE, 2019.

\bibitem{wang2019cnngenerated}
Sheng-Yu Wang, Oliver Wang, Richard Zhang, Andrew Owens, and Alexei~A Efros.
\newblock Cnn-generated images are surprisingly easy to spot ... for now.
\newblock In {\em IEEE/CVF Conf. Comput. Vis. Pattern Recog. (CVPR)}, 2020.

\bibitem{wang2021repo}
Sheng-Yu Wang, Oliver Wang, Richard Zhang, Andrew Owens, and Alexei~A Efros.
\newblock {CNNDetection}.
\newblock \url{https://github.com/peterwang512/CNNDetection}, 2020.

\bibitem{yaseryacoob}
Yaser Yacoob.
\newblock {GAN-Scanner}.
\newblock \url{https://github.com/yaseryacoob/GAN-Scanner}, 2021.

\bibitem{zhang2020human}
Ruohan Zhang, Akanksha Saran, Bo Liu, Yifeng Zhu, Sihang Guo, Scott Niekum,
  Dana Ballard, and Mary Hayhoe.
\newblock Human gaze assisted artificial intelligence: a review.
\newblock In {\em Int. Joint Conf. on Art. Intell. (IJCAI)}, volume 2020, page
  4951. NIH Public Access, 2020.

\bibitem{zhou2016learning}
Bolei Zhou, Aditya Khosla, Agata Lapedriza, Aude Oliva, and Antonio Torralba.
\newblock Learning deep features for discriminative localization.
\newblock In {\em IEEE/CVF Conf. Comput. Vis. Pattern Recog. (CVPR)}, pages
  2921--2929, 2016.

\bibitem{zhu2017unpaired}
Jun-Yan Zhu, Taesung Park, Phillip Isola, and Alexei~A Efros.
\newblock Unpaired image-to-image translation using cycle-consistent
  adversarial networks.
\newblock In {\em IEEE/CVF Int. Conf. Comput. Vis. (ICCV)}, 2017.

\bibitem{bisenet2019}
zll.
\newblock {BisSeNet -- Face Parsing Tool}.
\newblock \url{https://github.com/zllrunning/face-parsing.PyTorch}, 2019.

\end{thebibliography}
}

\appendix


\section{Online Annotation Tool}

\begin{figure*}[!htb]
\centering
\includegraphics[width=\textwidth]{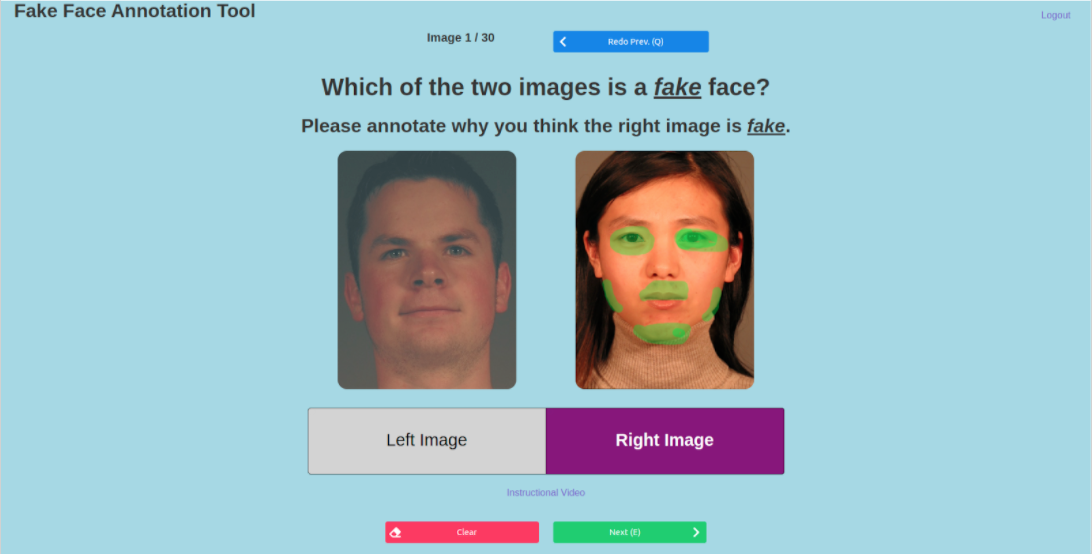}
\caption{A screenshot from the online annotation tool designed for this work and used to collect human annotation data.}
\label{fig:annot_tool}
\end{figure*}

Fig. \ref{fig:annot_tool} shows a screenshot from the tool used to collect the human annotation data, as described in Section 4 of the main paper. An example pair of face images is shown, where the right image is fake. The subject classifying this particular pair of images answered the prompt question correctly by picking the right image as being fake. Green highlights correspond to ``salient'' regions annotated by the user as to which part(s) of the image led to their classification decision. 

\section{Datasets}

We will release all data collected for this work with the camera-ready submission, along with a standard data sharing license agreement (human annotations were collected under IRB protocol). The data shared with the camera-ready paper will allow for replicating all experiments presented in the paper. Some test sets used in this work are already in the public domain, and their copies can be obtained by following the references provided in the main paper. The data can be used to run the testing codes after performing image pre-processing as described in the main paper.

\section{Source Codes}

Example source codes demonstrating how CYBORG loss was implemented are available in \textit{\textbf{code.zip}}. This includes the training and testing code used to compute the results in the paper, along with one example DenseNet121 model that was trained with CYBORG loss. The full suite of pretrained models used for the generation of the paper results will be released with the camera-ready submission. The Xception network code can be downloaded from \url{https://github.com/ondyari/FaceForensics/tree/master/classification}; the other three architectures can be used natively with PyTorch.

\section{Deep Learning-Based Face Segmentation}

\begin{figure*}[t]
\centering
  \includegraphics[width=0.7\textwidth]{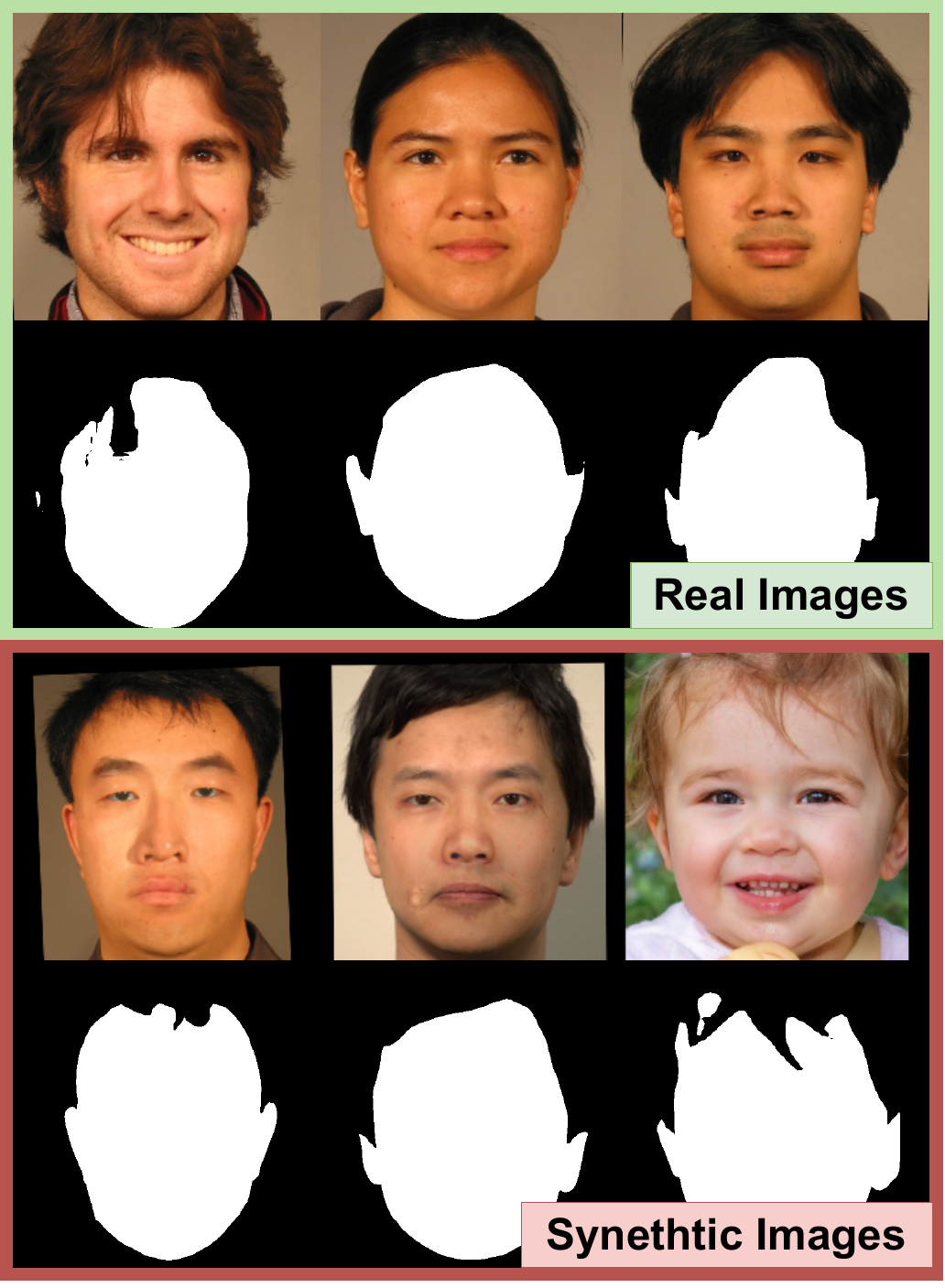}

  \caption{Three examples of real images and corresponding deep learning-based segmentations (top two rows) and three examples of synthetic images and corresponding deep learning-based segmentations (bottom two rows). Bottom two rows: image 1 and 2 from SREFI dataset, image 3 generated by StyleGAN2.}
  \label{fig:face_segmentations}
\end{figure*}

Examples of the output from the deep learning-based face segmentation tool \cite{bisenet2019} can be found in Fig. \ref{fig:face_segmentations}. The top row shows three examples of real (genuine) face images as well as their corresponding face segmentations. The bottom rows show three examples of synthetic (fake) face images as well as their corresponding face segmentations. The first two synthetic images shown are from the SREFI dataset~\cite{Banerjee_IJCB_2017}) and the third image is generated by StyleGAN2~\cite{karras2020analyzing}.

\section{Supplementary Experimental Results \\(Sec. 7 in the main paper)}

\subsection{Tabulated AUCs, Training/Validation Plots, and ROC Curves}

Tab. \ref{tab:we_are_simply_the_best} outlines the individual performances of each of the studied approaches on the individual GAN sources; this supplements the plots for all GANs combined, as presented in the main paper.

\begin{figure*}[!htb] 
     \centering
     \includegraphics[width=0.8\textwidth]{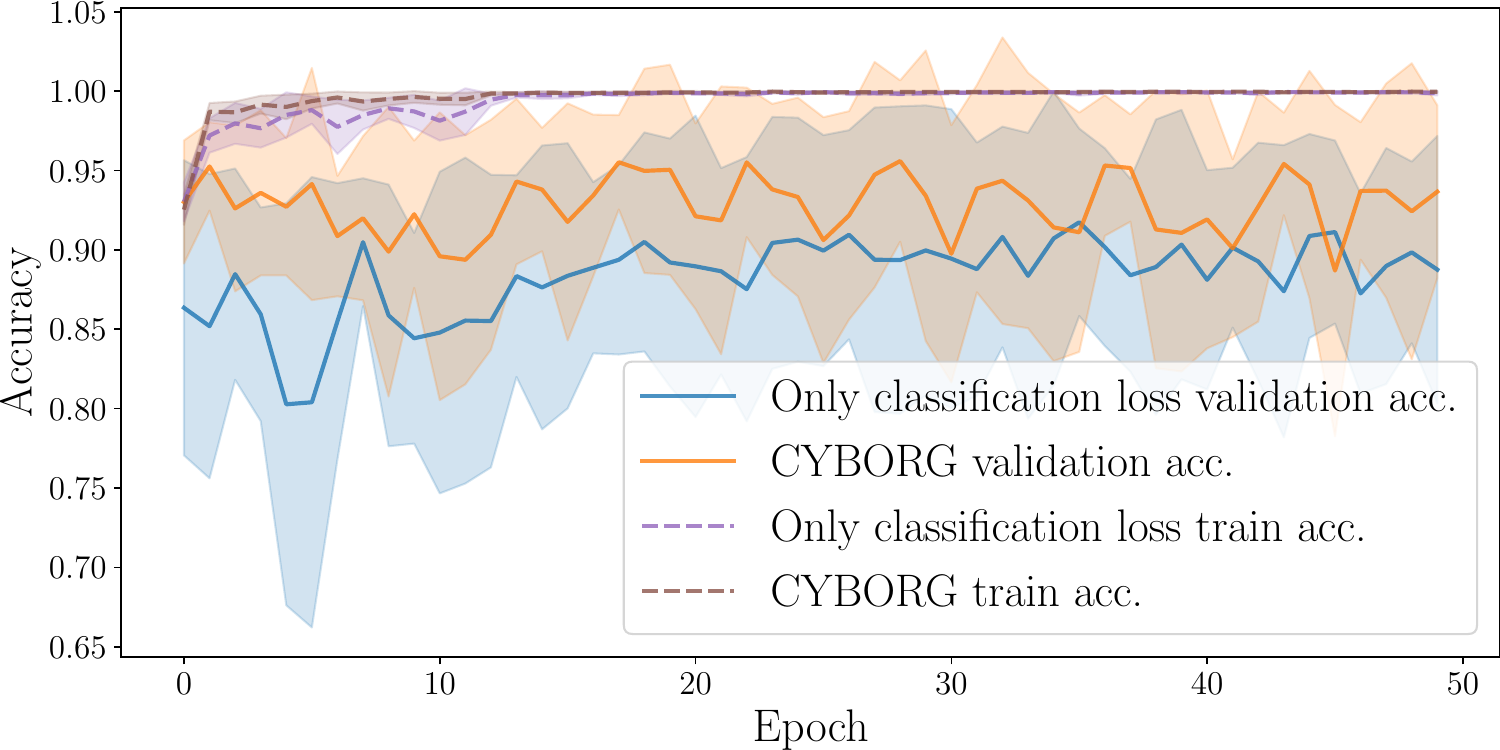}\vskip-1mm
     \caption{Training / validation accuracy: {\bf DenseNet121}. ``Only classification loss'' corresponds to Scenario 1 (``Classical Training''), and ``CYBORG'' corresponds to Scenario 3, as defined in Sec. 6.1 in the main paper.}
     \label{fig:densenet_training}
     \null\vskip-5mm
\end{figure*}

\begin{figure*}[!htb] 
     \centering
     \includegraphics[width=0.8\textwidth]{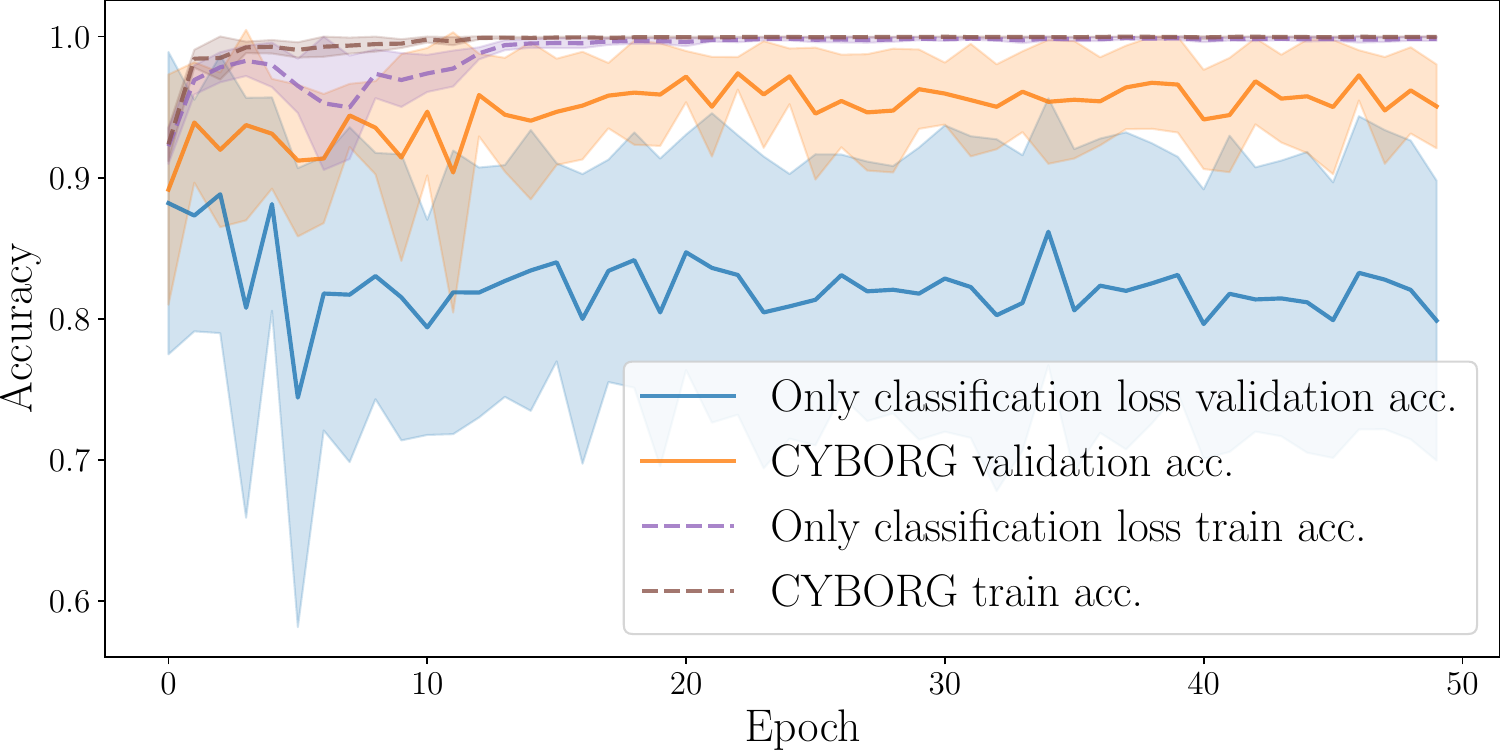}\vskip-1mm
     \caption{Same as in Fig. \ref{fig:densenet_training}, but for {\bf ResNet50}}
     \label{fig:resnet_training_supp}
     \null\vskip-5mm
\end{figure*}

\begin{figure*}[!htb] 
     \centering
     \includegraphics[width=0.8\textwidth]{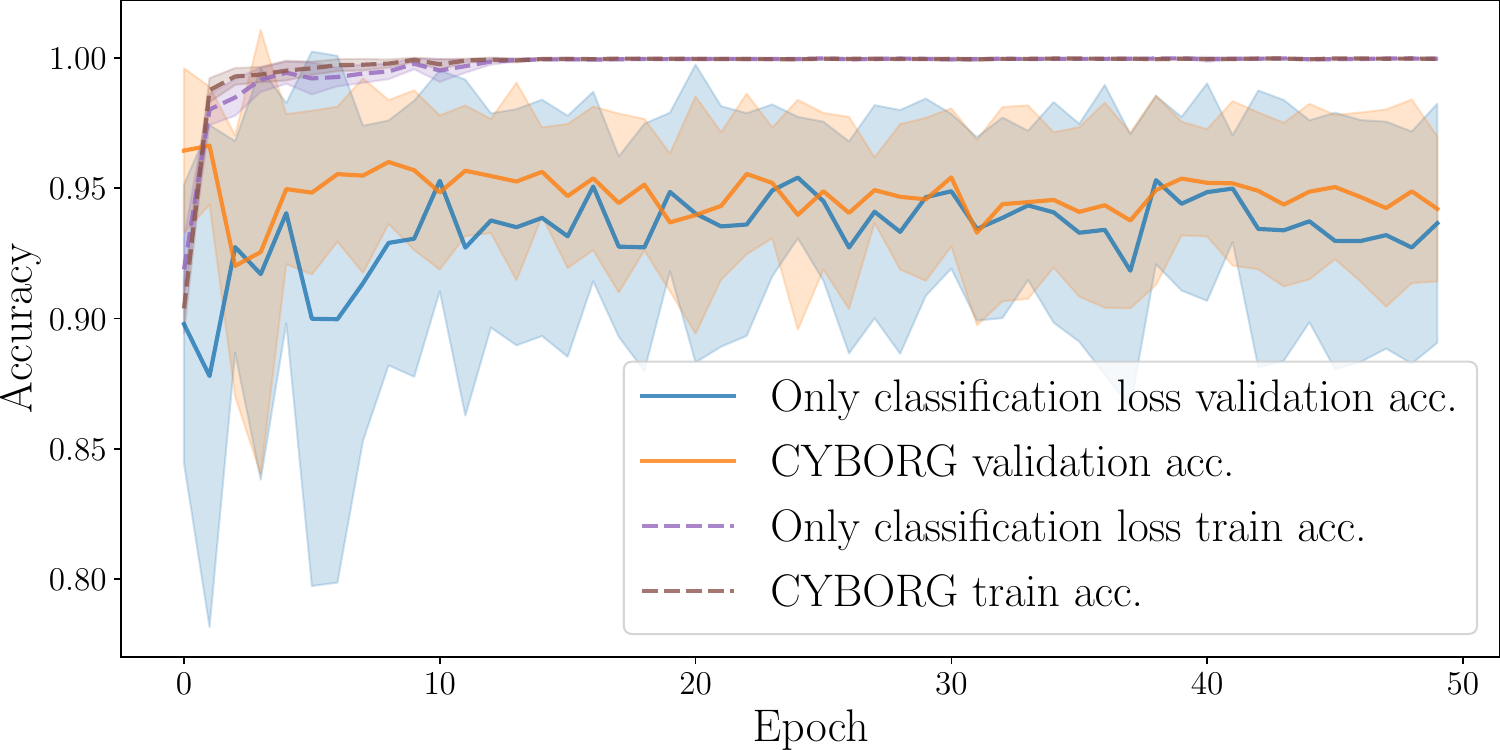}\vskip-1mm
     \caption{Same as in Fig. \ref{fig:densenet_training}, but for {\bf Inception-v3}}
     \label{fig:inception_training}
     \null\vskip-5mm
\end{figure*}

\begin{figure*}[!htb] 
     \centering
     \includegraphics[width=0.8\textwidth]{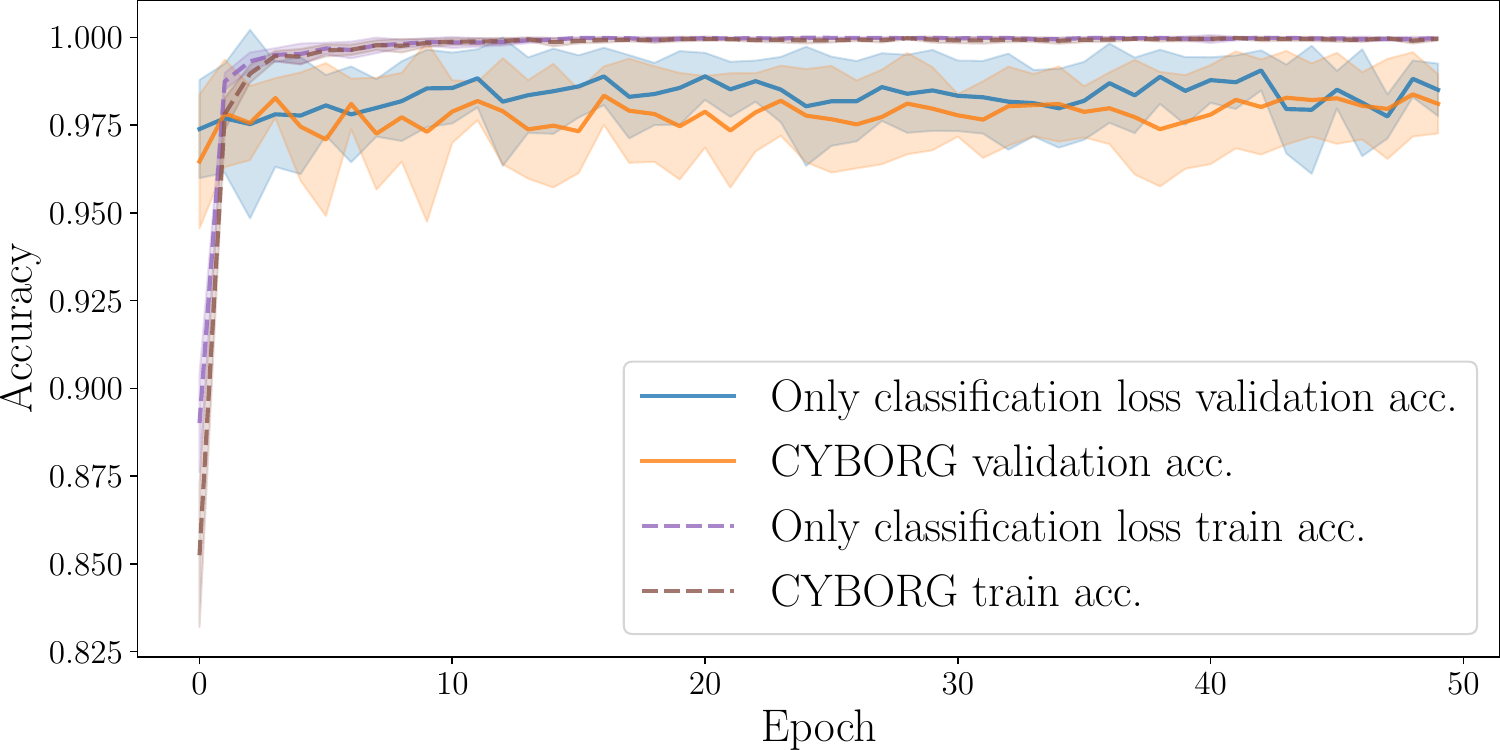}\vskip-1mm
     \caption{Same as in Fig. \ref{fig:densenet_training}, but for {\bf Xception Net}}
     \label{fig:xception_training}
     \null\vskip-5mm
\end{figure*}

Figures \ref{fig:densenet_training}-\ref{fig:xception_training}  outline the training and validation accuracy during the training processes for all four out-of-the-box architectures studied in this work.

Figures \ref{fig:dn_roc}-\ref{fig:xcp_roc} show the Receiver Operating Characteristics (ROC) curves for the results from Tab. \ref{tab:we_are_simply_the_best} for all four out-of-the-box architectures. The AUCs from Tab. \ref{tab:we_are_simply_the_best} can be found in the legends of the associated ROC plots. 

\subsection{Visualization of Model Output CAMs}

\begin{figure*}[t]
  \begin{subfigure}[b]{1\textwidth}
      \begin{subfigure}[b]{0.371\textwidth}
            \includegraphics[width=1\columnwidth]{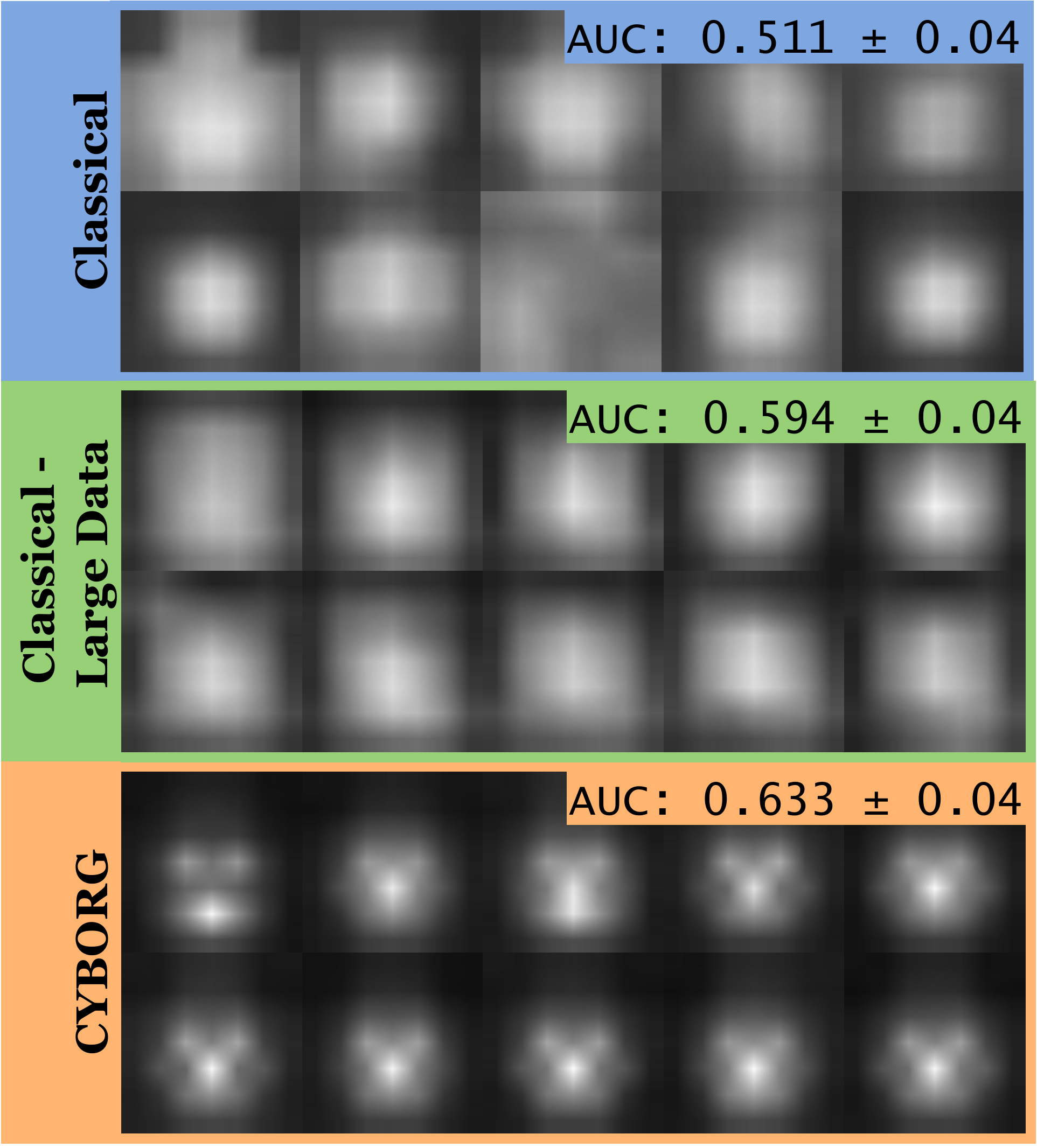}
          \caption{DenseNet-121}
      \end{subfigure}
      \begin{subfigure}[b]{0.333\textwidth}
          \includegraphics[width=1\columnwidth]{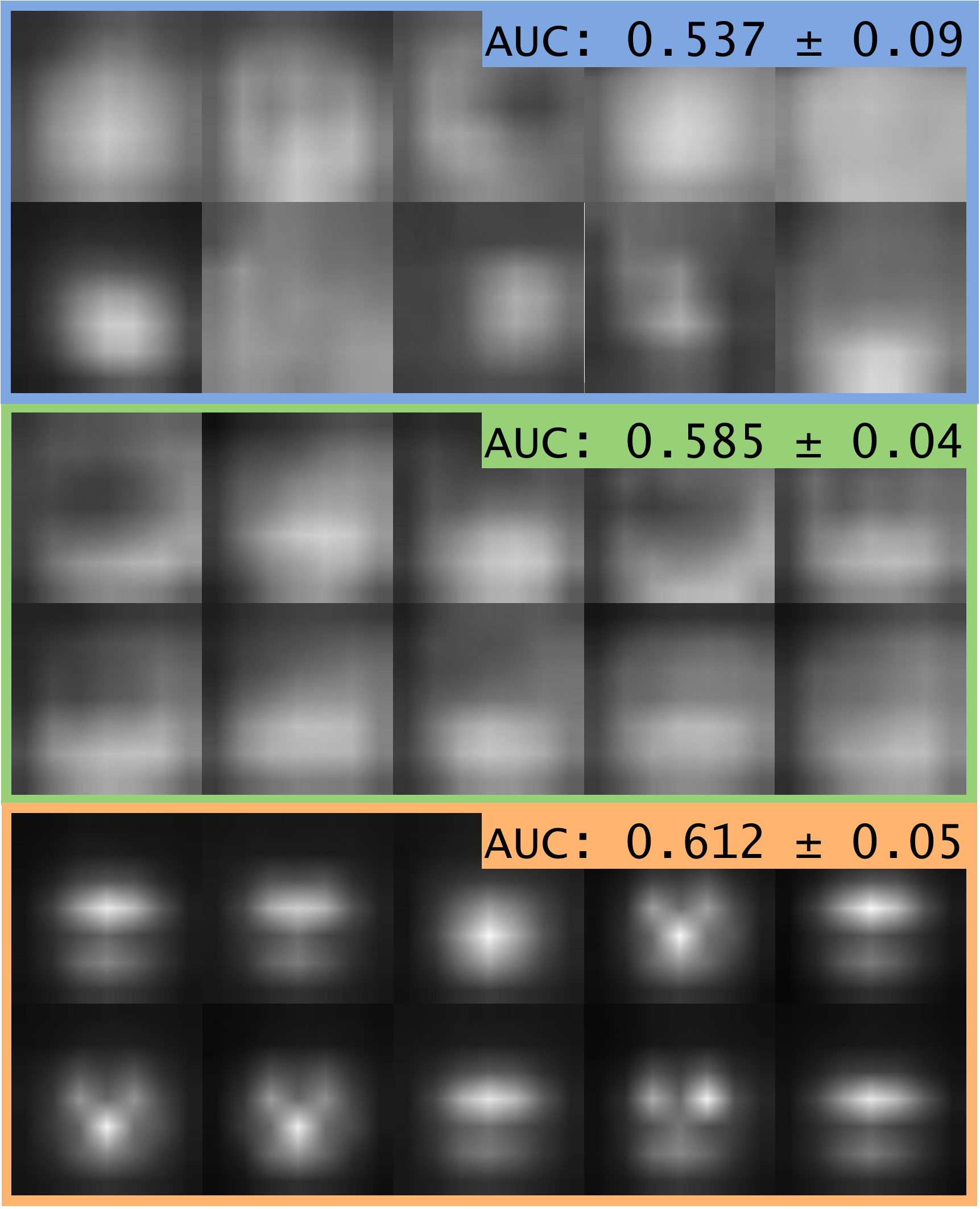}
          \caption{ResNet50}
      \end{subfigure}
    \end{subfigure}\vskip3mm
    
  \begin{subfigure}[b]{1\textwidth}
      \begin{subfigure}[b]{0.371\textwidth}
          \includegraphics[width=1\columnwidth]{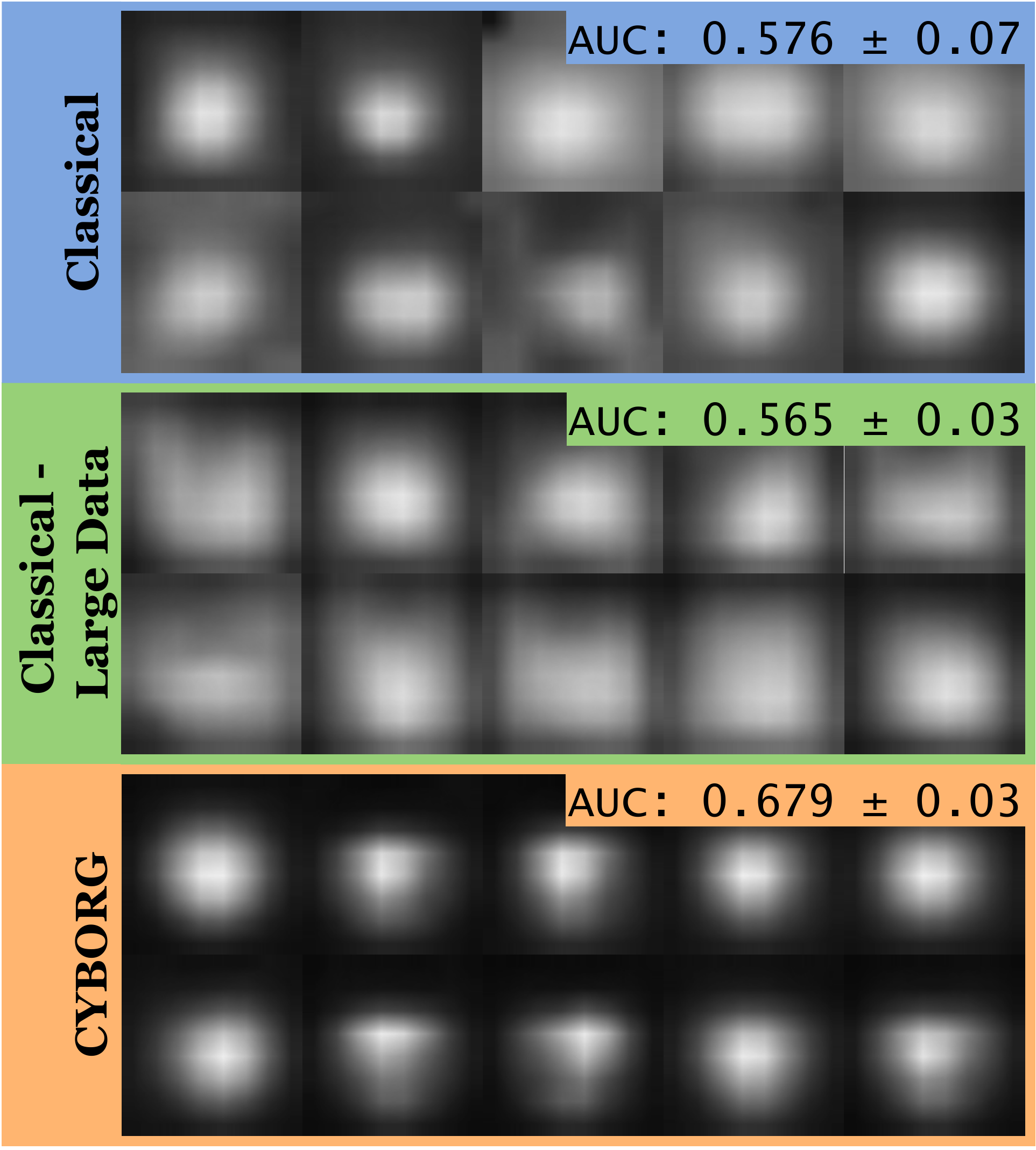}
          \caption{Inception-v3}
      \end{subfigure}
      \begin{subfigure}[b]{0.332\textwidth}
          \includegraphics[width=1\columnwidth]{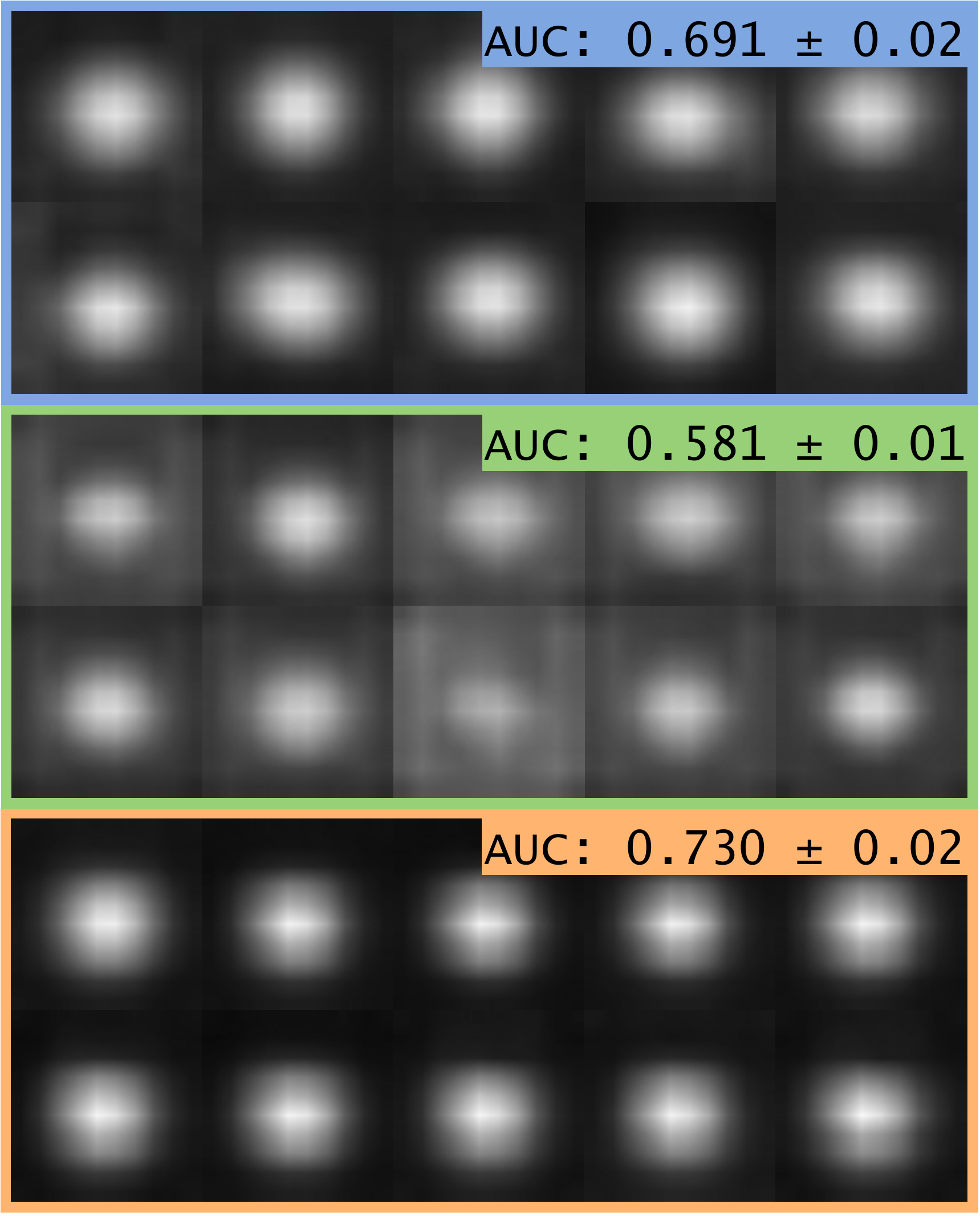}
          \caption{Xception}
      \end{subfigure}
    \begin{subfigure}[]{0.27\textwidth}
        \vskip-15cm
          \includegraphics[width=1\columnwidth]{figures/Visualizations/correct_annotations.png}
          \caption{Average Correct Human Annotation}
  \end{subfigure}
  \end{subfigure}
  \caption{Average CAMs across the entire test set for 10 independently trained models in three experimental settings and four different architectures.}
  \label{fig:visualizations_supp}
  \null\vskip-5mm
\end{figure*}

\begin{figure*}[!htb]
\vskip3mm
\centering
  \begin{subfigure}[b]{1\textwidth}
      \begin{subfigure}[b]{0.32\textwidth}
          \centering
            \includegraphics[width=1\columnwidth]{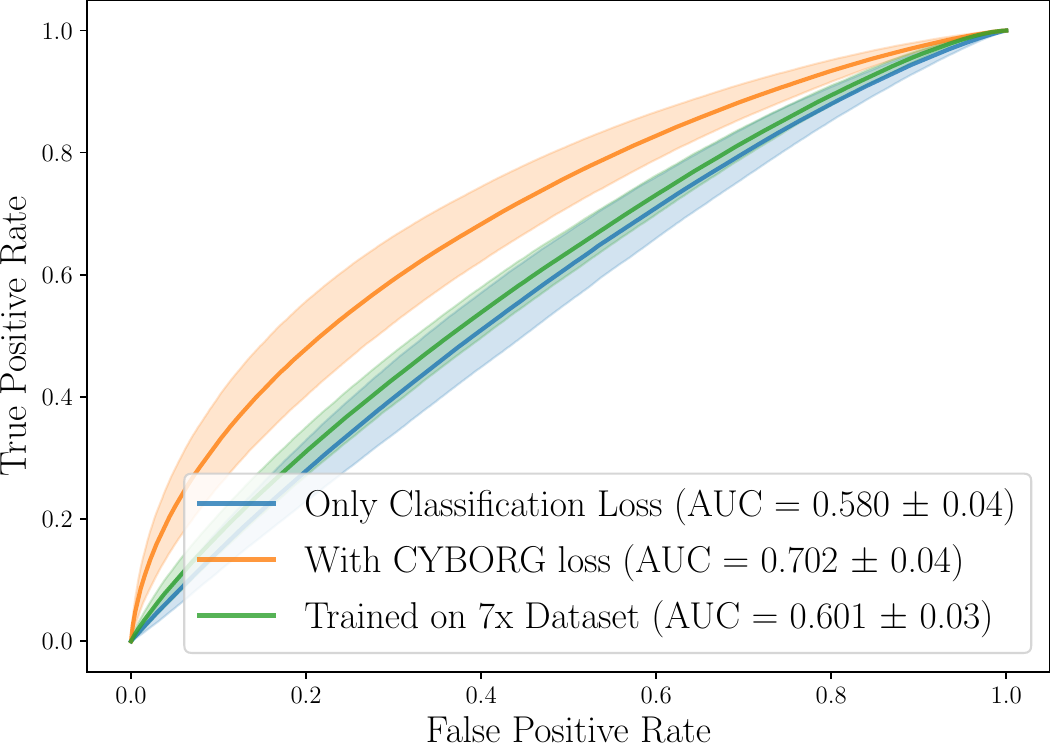}
          \caption{ProGAN}
      \end{subfigure}
      \hfill
      \begin{subfigure}[b]{0.32\textwidth}
          \centering
          \includegraphics[width=1\columnwidth]{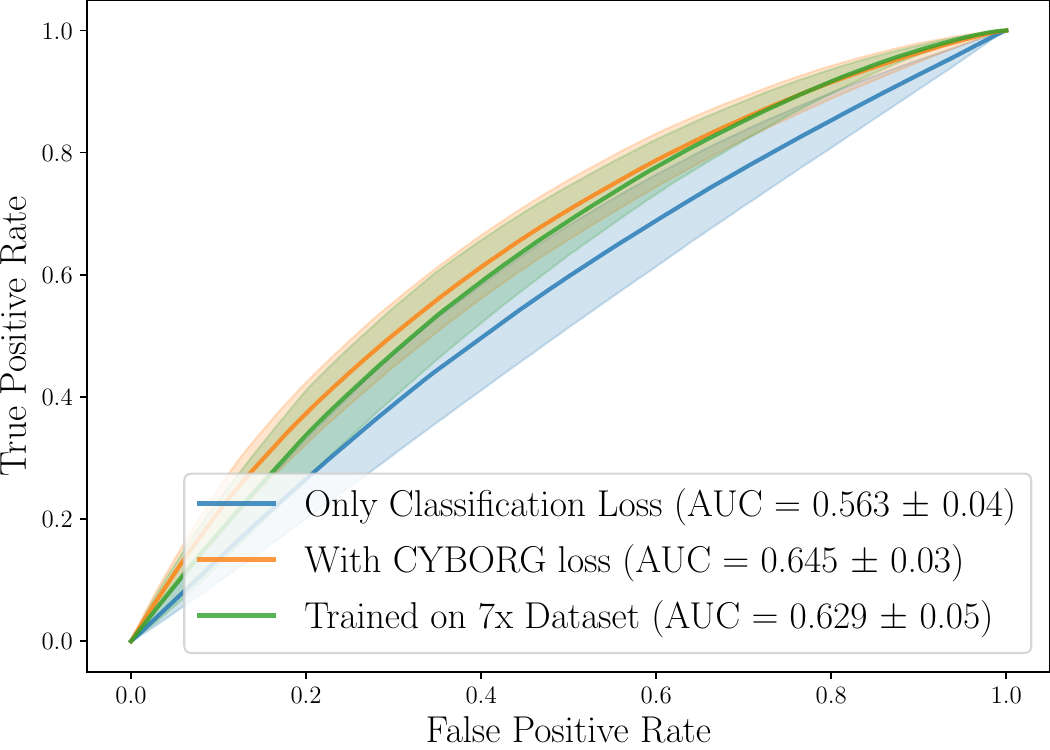}
          \caption{StyleGAN}
      \end{subfigure}
      \hfill
      \begin{subfigure}[b]{0.32\textwidth}
          \centering
          \includegraphics[width=1\columnwidth]{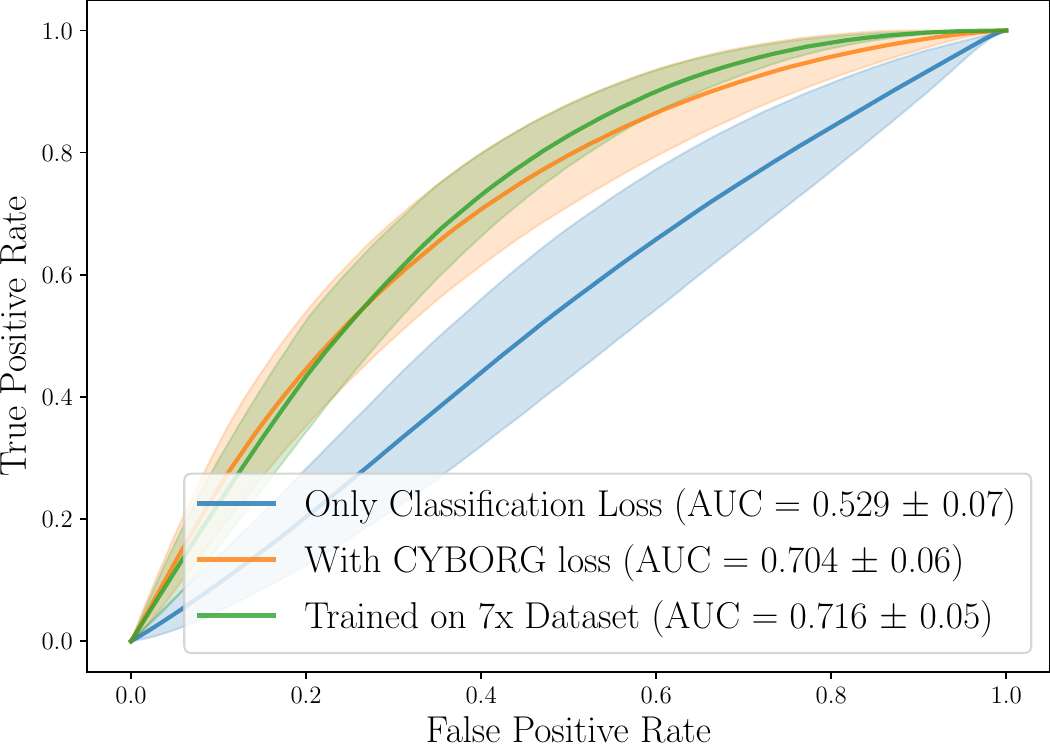}
          \caption{StyleGAN2}
      \end{subfigure}

  \end{subfigure}\vskip3mm
  \begin{subfigure}[b]{1\textwidth}
      \centering
      \begin{subfigure}[b]{0.32\textwidth}
          \centering
          \includegraphics[width=1\columnwidth]{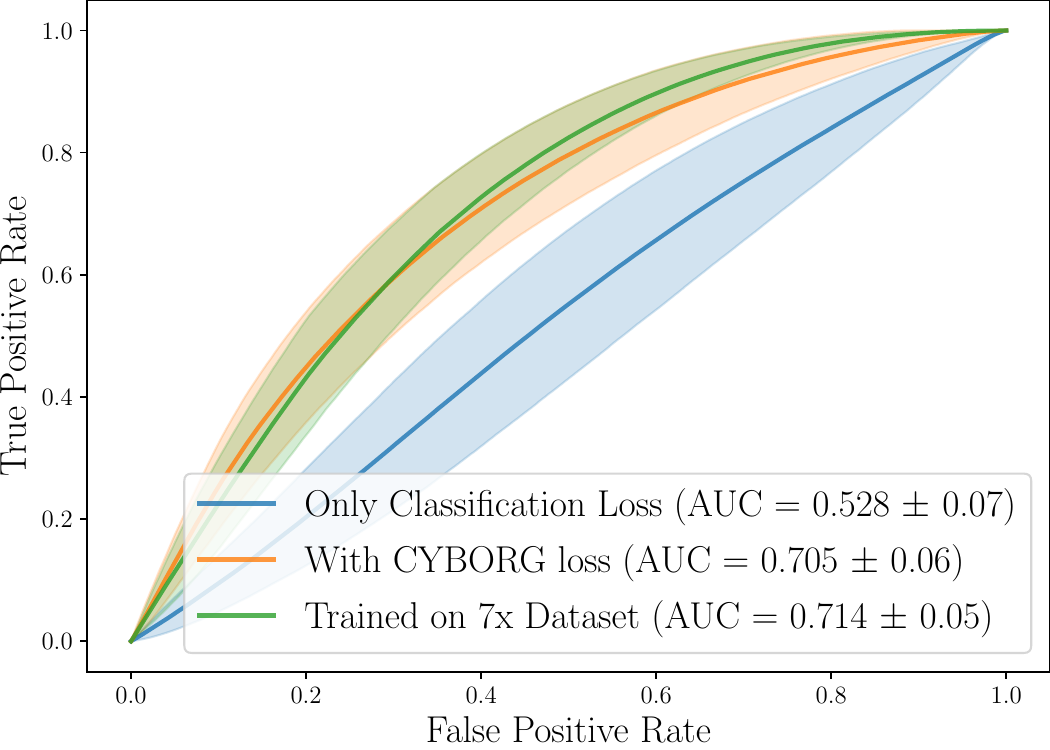}
          \caption{StyleGAN2-ADA}
      \end{subfigure}
      \begin{subfigure}[b]{0.32\textwidth}
          \centering
          \includegraphics[width=1\columnwidth]{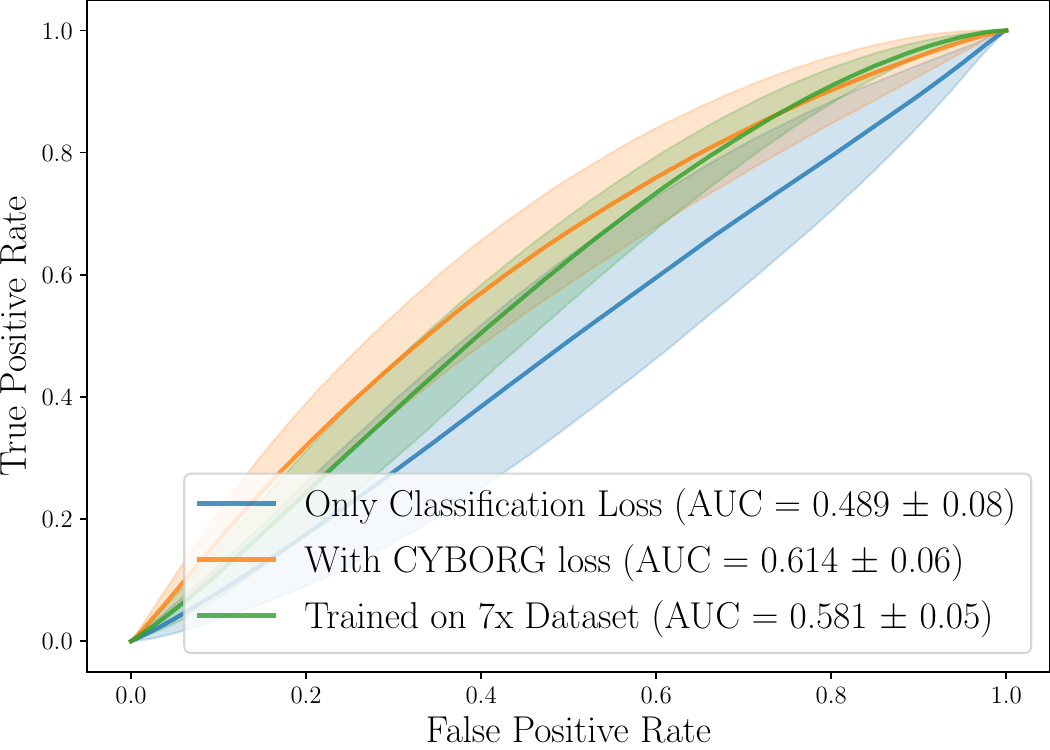}
          \caption{StyleGAN3}
      \end{subfigure}
      \begin{subfigure}[b]{0.32\textwidth}
          \centering
          \includegraphics[width=1\columnwidth]{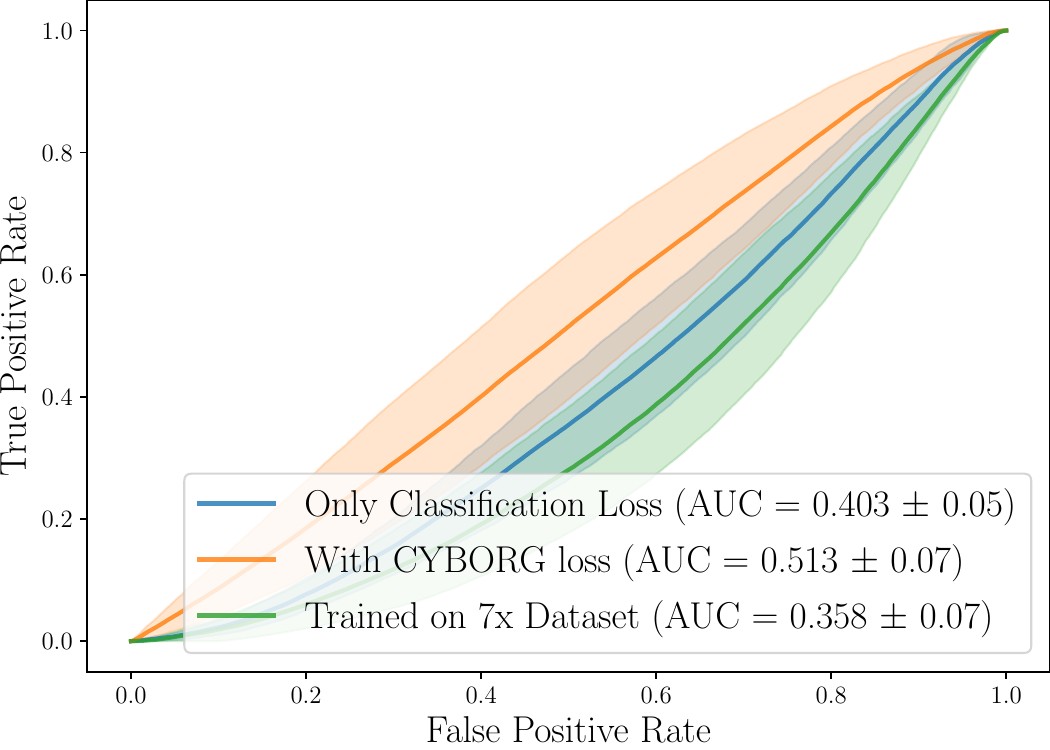}
          \caption{StarGAN v2}
      \end{subfigure}
  \end{subfigure}\vskip-1mm
  \caption{ROC curves associated with results shown in Tab. 1. Classification model: {\bf DenseNet121}. ``Only classification loss'' corresponds to Scenario 1 (``Classical Training''), ``Trained with 7x Dataset'' corresponds to Scenario 2 (``Classical -- Large Data''), and ``CYBORG'' corresponds to Scenario 3, as defined in Sec. 6.1 in the main paper.}
  \label{fig:dn_roc}
\end{figure*}

\begin{figure*}[!htb]
\centering
  \begin{subfigure}[b]{1\textwidth}
      \begin{subfigure}[b]{0.32\textwidth}
          \centering
            \includegraphics[width=1\columnwidth]{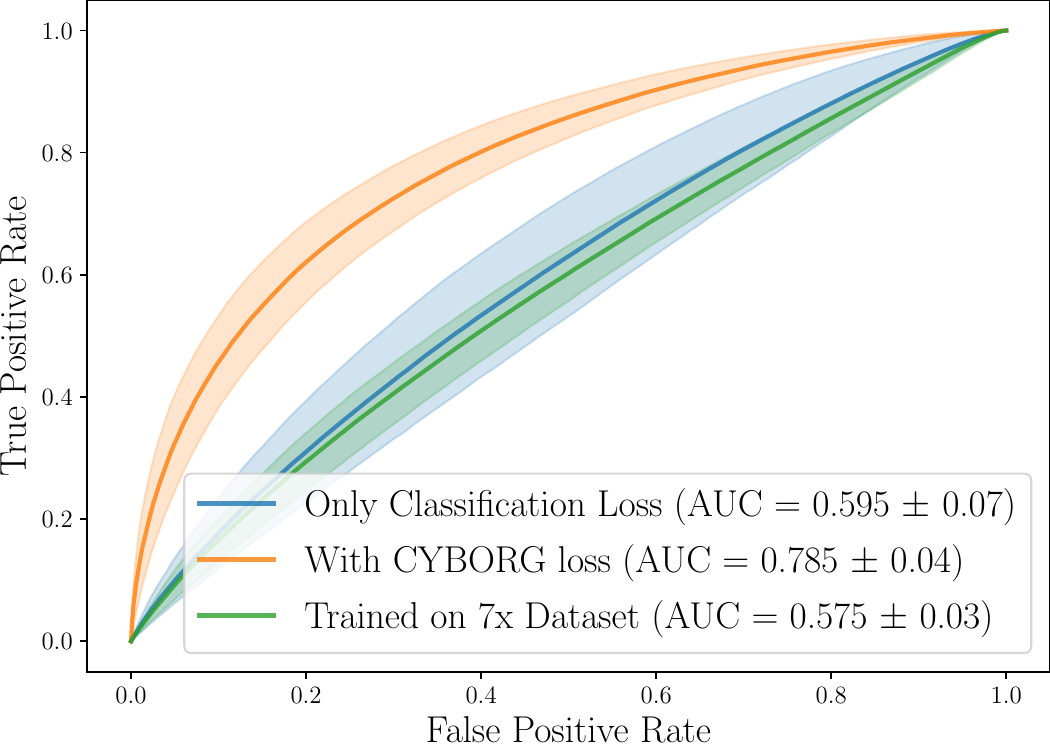}
          \caption{ProGAN}
      \end{subfigure}
      \hfill
      \begin{subfigure}[b]{0.32\textwidth}
          \centering
          \includegraphics[width=1\columnwidth]{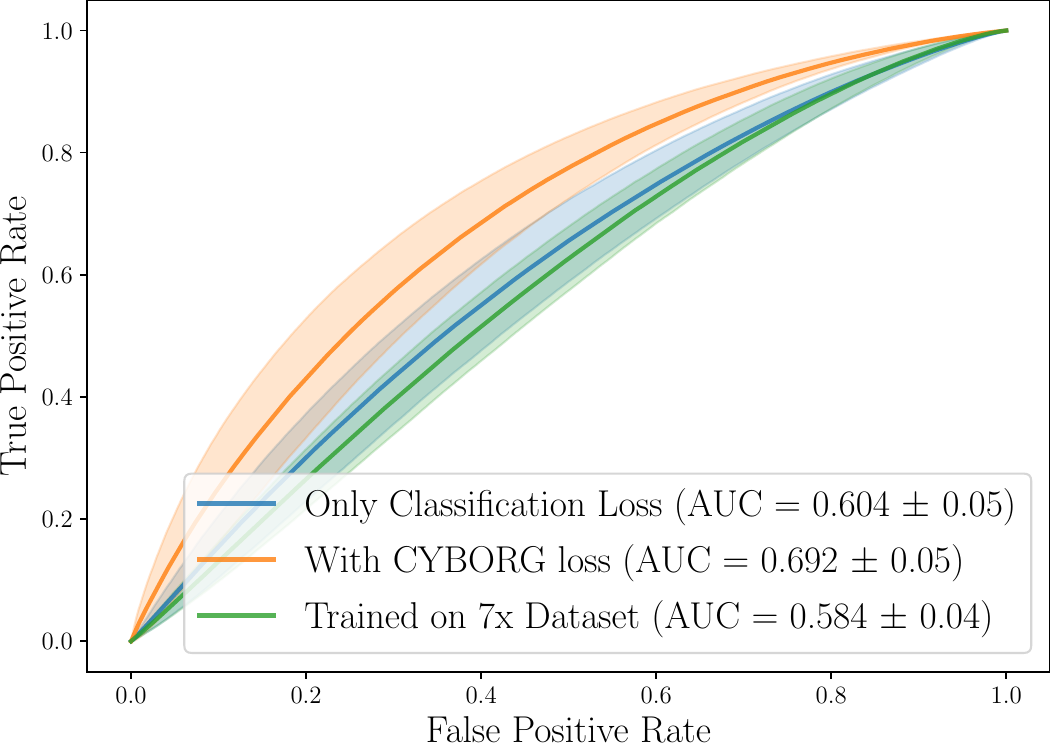}
          \caption{StyleGAN}
      \end{subfigure}
      \hfill
      \begin{subfigure}[b]{0.32\textwidth}
          \centering
          \includegraphics[width=1\columnwidth]{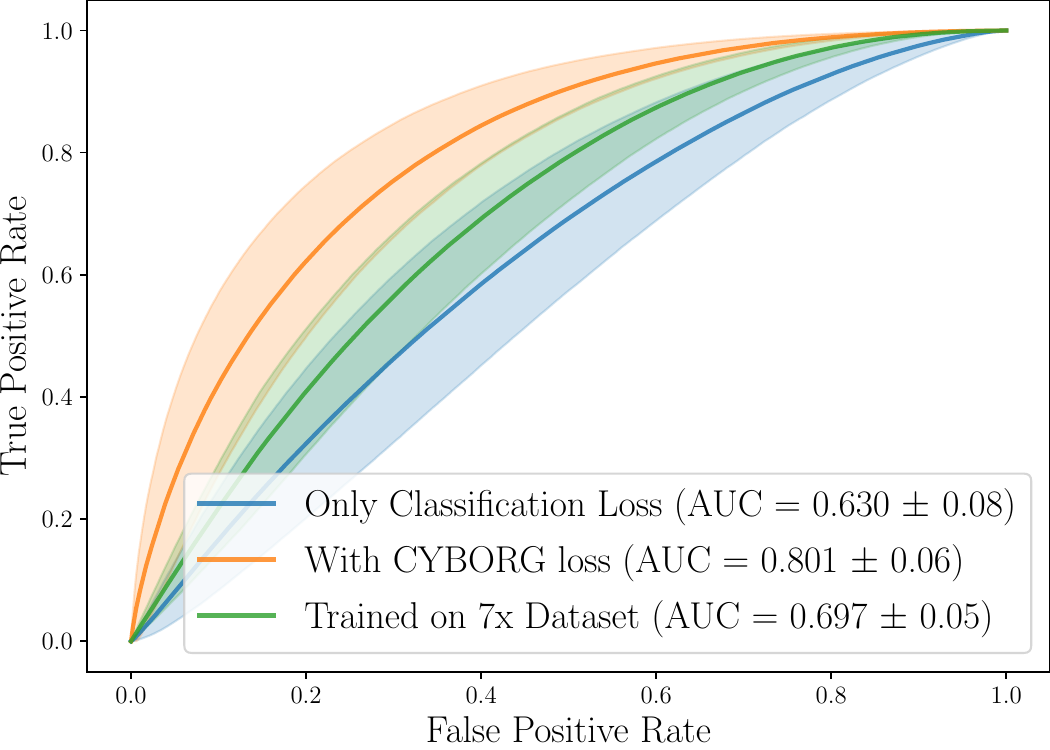}
          \caption{StyleGAN2}
      \end{subfigure}

  \end{subfigure}\vskip3mm
  \begin{subfigure}[b]{1\textwidth}
      \centering
      \begin{subfigure}[b]{0.32\textwidth}
          \centering
          \includegraphics[width=1\columnwidth]{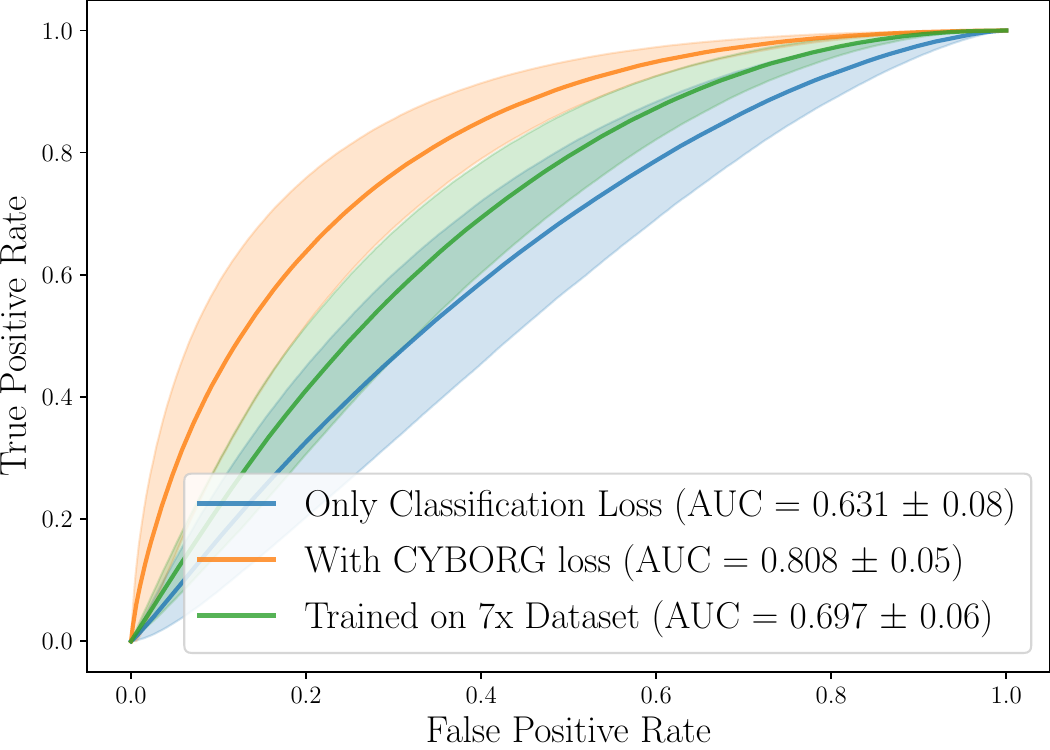}
          \caption{StyleGAN2-ADA}
      \end{subfigure}
      \begin{subfigure}[b]{0.32\textwidth}
          \centering
          \includegraphics[width=1\columnwidth]{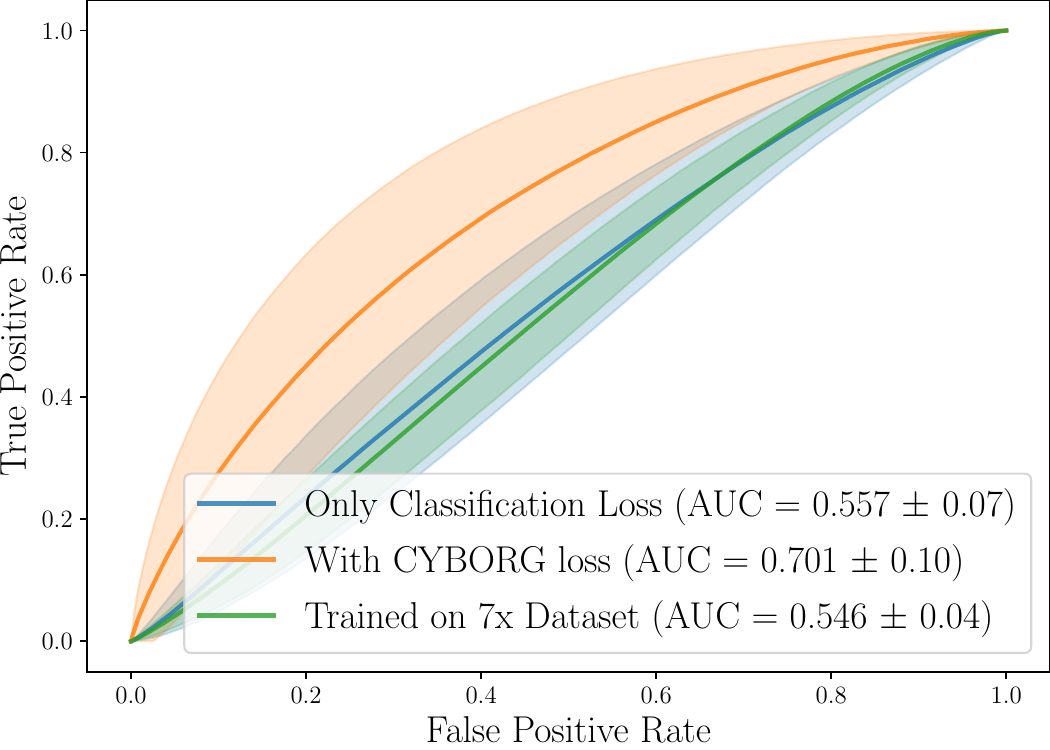}
          \caption{StyleGAN3}
      \end{subfigure}
      \begin{subfigure}[b]{0.32\textwidth}
          \centering
          \includegraphics[width=1\columnwidth]{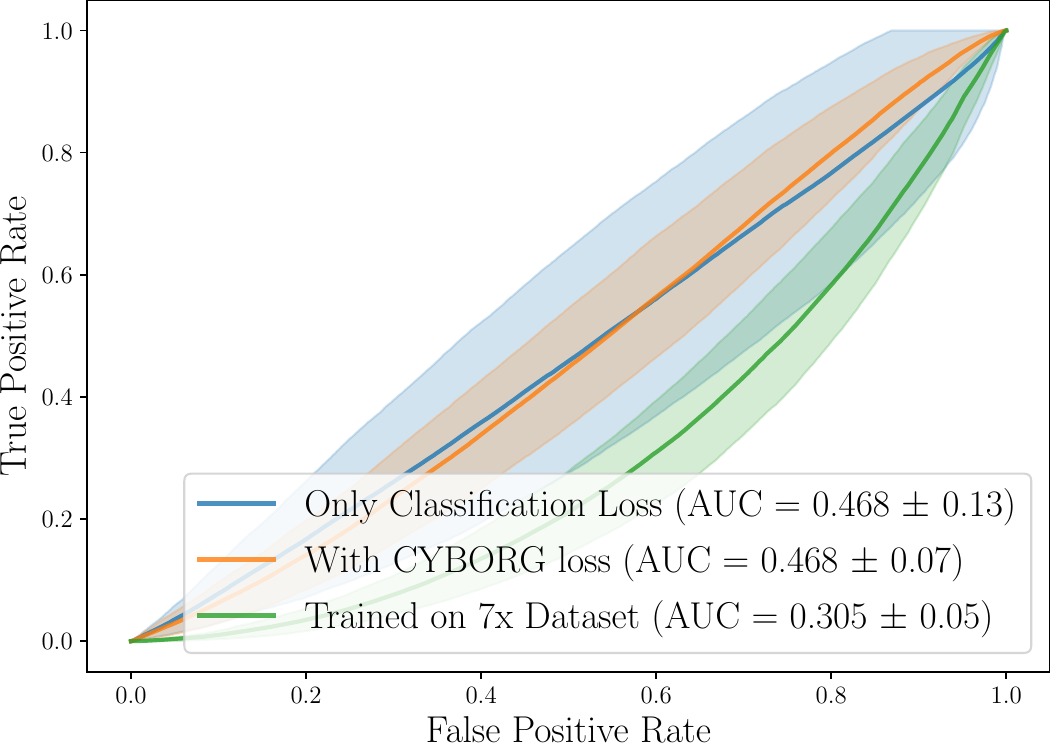}
          \caption{StarGAN v2}
      \end{subfigure}
  \end{subfigure}\vskip-1mm
\caption{Same as in Fig. \ref{fig:dn_roc}, except for classification model: {\bf ResNet50}}
\label{fig:rn_roc}
\end{figure*}

\begin{figure*}[!htb]
\centering
  \begin{subfigure}[b]{1\textwidth}
      \begin{subfigure}[b]{0.32\textwidth}
          \centering
            \includegraphics[width=1\columnwidth]{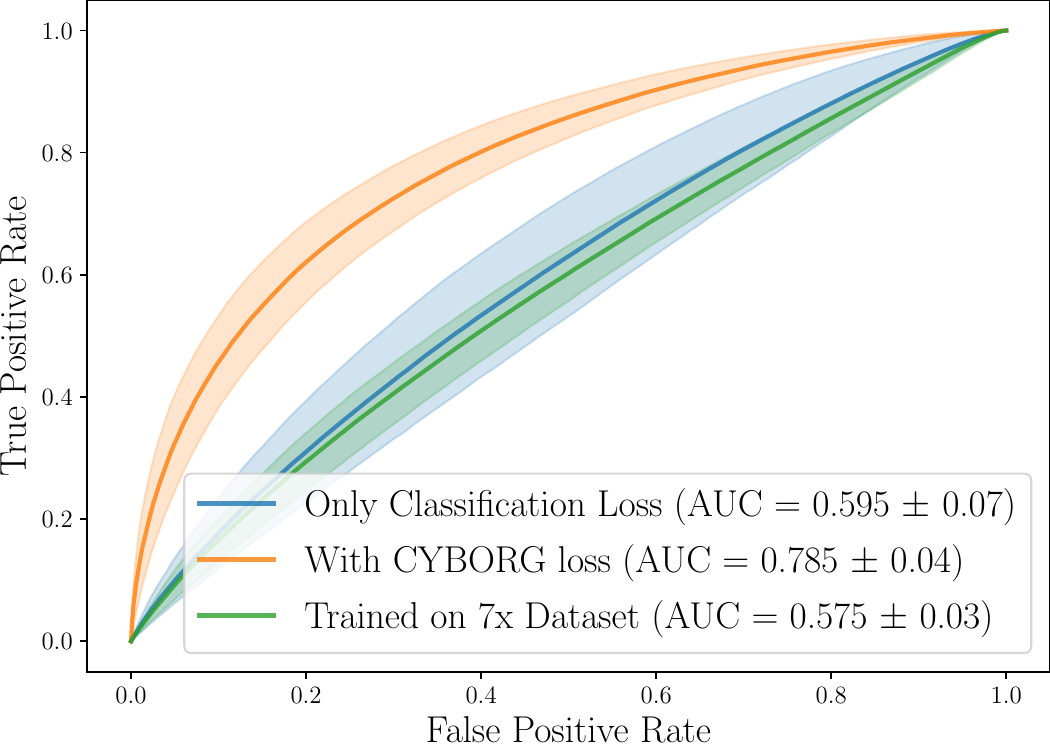}
          \caption{ProGAN}
      \end{subfigure}
      \hfill
      \begin{subfigure}[b]{0.32\textwidth}
          \centering
          \includegraphics[width=1\columnwidth]{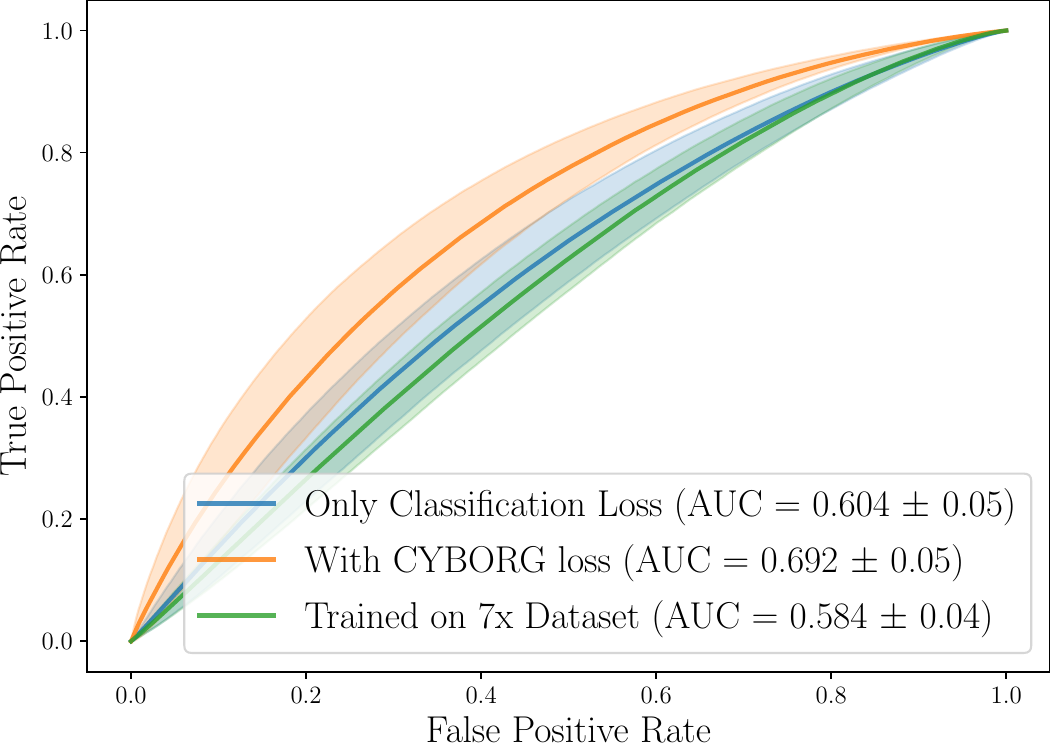}
          \caption{StyleGAN}
      \end{subfigure}
      \hfill
      \begin{subfigure}[b]{0.32\textwidth}
          \centering
          \includegraphics[width=1\columnwidth]{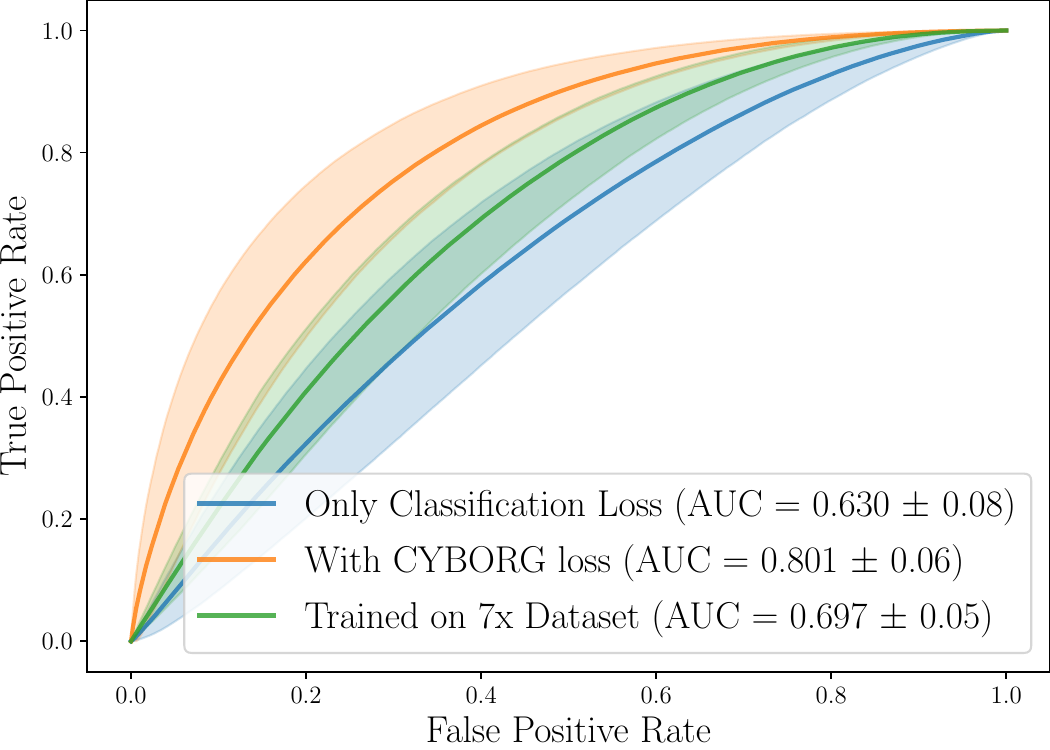}
          \caption{StyleGAN2}
      \end{subfigure}

  \end{subfigure}\vskip3mm
  \begin{subfigure}[b]{1\textwidth}
      \centering
      \begin{subfigure}[b]{0.32\textwidth}
          \centering
          \includegraphics[width=1\columnwidth]{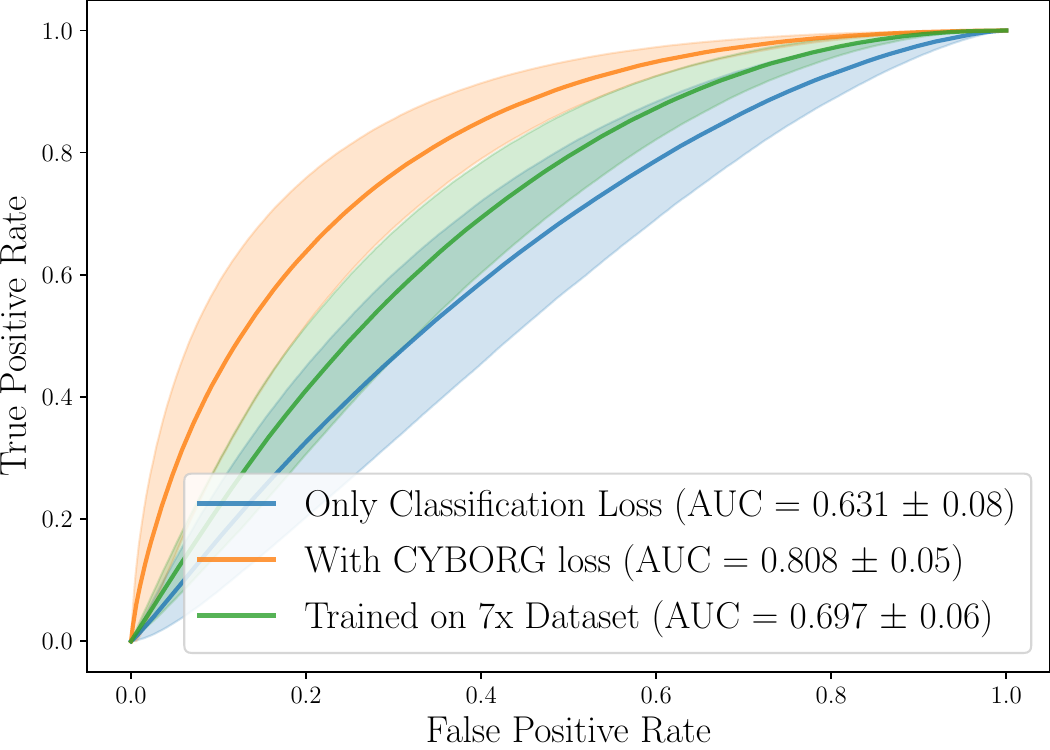}
          \caption{StyleGAN2-ADA}
      \end{subfigure}
      \begin{subfigure}[b]{0.32\textwidth}
          \centering
          \includegraphics[width=1\columnwidth]{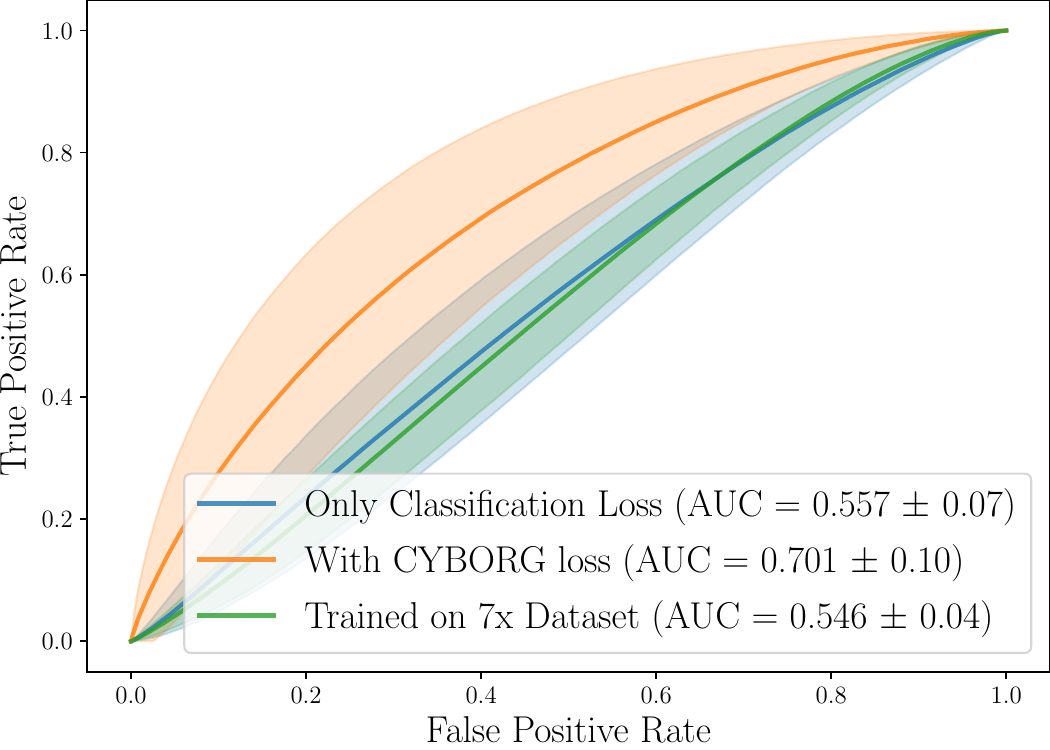}
          \caption{StyleGAN3}
      \end{subfigure}
      \begin{subfigure}[b]{0.32\textwidth}
          \centering
          \includegraphics[width=1\columnwidth]{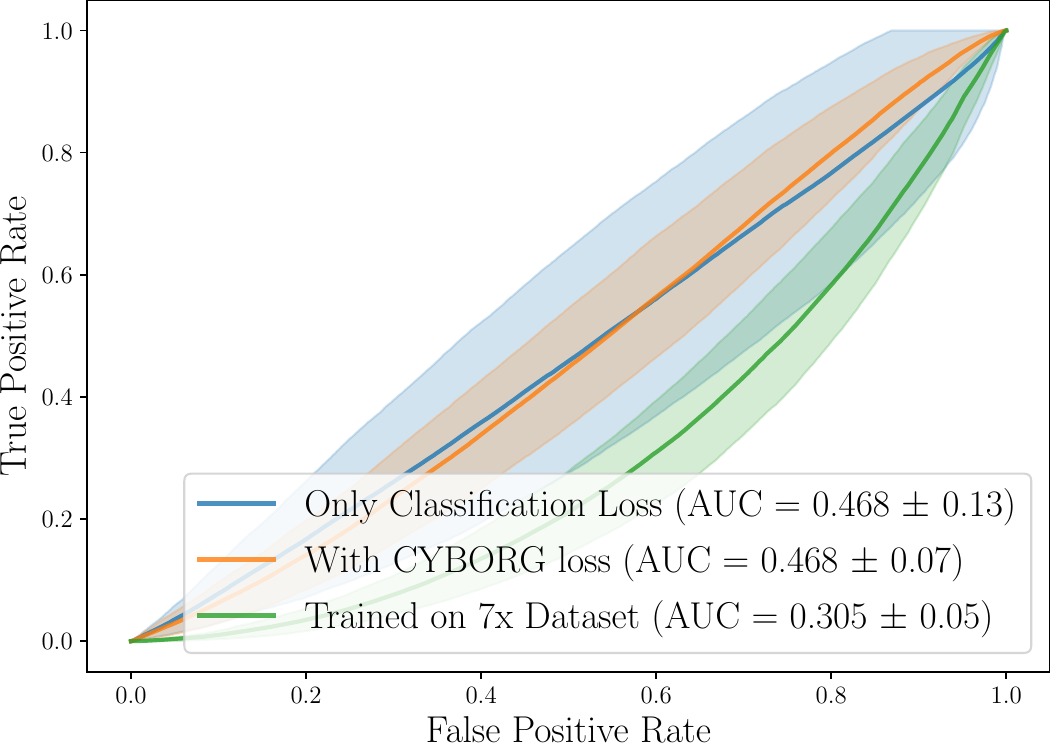}
          \caption{StarGAN v2}
      \end{subfigure}
  \end{subfigure}\vskip-1mm
\caption{Same as in Fig. \ref{fig:dn_roc}, except for classification model: {\bf Inception-v3}}
\label{fig:inc_roc}
\end{figure*}

\begin{figure*}[!htb]
\centering
  \begin{subfigure}[b]{1\textwidth}
      \begin{subfigure}[b]{0.32\textwidth}
          \centering
            \includegraphics[width=1\columnwidth]{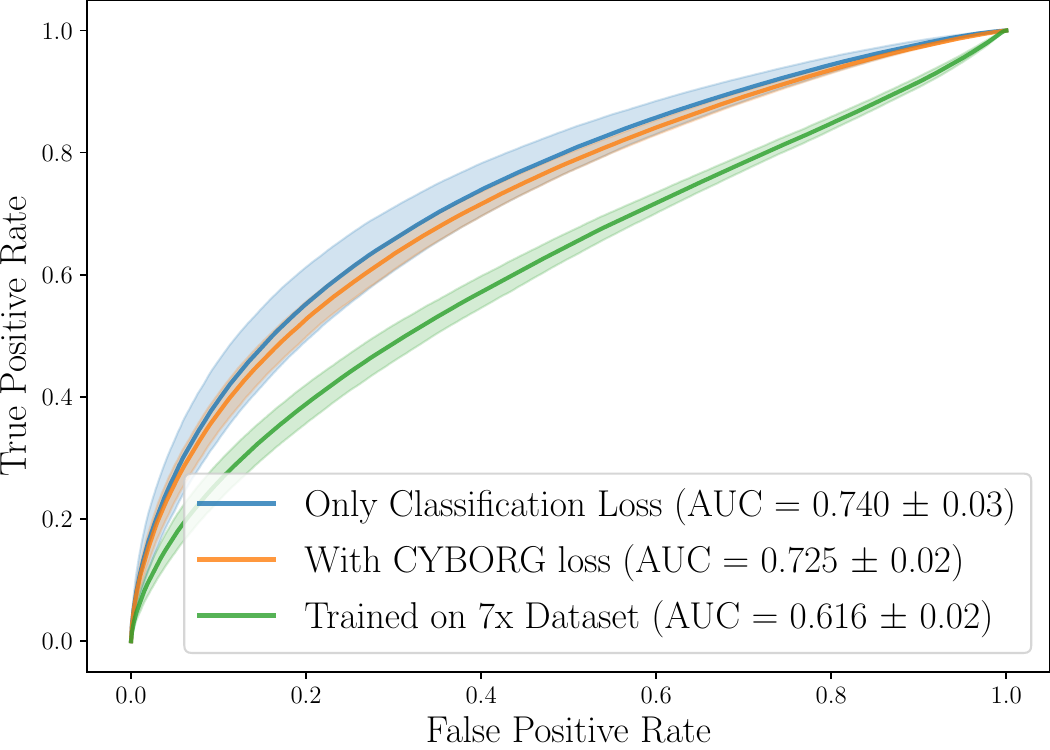}
          \caption{ProGAN}
      \end{subfigure}
      \hfill
      \begin{subfigure}[b]{0.32\textwidth}
          \centering
          \includegraphics[width=1\columnwidth]{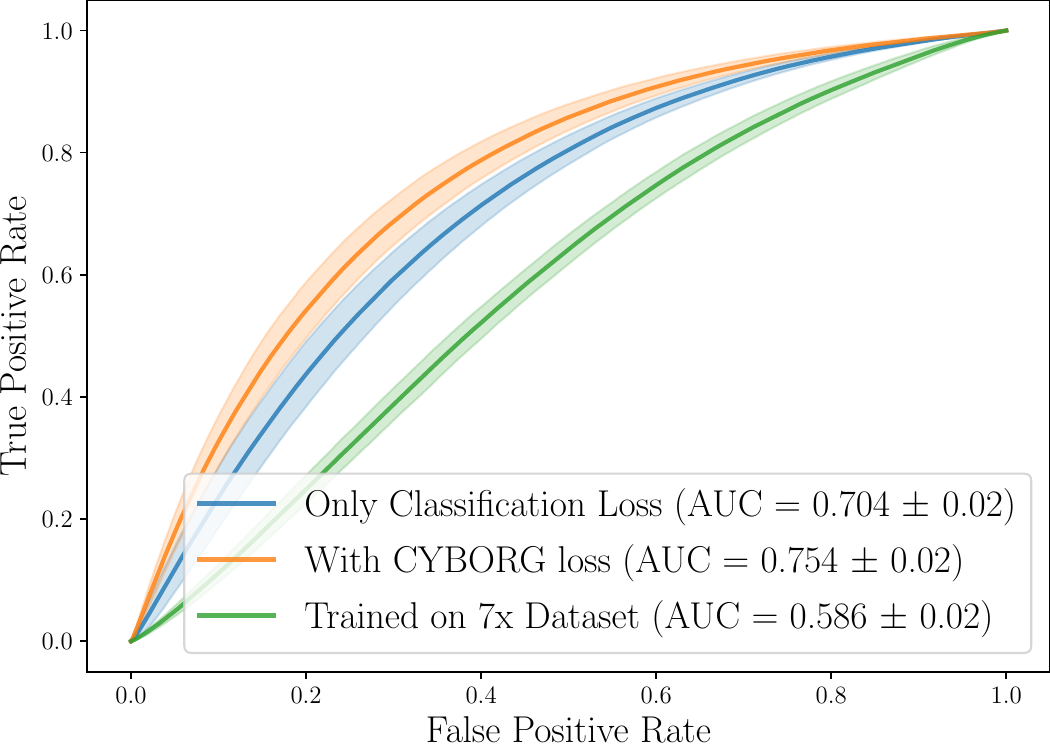}
          \caption{StyleGAN}
      \end{subfigure}
      \hfill
      \begin{subfigure}[b]{0.32\textwidth}
          \centering
          \includegraphics[width=1\columnwidth]{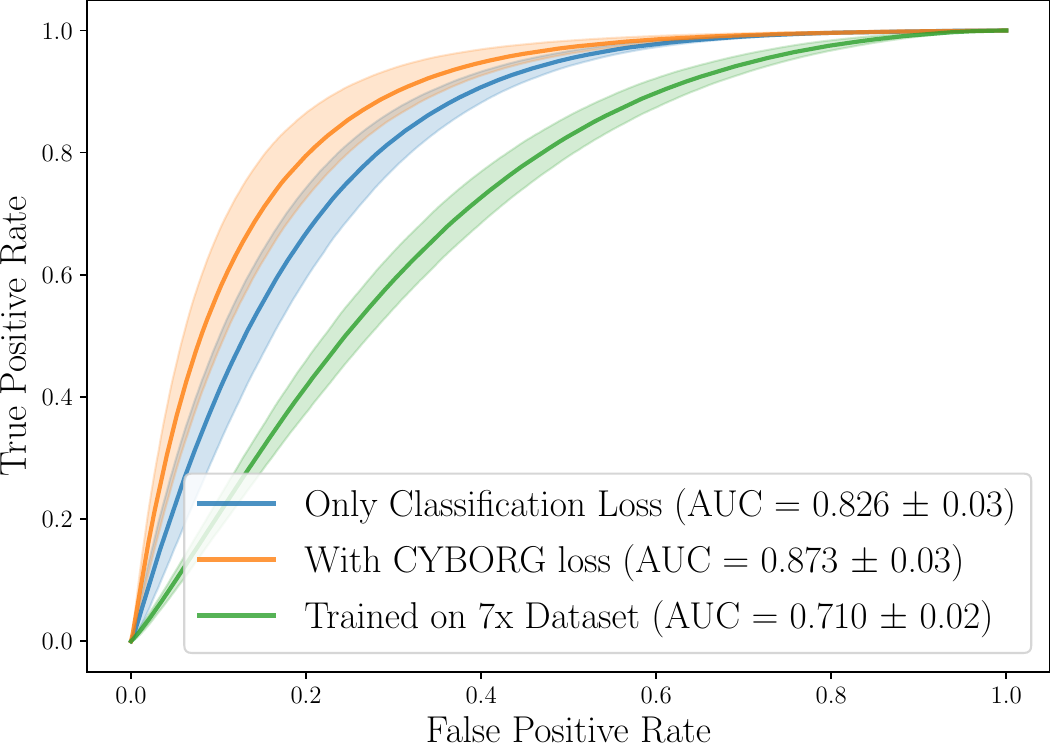}
          \caption{StyleGAN2}
      \end{subfigure}

  \end{subfigure}\vskip3mm
  \begin{subfigure}[b]{1\textwidth}
      \centering
      \begin{subfigure}[b]{0.32\textwidth}
          \centering
          \includegraphics[width=1\columnwidth]{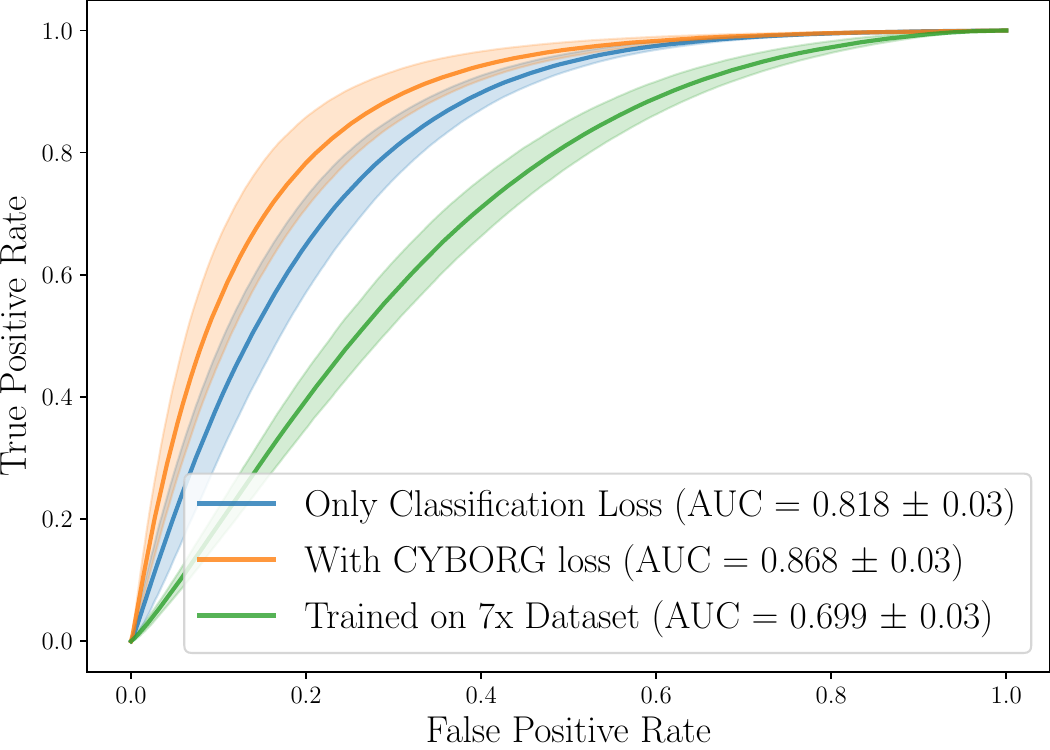}
          \caption{StyleGAN2-ADA}
      \end{subfigure}
      \begin{subfigure}[b]{0.32\textwidth}
          \centering
          \includegraphics[width=1\columnwidth]{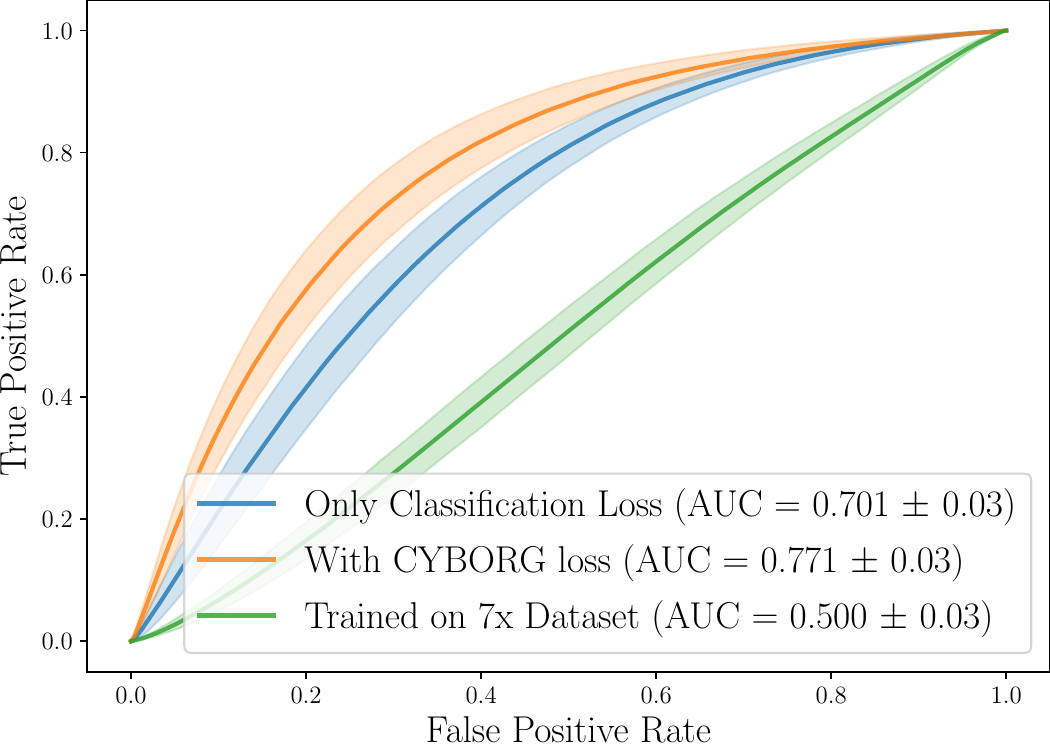}
          \caption{StyleGAN3}
      \end{subfigure}
      \begin{subfigure}[b]{0.32\textwidth}
          \centering
          \includegraphics[width=1\columnwidth]{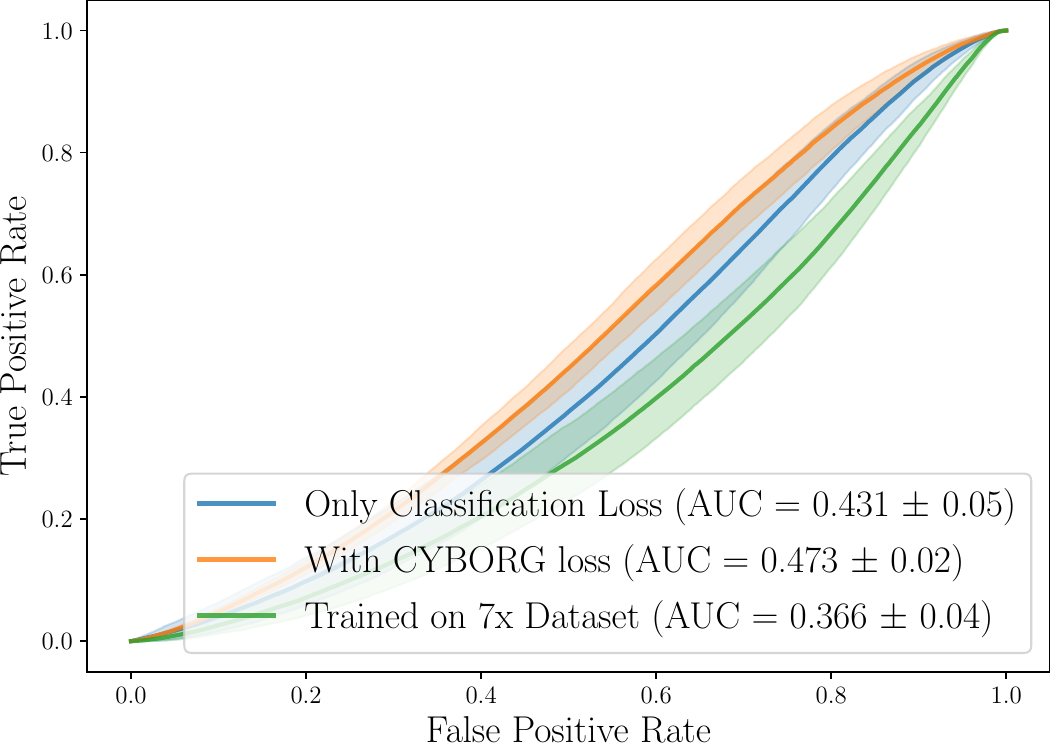}
          \caption{StarGAN v2}
      \end{subfigure}
  \end{subfigure}\vskip-1mm
\caption{Same as in Fig. \ref{fig:dn_roc}, except for classification model: {\bf Xception Net}}
\label{fig:xcp_roc}
\end{figure*}

Detailed here are the model visualizations for all studied architectures. In the manuscript Fig. 7 only ResNet is shown but here all four networks are displayed. This Fig. \ref{fig:visualizations_supp} in this supplementary materials complements Fig. 7 in the paper.

By direct comparison to the average correct human annotation in Fig. \ref{fig:visualizations_supp}(e), the models trained with CYBORG focus on features that are more similar to human detailed salient regions than models trained with classical cross-entropy loss in all cases. Thus, it can be concluded these models are effectively guided by the human annotations supplied during the training process.

\subsection{Evaluating an Off-the-shelf Deepfake Detector on Test Data}

While deepfake technology manipulates real videos by inter-splicing real identities, GAN-generated images are entirely synthetic. Given the slight difference between the two, we wanted to inspect whether or not existing \textit{deepfake} detection methods can be applied to our task of \textit{synthetic image} detection. To verify, we ran a state-of-the-art deepfake detector~\cite{bonettini2021video} on our test set of synthetic images.  

\begin{figure*}[!htb] 
    \centering
    \begin{subfigure}[b]{1\textwidth}
        \begin{subfigure}[b]{0.49\textwidth}
          \centering
            \includegraphics[width=1\columnwidth]{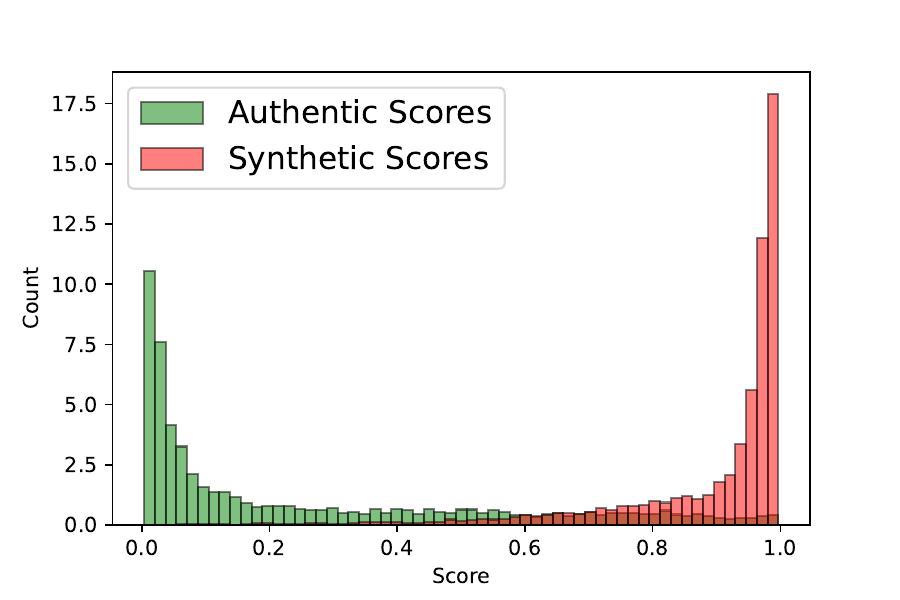}
          \caption{DFDC}
        \end{subfigure}
        \hfill
        \begin{subfigure}[b]{0.49\textwidth}
          \centering
          \includegraphics[width=1\columnwidth]{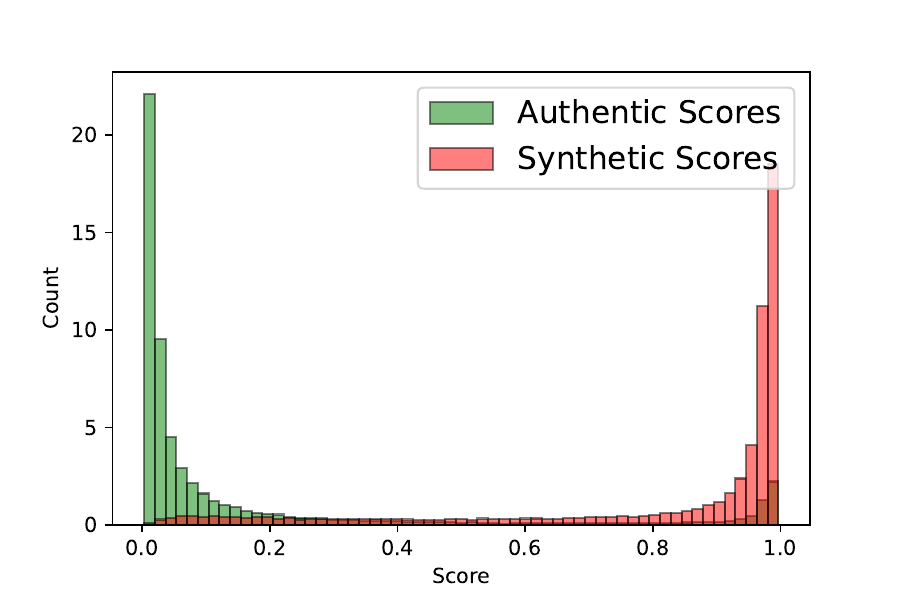}
          \caption{FF++}
         \end{subfigure}
    \end{subfigure}
    \caption{Best reported ensemble method among off-the-shelf deepfake detector models when applied to \textbf{the original authors'} respective DFDC and FF++ deepfake test data. As in the original paper, this deepfake detector performs very well on deepfake samples.}
    \label{fig:ensemble_theirs}
    \null\vskip-5mm
\end{figure*}

Prior to evaluating the deepfake models on our test images, model use and accuracy were validated for the ten pretrained models on their respective test sets: DFDC~\cite{DFDC2020} and FFPP~\cite{roessler2019faceforensicspp}. Figure \ref{fig:ensemble_theirs} shows the results from the authors' self-reported best ensemble deepfake detection method. On \textit{their} own DFDC and FF++ deepfake test data, the \cite{bonettini2021video} ensemble methods show AUCs of 0.95789 and 0.92047, respectively. 

\begin{figure*}[!htb] 
    \centering
    \begin{subfigure}[b]{1\textwidth}
        \begin{subfigure}[b]{0.49\textwidth}
          \centering
            \includegraphics[width=1\columnwidth]{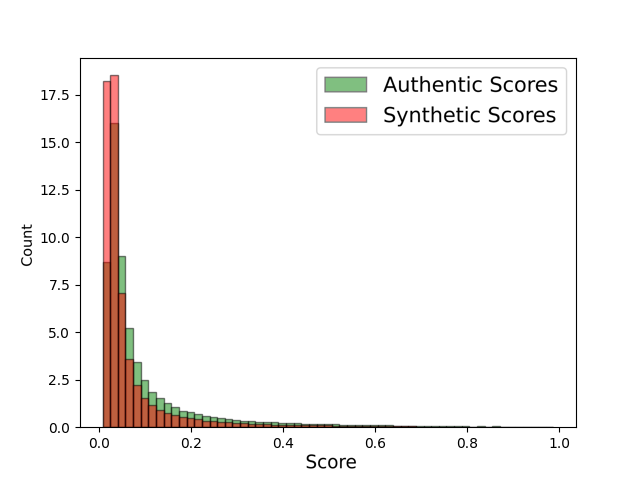}
          \caption{DFDC}
        \end{subfigure}
        \hfill
        \begin{subfigure}[b]{0.49\textwidth}
          \centering
          \includegraphics[width=1\columnwidth]{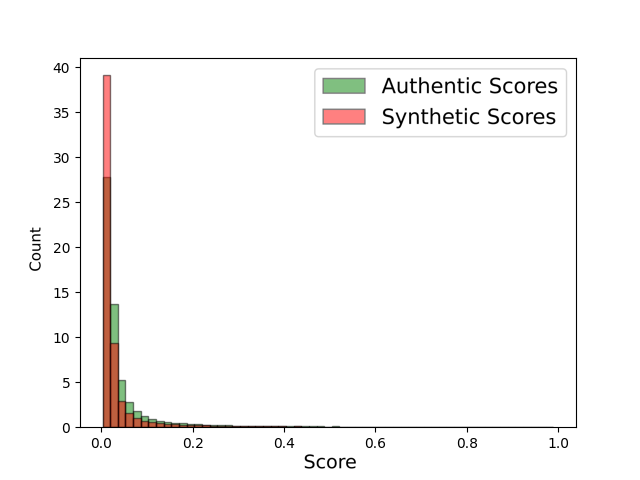}
          \caption{FF++}
         \end{subfigure}
    \end{subfigure}
    \caption{Best reported ensemble method among off-the-shelf deepfake detector models when applied to \textbf{our synthetic face test data}. As it can be seen, the method designed for deep fakes detection is not able to detect synthetically-generated faces.}
    \label{fig:ensemble-ours}
    \null\vskip-5mm
\end{figure*}

However, when applied in the different domain of synthetic image detection (on \textit{our} synthetic test data), the ensembles of pretrained models are not able to adequately distinguish between real and entirely synthetic images, as seen in Figure \ref{fig:ensemble-ours}; the DFDC ensemble method shows an AUC of 0.373 while the FFPP ensemble method shows an AUC of 0.385. These results support the claim that deepfake detection models and synthetic image detection models are \textbf{\textit{not}} interchangeable. 

\subsection{Incorporation Of CYBORG Into An Existing Synthetic Face Detector}
In~\cite{wang2019cnngenerated},  Wang \etal design a synthesizer-agnostic classifier of CNN-generated images. To test model generalizability on novel synthesizers, Wang \etal trained their model exclusively on ProGAN-generated images. They then tested their model on ``never-before-seen'' GAN-generated images from StyleGAN~\cite{karras2019style}, CycleGAN~\cite{zhu2017unpaired}, and other state-of-the-art methods. 

To determine whether the incorporation of CYBORG loss would improve upon this existing method, we conducted experiments under the training scenarios and testing protocols described in the main paper (Sec. 6.1), adding CYBORG loss to Wang \etal's publicly available, re-trainable model~\cite{wang2021repo}.

\begin{figure*}[!htb]
\centering
  \begin{subfigure}[b]{1\textwidth}
      \begin{subfigure}[b]{0.32\textwidth}
          \centering
            \includegraphics[width=1\columnwidth]{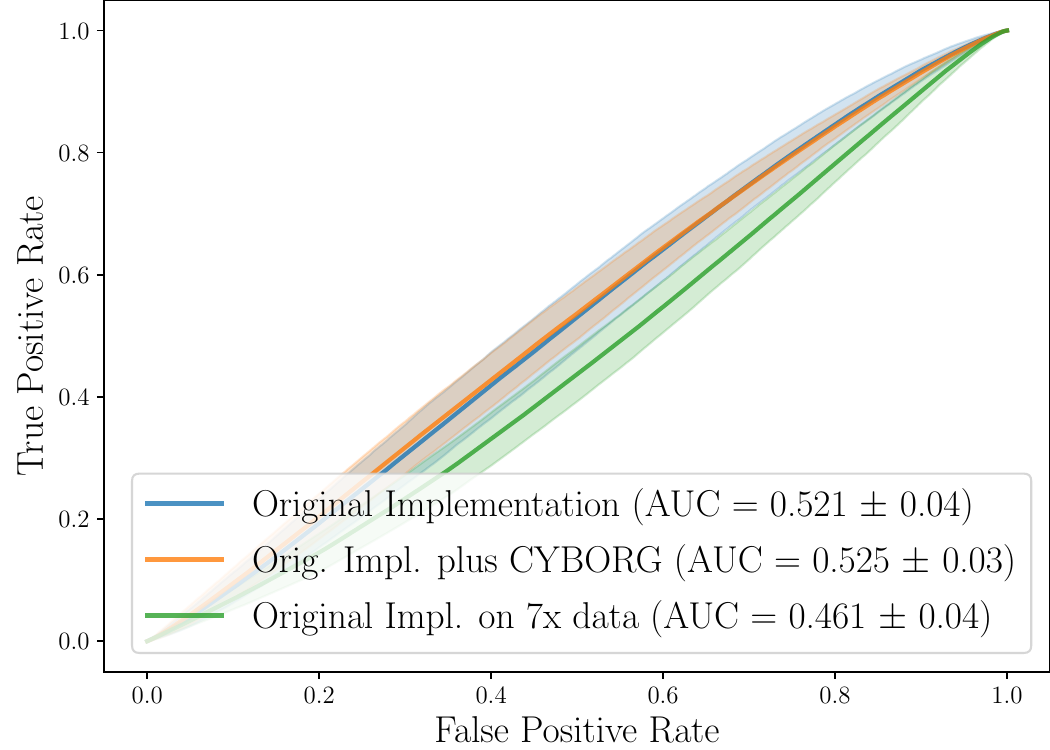}
          \caption{ProGAN}
      \end{subfigure}
      \hfill
      \begin{subfigure}[b]{0.32\textwidth}
          \centering
          \includegraphics[width=1\columnwidth]{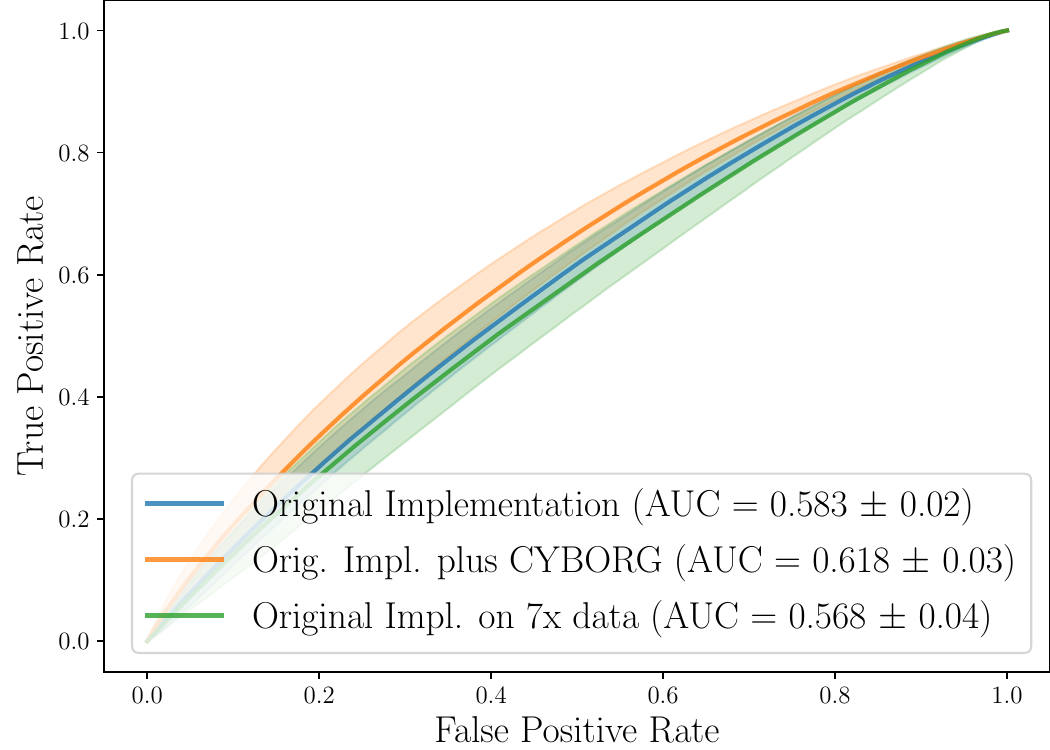}
          \caption{StyleGAN}
      \end{subfigure}
      \hfill
      \begin{subfigure}[b]{0.32\textwidth}
          \centering
          \includegraphics[width=1\columnwidth]{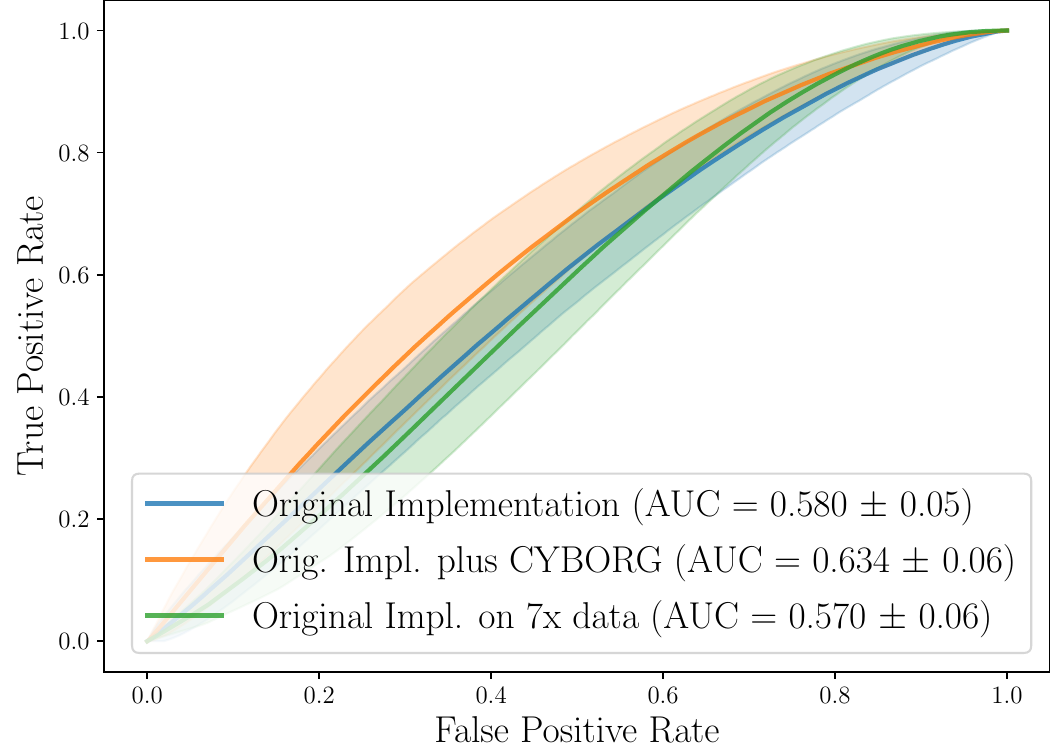}
          \caption{StyleGAN2}
      \end{subfigure}

  \end{subfigure}\vskip3mm
  \begin{subfigure}[b]{1\textwidth}
      \centering
      \begin{subfigure}[b]{0.32\textwidth}
          \centering
          \includegraphics[width=1\columnwidth]{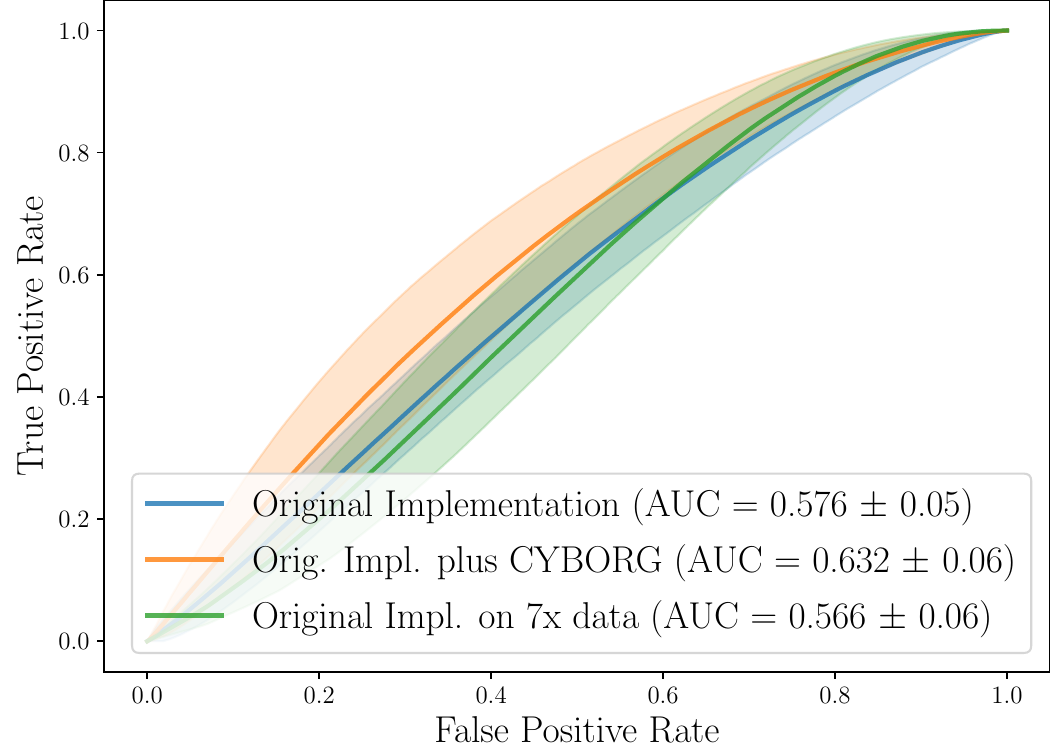}
          \caption{StyleGAN2-ADA}
      \end{subfigure}
      \begin{subfigure}[b]{0.32\textwidth}
          \centering
          \includegraphics[width=1\columnwidth]{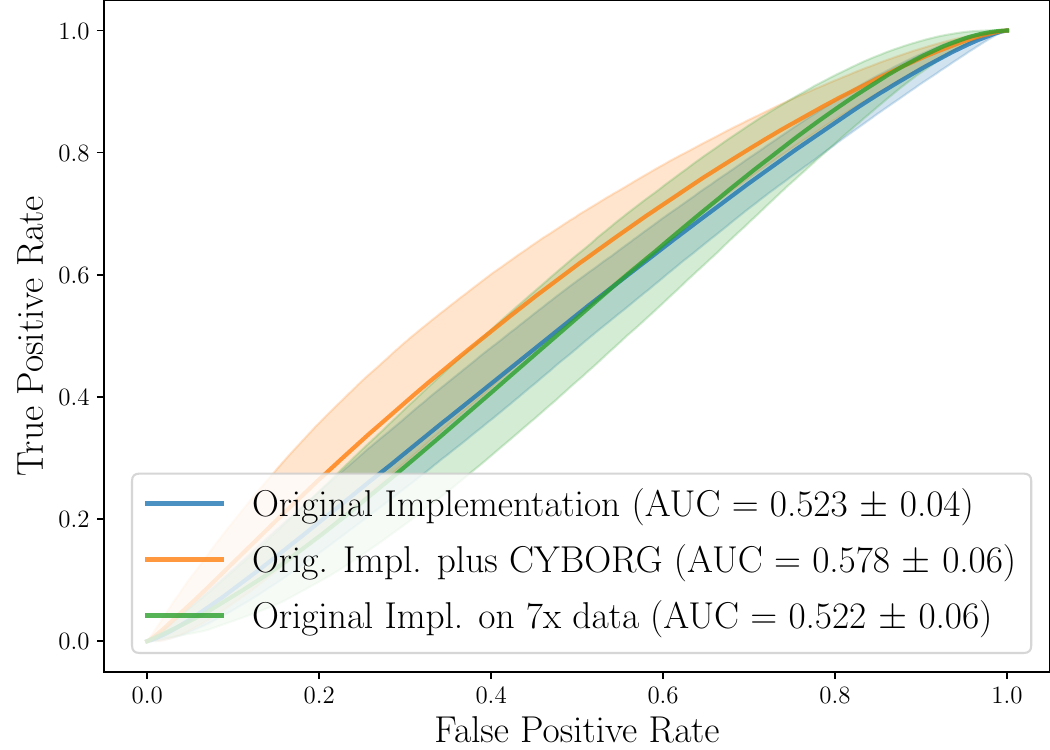}
          \caption{StyleGAN3}
      \end{subfigure}
      \begin{subfigure}[b]{0.32\textwidth}
          \centering
          \includegraphics[width=1\columnwidth]{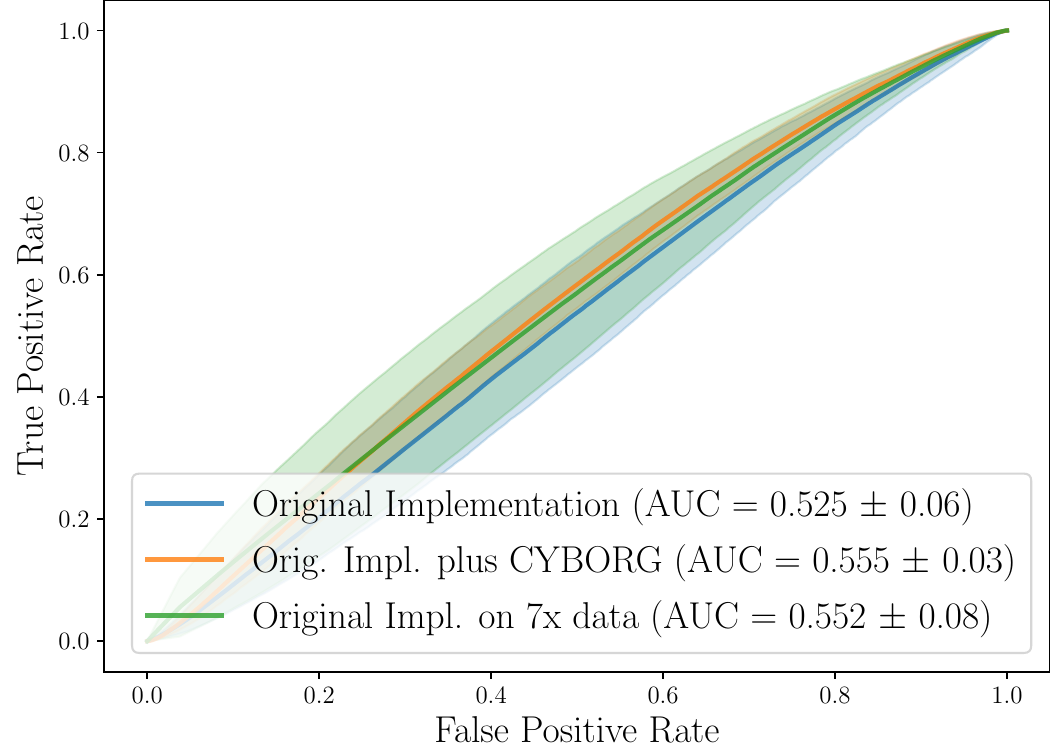}
          \caption{StarGAN v2}
      \end{subfigure}
  \end{subfigure}\vskip-1mm
\caption{Same as in Fig. \ref{fig:dn_roc}, except for classification model: {\bf CNNDetection}}
\label{roc:CNNDetection}
\end{figure*}

Results (based on the same testing scenarios) can be found in the ``CNNDetection'' row in Tab. \ref{tab:we_are_simply_the_best}, with ROCs for the individual GAN sources in Fig. \ref{roc:CNNDetection}. As can be seen, there is an increase in performance, although relatively minimal compared to that of other models.

\subsection{Incorporation Of Human Saliency Into An Attention Mechanism.}

A popular approach to force networks to focus on specified regions is self-attention. In~\cite{dang2020detection}, the authors propose a method of self-attention to give extra context to deepfake detection models via image masks. The masks in this approach signal broad areas of ``tampered'' information, instructing the model where the image alteration has been performed during training. However, attention maps are much coarser as they only provide binary information (tampered / not tampered). Additionally, attention maps are not based on human judgment, but are instead bona-fide ground truth maps of modified regions within an image.

\begin{figure*}[!htb]
\centering
  \begin{subfigure}[b]{1\textwidth}
      \begin{subfigure}[b]{0.32\textwidth}
          \centering
            \includegraphics[width=1\columnwidth]{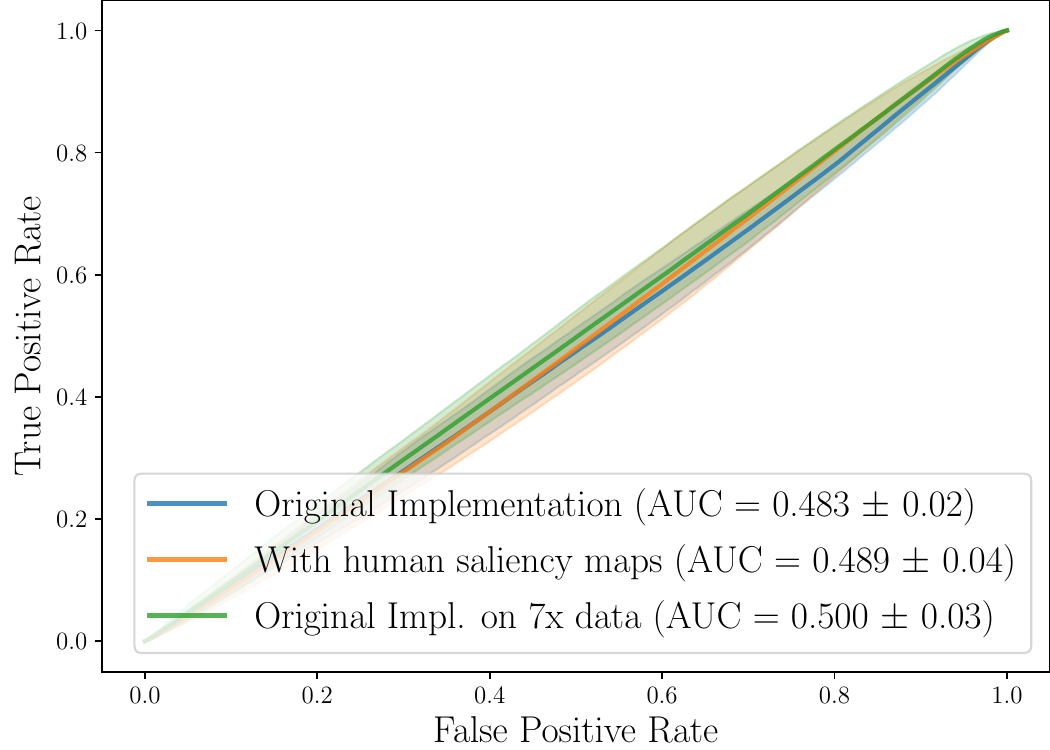}
          \caption{ProGAN}
      \end{subfigure}
      \hfill
      \begin{subfigure}[b]{0.32\textwidth}
          \centering
          \includegraphics[width=1\columnwidth]{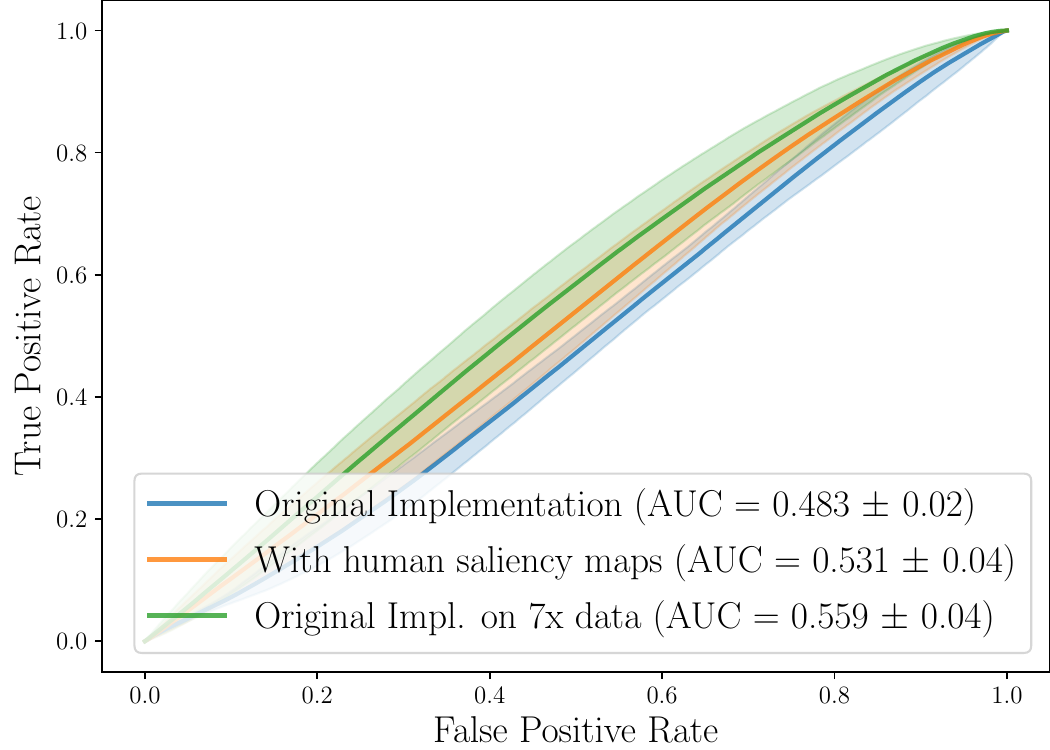}
          \caption{StyleGAN}
      \end{subfigure}
      \hfill
      \begin{subfigure}[b]{0.32\textwidth}
          \centering
          \includegraphics[width=1\columnwidth]{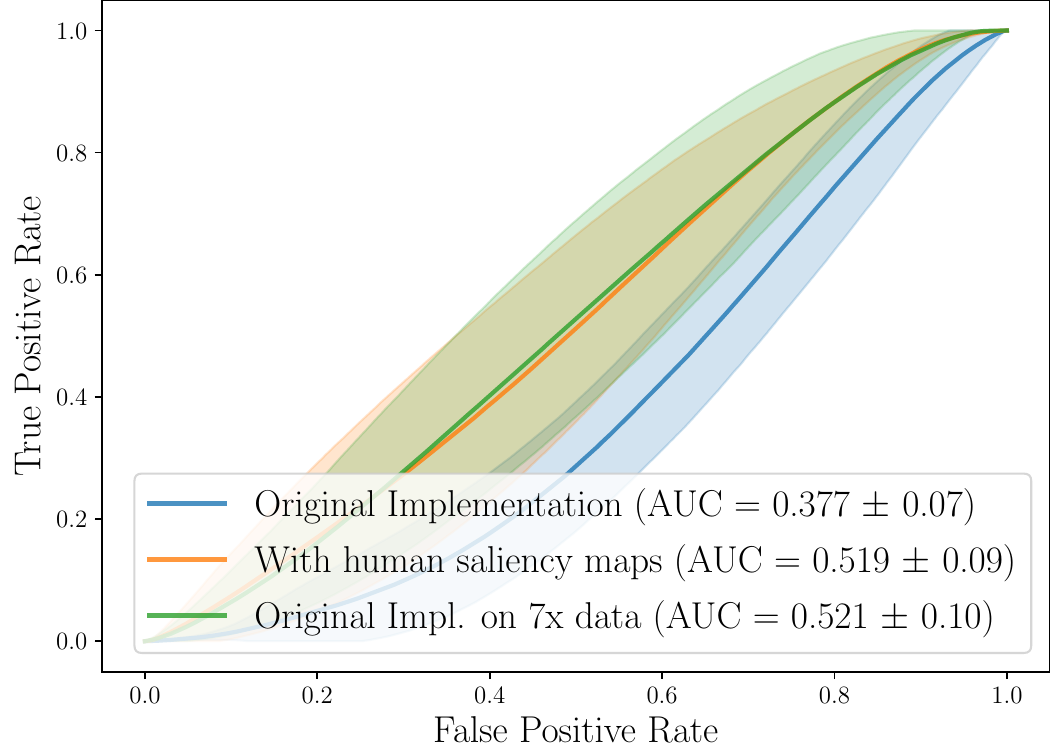}
          \caption{StyleGAN2}
      \end{subfigure}

  \end{subfigure}\vskip3mm
  \begin{subfigure}[b]{1\textwidth}
      \centering
      \begin{subfigure}[b]{0.32\textwidth}
          \centering
          \includegraphics[width=1\columnwidth]{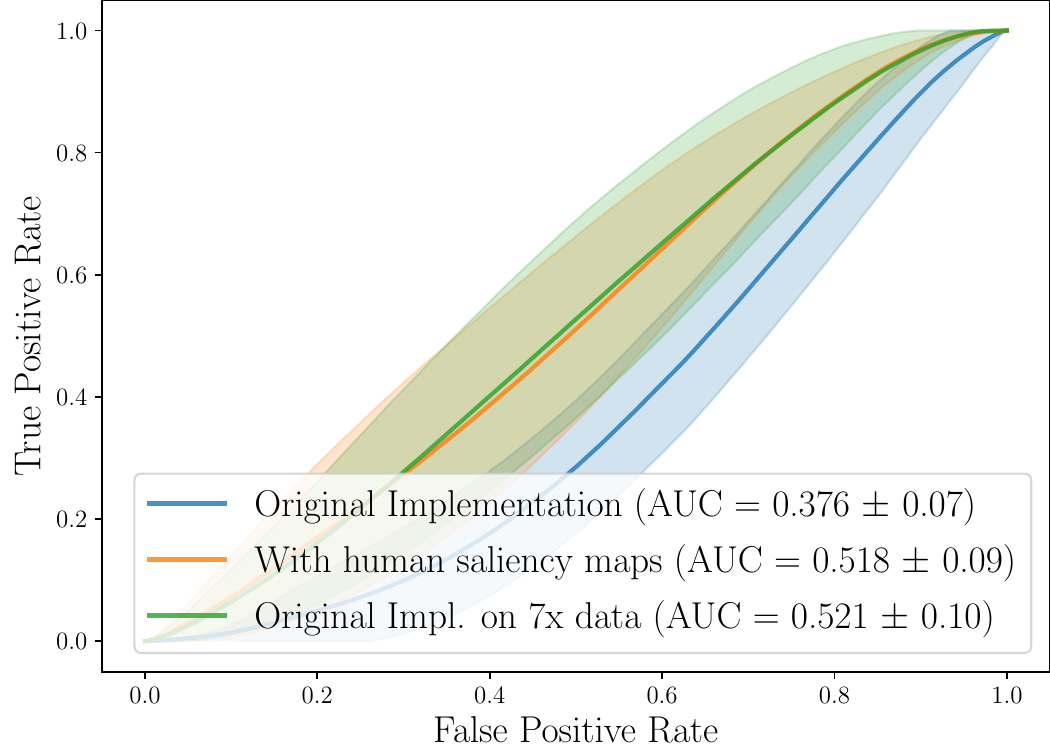}
          \caption{StyleGAN2-ADA}
      \end{subfigure}
      \begin{subfigure}[b]{0.32\textwidth}
          \centering
          \includegraphics[width=1\columnwidth]{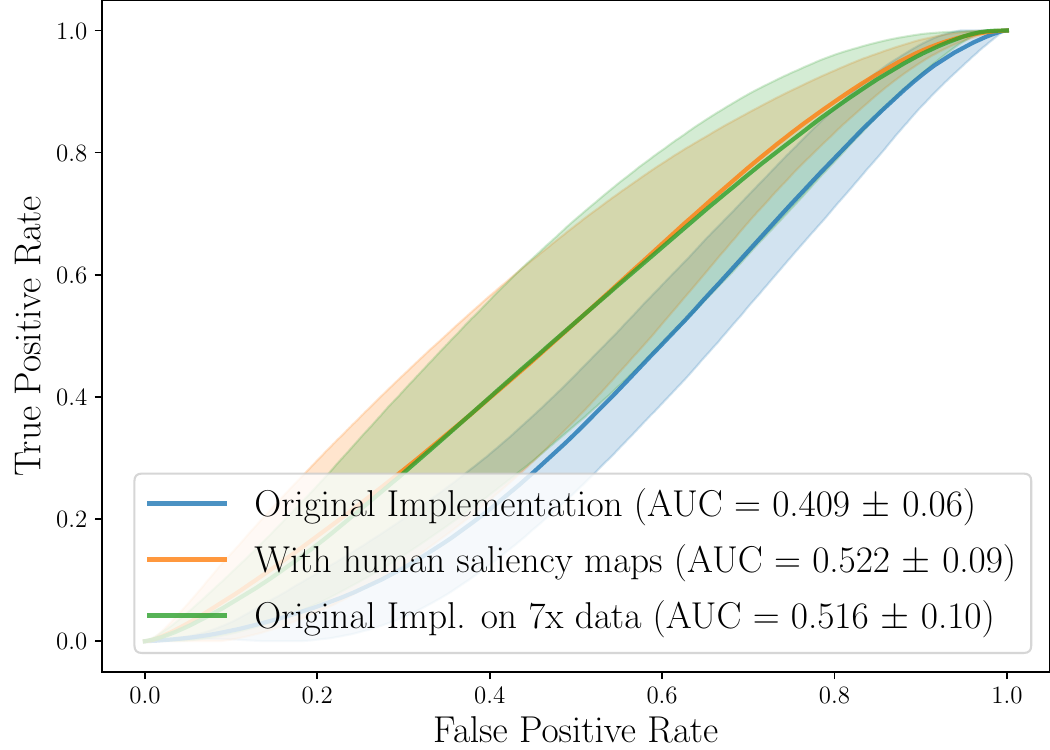}
          \caption{StyleGAN3}
      \end{subfigure}
      \begin{subfigure}[b]{0.32\textwidth}
          \centering
          \includegraphics[width=1\columnwidth]{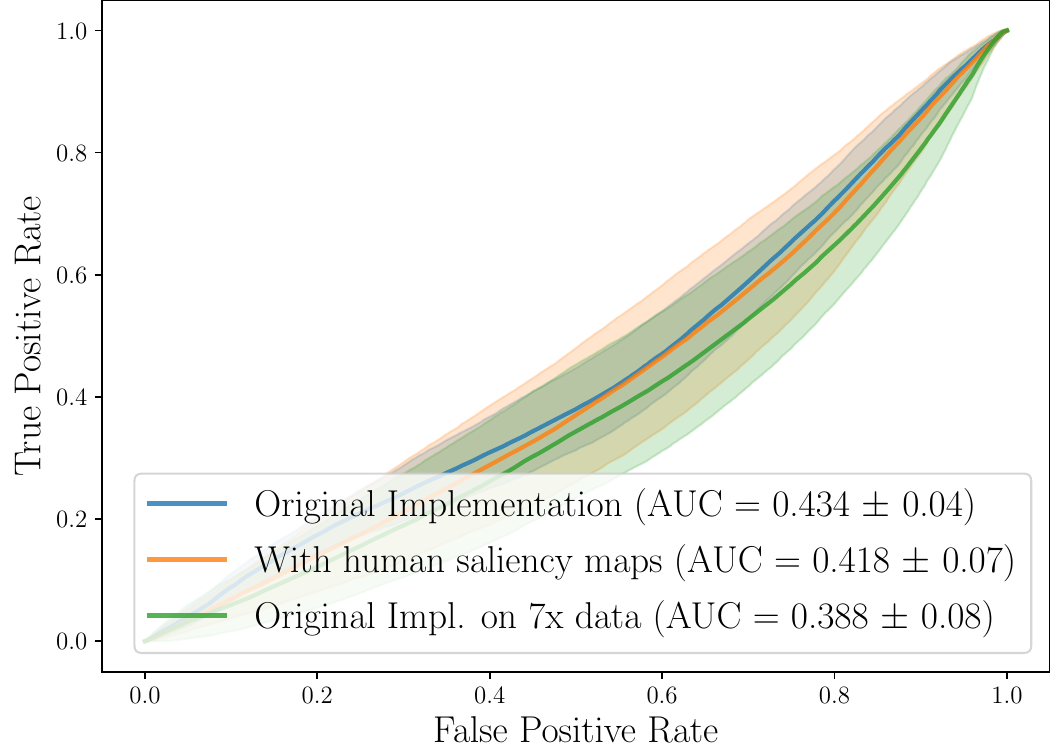}
          \caption{StarGAN v2}
      \end{subfigure}
  \end{subfigure}\vskip-1mm
\caption{Same as in Fig. \ref{fig:dn_roc}, except for classification model: {\bf Self-attention}}
\label{roc:attention}
\end{figure*}

Although not the main goal of this work, we investigate whether the replacement of the masks proposed in~\cite{dang2020detection} with our human saliency maps 
results in higher accuracy. Because our work focuses on detecting synthetically generated images as a whole, rather than tampered images, real image masks are all zeros and synthetically-generated image masks are all ones, as described in~\cite{dang2020detection}. We train two models in this case: (1) using the original approach with no human saliency, and (2) using our human saliency maps as the masks for real and synthetic images. In both cases, the parameters proposed by the authors are used. The goal of this evaluation is to see whether addition of a self-attention module and human saliency information can also provide a boost in generalization.
The ``Self-Attention'' rows in Tab. \ref{tab:we_are_simply_the_best} illustrate that the replacement of the ground truth masks used in~\cite{dang2020detection} with our human saliency maps increased performance. ROCs on individual GAN sources can also be found in Fig. \ref{roc:attention}. Since GAN-generated images are entirely synthetic, ground truth data regarding synthetic \textit{regions} within an image typically does not exist. Tab. \ref{tab:we_are_simply_the_best} suggests that implanting human saliency maps into the self-attention module narrows the model's search for areas of importance (even in the absence of ground truth) and boosts performance.

\newcolumntype{B}{<{\hspace{1ex}}c}
\setlength{\dashlinedash}{1pt}
\setlength{\dashlinegap}{1pt}
\setlength{\arrayrulewidth}{0.4pt}

\clearpage
\noindent

\begin{landscape}

\begin{table}[!htb]

\caption{Area Under Curve (AUC) with $\pm$ one standard deviation (over 10 independent runs of the training-validation experiments) for all combinations of classification models (rows), synthetic face generative models (table section headers) and training strategies (Scenario 1: Classical with cross-entropy loss only; Scenario 2: Classical with Large Data and cross-entropy loss only; Scenario 3: CYBORG -- the proposed approach penalizing both the divergence of the model from human saliency and classification performance). Best average AUCs among all scenarios are bold and color-coded: \bfstatB{blue for Classical (Scenario 1)}, \bfstatG{green for Classical with Large Data (Scenario 2)}, and \bfstatO{orange for CYBORG (Scenario 3)}. 
We can see that in majority of cases, the CYBORG approach results in higher AUCs in a task of recognition of synthetic faces generated by unknown GAN models. 
}
\label{tab:we_are_simply_the_best}

\scriptsize
\vskip5mm
\begin{tabular}{rBBBBBBBBB}
\toprule

& \multicolumn{3}{c}{\bf ProGAN} & \multicolumn{3}{c}{\bf StyleGAN} & \multicolumn{3}{c}{\bf StyleGAN2} \\\cmidrule(lr){2-4}\cmidrule(lr){5-7}\cmidrule(lr){8-10} & 
Classical & CYBORG & Classical -- & 
Classical & CYBORG & Classical -- & 
Classical & CYBORG & Classical -- \\ &
 &  &  Large Data & 
 &  &  Large Data & 
 &  &  Large Data \\\hline

\bf DenseNet121 & 
0.580 $\pm$ 0.04 & \bfstatO{0.702 $\pm$ 0.04} & 0.601 $\pm$ 0.03 &
0.563 $\pm$ 0.04 & \bfstatO{0.645 $\pm$ 0.03} & 0.629 $\pm$ 0.05 &
0.529 $\pm$ 0.07 & 0.704 $\pm$ 0.06 & \bfstatG{0.716 $\pm$ 0.05} \\

\bf ResNet50 & 
0.554 $\pm$ 0.06 & \bfstatO{0.668 $\pm$ 0.04} & 0.599 $\pm$ 0.03 &
0.561 $\pm$ 0.09 & \bfstatO{0.629 $\pm$ 0.04} & 0.611 $\pm$ 0.05 &
0.556 $\pm$ 0.14 & 0.664 $\pm$ 0.06 & \bfstatG{0.690 $\pm$ 0.07} \\

\bf Inception v3 & 
0.595 $\pm$ 0.07 & \bfstatO{0.785 $\pm$ 0.04} & 0.575 $\pm$ 0.03 &
0.604 $\pm$ 0.05 & \bfstatO{0.692 $\pm$ 0.05} & 0.584 $\pm$ 0.04 &
0.630 $\pm$ 0.08 & \bfstatO{0.801 $\pm$ 0.06} & 0.697 $\pm$ 0.05 \\

\bf Xception Net & 
\bfstatB{0.740 $\pm$ 0.03} & 0.725 $\pm$ 0.02 & 0.616 $\pm$ 0.02 &
0.704 $\pm$ 0.02 & \bfstatO{0.754 $\pm$ 0.02} & 0.586 $\pm$ 0.02 &
0.826 $\pm$ 0.03 & \bfstatO{0.873 $\pm$ 0.03} & 0.710 $\pm$ 0.02\\\hdashline

\bf CNN Det. & 
0.521 $\pm$ 0.04 & \bfstatO{0.525 $\pm$ 0.03} & 0.461 $\pm$ 0.04 &
0.583 $\pm$ 0.02 & \bfstatO{0.618 $\pm$ 0.03} & 0.568 $\pm$ 0.04 &
0.580 $\pm$ 0.05 & \bfstatO{0.634 $\pm$ 0.06} & 0.570 $\pm$ 0.06\\\hdashline

\bf Self-Attention & 
0.483 $\pm$ 0.02 & \bfstatO{0.489 $\pm$ 0.04} & 0.500 $\pm$ 0.03 &
0.483 $\pm$ 0.02 & \bfstatO{0.531 $\pm$ 0.04} & 0.559 $\pm$ 0.04 &
0.377 $\pm$ 0.07 & \bfstatO{0.519 $\pm$ 0.09} & 0.521 $\pm$ 0.10 \\\hdashline

\bf Deepfake Det. & 
& & &
& & &
& &\\

\midrule

& \multicolumn{3}{c}{\bf StyleGAN2-ADA} & \multicolumn{3}{c}{\bf StyleGAN3} & \multicolumn{3}{c}{\bf StarGANv2} \\
\cmidrule(lr){2-4}\cmidrule(lr){5-7}\cmidrule(lr){8-10} &
Classical & CYBORG & Classical -- & 
Classical & CYBORG & Classical -- & 
Classical & CYBORG & Classical -- \\ &
 &  &  Large Data & 
 &  &  Large Data & 
 &  &  Large Data \\\hline

\bf DenseNet121 & 
0.528 $\pm$ 0.07 & \bfstatO{0.705 $\pm$ 0.06} & 0.714 $\pm$ 0.05 &
0.489 $\pm$ 0.08 & \bfstatO{0.614 $\pm$ 0.06} & 0.581 $\pm$ 0.05 &
0.403 $\pm$ 0.05 & \bfstatO{0.513 $\pm$ 0.07} & 0.358 $\pm$ 0.07 \\

\bf ResNet50 & 
0.552 $\pm$ 0.14 & 0.665 $\pm$ 0.07 & \bfstatG{0.681 $\pm$ 0.06} &
0.520 $\pm$ 0.12 & \bfstatO{0.594 $\pm$ 0.07} & 0.573 $\pm$ 0.06 &
\bfstatB{0.520 $\pm$ 0.10} & 0.511 $\pm$ 0.07 & 0.372 $\pm$ 0.07 \\

\bf Inception v3 & 
0.631 $\pm$ 0.08 & \bfstatO{0.808 $\pm$ 0.05} & 0.697 $\pm$ 0.06 &
0.557 $\pm$ 0.07 & \bfstatO{0.701 $\pm$ 0.10} & 0.546 $\pm$ 0.04 &
\bfstatB{0.468 $\pm$ 0.13} & \bfstatO{0.468 $\pm$ 0.07} & 0.305 $\pm$ 0.05 \\

\bf Xception Net & 
0.818 $\pm$ 0.03 & \bfstatO{0.868 $\pm$ 0.03} & 0.699 $\pm$ 0.03 &
0.701 $\pm$ 0.03 & \bfstatO{0.771 $\pm$ 0.03} & 0.500 $\pm$ 0.03 &
0.431 $\pm$ 0.05 & \bfstatO{0.473 $\pm$ 0.02} &  0.366 $\pm$ 0.04\\\hdashline

\bf CNN Det. & 
0.576 $\pm$ 0.05 & \bfstatO{0.632 $\pm$ 0.06} & 0.566 $\pm$ 0.06 &
0.523 $\pm$ 0.04 & \bfstatO{0.578 $\pm$ 0.06} & 0.522 $\pm$ 0.06 &
0.525 $\pm$ 0.06 & \bfstatO{0.555 $\pm$ 0.03} & 0.552 $\pm$ 0.08\\\hdashline

\bf Self-Attention & 
0.376 $\pm$ 0.07 & 0.518 $\pm$ 0.09 & \bfstatG{0.521 $\pm$ 0.10} &
0.409 $\pm$ 0.06 & \bfstatO{0.522 $\pm$ 0.09} & 0.516 $\pm$ 0.10 &
\bfstatB{0.434 $\pm$ 0.04} & 0.418 $\pm$ 0.07 & 0.388 $\pm$ 0.08 \\\hdashline

\bf Deepfake Det. & 
& & &
& & &
& &\\
\bottomrule

\end{tabular}

\end{table}

\end{landscape}

\end{document}